\documentclass[12pt,reqno]{amsart}

\usepackage[left=1in,right=1in,top=1in,bottom=1in]{geometry}
\usepackage{amsmath, amssymb, graphicx, url, algorithm2e,mathtools}

\usepackage{xcolor}
\usepackage{wrapfig}
\usepackage{tikz}
\usepackage{parskip}
\usepackage{algorithmic}
\usepackage{booktabs}
\usepackage{float}
\newcommand{\bbI}{\mathrm{I}}
\newcommand{\bbA}{\mathrm{A}}

\newcommand{\ldot}[2]{\langle #1, #2 \rangle}

\newcommand{\starhull}{\mathrm{star}}

\usepackage{mathptmx}
\usepackage{tikz}
\usepackage{multirow}
\usepackage{mathtools} 
\usepackage{amsthm}
\usepackage{thmtools}

\newcommand{\vertiii}[1]{{\left\vert\kern-0.25ex\left\vert\kern-0.25ex\left\vert #1 
    \right\vert\kern-0.25ex\right\vert\kern-0.25ex\right\vert}}

\usepackage{xr}

\usepackage{graphicx}
\usepackage{cancel}
\usepackage{subcaption}
\usepackage{color}
\usepackage{wrapfig}
\usepackage{mathtools}
\usepackage{placeins}
\usepackage[framemethod=TikZ]{mdframed}
\usepackage{xcolor}
\usepackage{tikz}
\usetikzlibrary {positioning}

\usepackage{bm}
\usepackage{paralist}
\usepackage{accents}
\usepackage{enumerate}
\usepackage[round, longnamesfirst]{natbib} 

\makeatletter
\newcommand*{\addFileDependency}[1]{% argument=file name and extension
  \typeout{(#1)}
  \@addtofilelist{#1}
  \IfFileExists{#1}{}{\typeout{No file #1.}}
}
\makeatother

\renewcommand{\Pr}{{\mathrm{P}}}

\newcommand{\E}{\mathrm{E}}

\usepackage{nccmath}
		{\end{pmatrix}\end{medsize}}%

\newtheorem{theorem}{Theorem}[section]
\newtheorem{assumption}{Assumption}[section]
\newtheorem{proposition}{Proposition}[section]
\newtheorem{corollary}{Corollary}[section]
\newtheorem{lemma}{Lemma}[section]

\theoremstyle{definition}

\newtheorem{example}{Example}[section]
\newtheorem{remark}{Remark}[section]

\renewcommand\thmcontinues[1]{Continued}

\numberwithin{equation}{section}

\usepackage{wrapfig}

\begin{document}

\setcounter{page}{1}

\title[ Welfare Analysis in Dynamic Models
]{Welfare Analysis in Dynamic Models}%
% VS: 

\author{Victor Chernozhukov}\thanks{MIT, vchern@mit.edu} 
\author{Whitney K. Newey}\thanks{MIT, wnewey@mit.edu }
\author{ Vira Semenova}\thanks{University of California, Berkeley, vsemenova@berkeley.edu} 
%\today
\date{\today; Initial ArXiv Submission: August 2019, arXiv ID. 1908.09173. This research was supported by NSF grants 1757140 and 224247. Benjamin Deaner, David Donaldson, Bryan Graham, Michael Jansson, Patrick Kline, Lihua Lei, Robert Miller, Demian Pouzo, John Rust, and participants of the 2019 and 2023 Dynamic Structural Estimation Conferences and ASSA 2024 Econometrica session provided helpful comments. }
\makeatletter
\def\thanks#1{\protected@xdef\@thanks{\@thanks
        \protect\footnotetext{1}}}
\makeatother

\makeatletter
\def\thanks#1{\protected@xdef\@thanks{\@thanks
        \protect\footnotetext{1}}}
\makeatother

\begin{abstract}

This paper introduces metrics for welfare analysis in dynamic models. We develop estimation and inference for these parameters even in the presence of a high-dimensional state space. Examples of welfare metrics include average welfare, average marginal welfare effects, and welfare decompositions into direct and indirect effects similar to \cite{oaxaca73} and \cite{blinder73}. 
We derive dual and doubly robust representations of welfare metrics that facilitate debiased inference. For average welfare, the value function does not have to be estimated.
In general, debiasing can be applied to any estimator of the value function, including neural nets, random forests, Lasso, boosting, and other high-dimensional methods. In particular, we derive Lasso and Neural Network estimators of the value function and associated dynamic dual representation and establish associated mean square convergence rates for these functions. 
Debiasing is automatic in the sense that it only requires knowledge of the welfare metric of interest, not the form of bias correction.
 The proposed methods are applied to estimate a dynamic behavioral model of teacher absenteeism in \cite{DHR} and associated average teacher welfare. 

\end{abstract}

\maketitle

\newpage

\section{Introduction}

Dynamic considerations are important in applied work \citep{Miller1984, Wolpin, Pakes:1986, Rust, HotzMiller, HotzMillerSandersSmith, keane1994solution, AMira2002,AMira2007,BBL}. 
These considerations are captured by the value function, defined as the present discounted value of agents' expected per-period utility. 
Prior work has focused on identifying utility parameters from the optimality of agents' behavior and conducting inference on them. Building on this foundation, this paper targets a different yet complementary class of parameters that depend on the value function which we introduce below.

Welfare analysis is a central objective in economics.
This paper develops identification, estimation, and inference results for welfare metrics in dynamic models.  
Examples of such metrics include the average value function and average marginal welfare effects, e.g., due to the change in observable time-invariant characteristics such as initial wealth. Most importantly, we decompose welfare effects into the direct effect through changes in utilities holding agents' behavior fixed and the indirect effect from  changes in agents' behavior. This decomposition is analogous in spirit to Oaxaca-Blinder decomposition of outcome distributions yet different since welfare metrics are based on latent utilities rather than observed outcomes.
% to welfare in dynamic models.

The economic motivation for the welfare decomposition we propose is drawn from various fields in applied economics, including labor economics, public finance, and health economics. In labor economics, \cite{kitagawa55}, \cite{oaxaca73} and \cite{blinder73} decompose the gender wage gap into structural and composition effects.  In public finance, it is common to
distinguish  between mechanical and behavioral  effects of taxation policies, see e.g. \cite{Chetty2009}. In health economics, the work by \cite{EinavFinkelsteinCullen2010}  differentiates the direct effect of a health insurance price change from the indirect effect arising from changes in selection. For an example of cancer screening \citep{EinavEtAl2020}, the total welfare effect corresponds to the introduction of a new screening option, and the direct effect is its counterpart as if the frequency of screening were mandated. One of the paper's contribution is to give identification, estimation, and inferential results for direct and indirect effects in dynamic models. 

The next contribution of the paper is the dynamic dual representation of welfare metrics, which directly maps the per-period utility to the welfare metric of interest. For the case of average welfare, the dynamic dual representation reduces to a known expression of expected per-period utility. Furthermore, neither value function nor  other dynamic object needs to be estimated for certain discrete choice models, such as those of   \cite{Rust} and \cite{HotzMiller}.  For cases beyond average welfare, the dynamic dual representation involves backward discounting. In these cases,  we derive a doubly robust representation that facilitates consistent estimation if at least one of the two nuisance components, the value function or the dynamic dual representation,  is correct. Appendix \ref{sec:atheory} gives large sample properties of the proposed estimators based on Neyman-orthogonal moment equations, in which the dynamic dual representation plays a central role. This paper is the first result in the literature to leverage duality in dynamic models whose state distribution is strictly stationary. 

Another contribution of the paper is to introduce a Lasso estimator for the value function, the dynamic dual representation, and associated mean square convergence rates in the setting of high-dimensional state variables. Specifically, the number of state variables may exceed the sample size, provided that only a small subset is relevant. We derive a novel least squares criterion that distinguishes our approach from previous IV-based methods. Adding an $\ell_1$ penalty to the sample analog of this criterion yields a Lasso estimator of the value function. In addition, we propose a neural network estimator and derive corresponding mean square convergence rates in the low-dimensional case. Although the results are presented for the value function, the methods apply to any fixed-point solution of an integral equation of the second kind -- such as the Q-function in reinforcement learning -- and do not require strict stationarity of the state distribution.

We revisit the study of teacher absenteeism in the nonformal education centers (NFEs) of Rajasthan, India, initially analyzed by \citet{DHR} (henceforth, DHR). Their approach emphasizes unobserved heterogeneity in teachers' leisure preferences—shaped by factors such as past effort, illness, fatigue, and informal obligations. We conjecture that this heterogeneity can be effectively proxied by a sufficiently long window of prior work history, even if the history itself may not have a  structural or causal interpretation. DHR's structural 
estimates remain robust to this exercise. In addition, we focus on teacher welfare and find that the failure to properly account for teacher heterogeneity results in the average teacher welfare overestimated by 13–20\%\%.  As a side contribution, we demonstrate that our methodological framework remains valid in finite-horizon settings, provided the panel is sufficiently long relative to standard discount factors.

 \subsection{Literature Review}
 \label{sec:litreview}

A large body of work is dedicated to welfare analysis in discrete choice models with unobserved heterogeneity, as studied, e.g., in \citet{Bhattacharya2015}. As \citet{Bhattacharya2024} discusses, welfare calculations are based on latent utilities rather than observed outcomes. For dynamic choice -- the focus of this paper -- important early references include \citet{Miller1984}, \citet{Wolpin}, and \citet{Pakes:1986}. Within this group, a notable subclass of models includes those with the terminal action property, as in \citet{HotzMiller} and \citet{HotzMillerSandersSmith}, and, more broadly, models with finite dependence, as in \citet{ArcidiaconoMillerFD}. In \citet{HotzMiller}, structural parameters are identified via a regression problem that depends on conditional choice probabilities. The dynamic dual representation we derive extends this insight to the average value function and other welfare metrics.

Estimation of dynamic discrete choice models has received substantial attention; see, e.g., \citet{AMira2002, AMira2007, BBL, PSM, ArcidiaconoMiller, Arcidiacono:2013, ArcidiaconoEtAl2013, ChenAck, AMagesan, BShum}. Recent work by \citet{AdsmEck2022} relaxes the terminal action property while allowing for an infinitely supported state space. Appendix~\ref{sec:nonlinear} of the present paper outlines an automatic debiasing approach that targets nonlinear functionals of the value function and thus could be applicable to the model in \citet{AdsmEck2022}.

Last but not least, we contribute to the literature on estimating fixed points of integral equations of the second kind; see, e.g., \citet{SLinton}. Prior work by \citet{chen2022wellposedness} has shown that this is a special case of a well-posed NPIV problem and has developed value function estimators that achieve minimax-optimal rates in low-dimensional settings; see also \citet{xia2022krylovbellmanboostingsuperlinearpolicy} for related IV-based approaches. However, these methods may not readily extend to penalized estimators that remain consistent in high-dimensional settings. The least squares criterion we derive circumvents this limitation. We also give a least squares criterion for estimating the dynamic dual representation, that only depends on the welfare metric of interest, and so enables automatic debiasing like  \cite{chernozhukov2021automatic} and \cite{chernozhukov2024qm}. 
 
 The paper is organized as follows.  Section \ref{sec:setup} presents the general framework and examples of welfare metrics.  
Section \ref{sec:oaxaca} decomposes the  differences in average welfare in the spirit of Oaxaca and Blinder.
Section \ref{sec:empirical} revisits \citet{DHR} study of teacher absenteeism.
Section \ref{sec:dual}  gives the dynamic dual and doubly robust representations for welfare metrics.
Section \ref{sec:estimation} describes orthogonal estimating equations for average welfare and related averages.
Least squares estimators of the value function and dynamic dual representation are presented in Section \ref{sec:valuefunction}.
Section \ref{sec:conclusion} concludes.
Appendix \ref{sec:dynamicbinary} considers the example of dynamic discrete choice.
Appendix \ref{sec:proofs} gives proofs of the results in main text.
Appendix \ref{sec:atheory} gives large sample properties of the estimators.
Appendices \ref{sec:fs}--\ref{sec:firststage} give mean square rates for first-stage estimators.
Appendix \ref{sec:nonlinear} extends the proposed method to nonlinear functionals of the value function.
Appendix \ref{sec:dynamicbinary2} demonstrates the method of Appendix \ref{sec:nonlinear} for dynamic binary choice. 

%As a simulation exercise in Appendix \ref{sec:dynamicbinary2}, we revisit the dynamic discrete choice model of \citet{Rust}, augmenting it with a high-dimensional representation of observable heterogeneity across bus types. In the proposed design, all engines share the same replacement cost but differ in maintenance costs. We focus on the average value function and compare its direct plug-in estimates to their dynamic dual representations. The dual representation produces lower bias and more accurate coverage, closer to nominal levels. 

\section{Setup}
\label{sec:setup}

We consider estimation and inference on  welfare metrics that depend on the value function.  To define the value function, 
let $X_t, (t=0,1,...)$, denote a time series of observed state variables, which we assume to be a time-homogeneous, first-order Markov process  with initial element $X:=X_0$. 
The value function is determined by a per-period reward $\zeta_0(X)$, or, in other words,  expected utility in a single period conditional on the state $X$, which we assume to be identifiable. 
The value function $V_0(X)$ is the present discounted value of per-period rewards given the current state, satisfying 
\begin{align}
\label{eq:npv2}
V_0(X) = \sum_{t=0}^{\infty} \beta^t \E [\zeta_0(X_t) \mid X],
\end{align}
where $\beta \in [0,1)$ is a known discount factor.
The value function satisfies the integral equation: \begin{align}
\label{eq:fdp}
V_0(X) = \zeta_0(X) + \beta\E[V_0(X_{+}) \mid X],
\end{align}
where $X$ is the current and $X_{+}$ is the next period element of the first-order Markov process (see Lemma \ref{lem:equivalence}). 
In what follows, we assume that the time series is strictly stationary. The welfare metrics we consider are linear functionals of the value function $V_0(X)$ having the form
\begin{align}
\label{eq:targetstationary}
\delta_0 = \E [ w_0(X) V_0(X) ],
\end{align}
where $w_0(X)$ is some function of the state.  

To give an example of per-period utility, we consider the dynamic discrete choice problem \citep{Rust,HotzMiller,AMira2002}. 
In each period, $(t=0,1,...),$ the agent chooses an action $j$ from a finite choice set $\mathcal{A}$. 
The utility of choice $j$ in period $t$ is additively separable in a function of the current state $u(X_t,j)$ and private shock $\varepsilon_t(j)$ and is given by
$$\bar{u} (X_t, j, \epsilon_t) = u(X_t,j) + \epsilon_t(j), \quad j \in \mathcal{A}.$$ 
Here the sequence ${(X_{t}, J_t)}$ is strictly stationary, so the time index $t$ can be dropped. 
The per-period reward is the expected utility $\zeta_0(x)$, obtained by taking the expectation over choices:
$$
\zeta_0(x) =   \sum_{j \in \mathcal{A}} (u(x,j) + \E [ \epsilon (j) \mid X=x, J=j]) \Pr (J=j \mid X=x),
$$
where $\Pr (J=j \mid X=x)$ is the probability that an agent chooses $j$ when $X=x$, and  $\E [ \epsilon (j) \mid X=x, J=j]$ is the expectation of $\epsilon (j)$ under the choice $J=j$ and given $X=x$. For example, in a special case where the choice is binary and the private shocks are distributed as Gumbel,  
\begin{align}
\label{eq:zetax}
\zeta_0(x) = u(x,1) p_0(x) + u(x,0) (1-p_0(x)) + H(p_0(x)),
\end{align}
where $p_0(x) = \Pr (J=1 \mid X=x)$,   $\mathcal{A}=\{1, 0\}$, and $H(t) = \gamma_e - t\ln t -(1-t) \ln (1-t),$
with  $\gamma_e =  0.5227$ denoting the Euler constant. Here and generally for dynamic discrete choice the per-period utility will be the sum of the expected value of the observable part of the utility plus the expected value over optimal choices of the private shock part. We assume that the utility components $u(x,1)$ and $u(x,0)$ are known up to a structural parameter that is identified. 

Within the discrete choice problems, our target parameter $\delta_0$
represents a welfare metric since $V_0(\cdot)$ is the expected value of an agent making optimal dynamic choices conditional on state $X$.
Our first example is the expected value function where $w_0(X)=1$.

\begin{example}[Average Welfare]
\label{ex:average}
If $w_0(X)=1$, the parameter  
\begin{align}
\delta_0 = \E [V_0(X)]
\end{align} 
represents the unconditional expected value or welfare of making optimal dynamic choices. 
\end{example}

We focus particularly on settings where the state variable include observable time-invariant heterogeneity in individual, per-period expected utility denoted by a vector $K$. Examples of $K$ include occupation type \citep{KeaneWolpin1997},  gender and class grade of a child \citep{ToddWolpin} and teacher test score \citep{DHR}. Time invariant state variables could also represent observable heterogeneity in individual, per-period expected utility.     In this case, the state vector $X_t$ can be decomposed as $X_t = (S_t, K),$  where $K_t= K$ does not vary over time. 
Here first-order time homogeneity of $X_t$ implies that $(S_t)_{t > 0 }$ is a first-order time-homogeneous Markov chain conditional on $K$. 

\begin{example}[Group Average Welfare]
\label{ex:group}
When $K$ is discrete, taking on a finite number of values, the group average welfare is 
\begin{align}
\label{eq:timeinvar}
\delta_0 = \E[1(K=k)V_0(X)]/\Pr(K=k) = \E[V_0(X)\mid K=k].
\end{align}
In this case $w_0(X)=1(K=k)/\Pr(K=k)$.

\end{example}

Extending this example to differences in average welfare across groups is straightforward by differencing the parameter of interest in Example \ref{ex:group} across different values of $K$.
In this example and the others, the weight $w_0(X)$ is unknown and will need to be estimated.
The identification and estimation of $w_0(X)$ will be accounted for in the results that follow.

When $K$ represents an endowment of some resource it may be of interest to consider the welfare effect of changing the distribution of that endowment. 

\begin{example}[Average Policy Effect]
\label{ex:stock}
Let  $\pi(k)$ and $\pi^{*}(k)$ be the probability density (or mass) function of $K$ with respect to a base measure, corresponding to the actual data and a proposed policy shift. 
The average policy effect from this shift is 
\begin{align}
\delta_0 &=\E[w_0(K)V_0(X)] , \quad w_0(K) = [\pi^{*} (K) - \pi (K)]/\pi(K).\nonumber
\end{align}
This object differs from the policy effect of \cite{Stock} in being the average effect of a policy on dynamic welfare rather than the average effect on some outcome variable.
\end{example}

For continuously distributed $K$ an effect of interest could be the average effect of changing $K$ on the value function.

\begin{example}[Average Marginal Effect]
\label{ex:avder}
For continuously distributed $K$, the average derivative of the value function with respect to $K$,
\begin{align}
\label{eq:avder}
\delta_0 = \E[\partial_k V_0(X)]=\E[\partial_k V_0(S,K)],
\end{align}
measures the average change in welfare due to varying $K$.  Letting $f(K|S)$ denote the conditional PDF of $K$ given $S$, integration by parts gives equation \eqref{eq:targetstationary} with  $w_0(X)=- \partial_k \ln f(K|S)$ as long as  $f(K|S)$ is equal to zero at the boundary of the support of $K$ conditional on $S$. 
This parameter differs from the average derivative\footnote{The average marginal welfare effects herein are different from the average marginal effects of \cite{ACarro} in panel data setup, where marginal effects are taken with respect to unobserved unit heterogeneity.} of \cite{Stoker1986} in being a dynamic welfare effect rather than an outcome effect.
\end{example}

\begin{example}
Let $\mathcal{X}$ be a set and $X$ be a state vector. Define
\begin{align}
\label{eq:timeinvar2}
\delta_0 = \E[1(X \in \mathcal{X}) V_0(X)]/\Pr(X \in  \mathcal{X}) = \E[V_0(X)\mid X \in \mathcal{X}].
\end{align}
In this case the weighting function $w_0(X)=1(X \in \mathcal{X})/\Pr(X \in  \mathcal{X})$ is time varying.
\end{example}

%%% I would like to move it somewhere. 
All of the above examples of target parameters fall in the following general theoretical  framework.  
Let $Z$ denote a data vector that includes $X$ and $X_+$, and let $V$ denote a possible value function. 
Also let $m(Z,V)$ denote a function of $Z$ and the function $V(\cdot)$ (i.e. $m(Z,V)$ is a functional of $V$.) 
We consider parameters of the form
\begin{align}
\label{eq:linfunc}
\delta_0=\E[m(Z,V_0)],
\end{align}
where $\E[m(Z, V)]$ is linear in $V$. We will impose throughout that the expectation $\E[m(Z, V)]$ is mean square continuous as a function of $V$, meaning that there is a constant $C>0$ such that for all $V(X)$ with $\E[V(X)^2]<\infty$,
\begin{align}
|\E[m(Z, V)]|\leq C (\E[V(X)^2])^{1/2}.
\end{align}
By the Riesz representation theorem mean square continuity of $\E[m(Z,V)]$ is equivalent to existence of a function $w_0(X)$ with $\E[w_0(X)^2]<\infty$ such that 
\begin{align}
\label{eq:rieszweight}
\E[m(Z,V)]=\E[w_0(X)V(X)], 
\end{align}
for all $V(\cdot)$ with $\E[V(X)^2]<\infty$.
Here, we see that under mean square continuity, any parameter as in equation \eqref{eq:linfunc} can be represented as a linear function of the value function. There are many other potentially interesting examples of such welfare metrics. In the next section we consider decomposing differences in average welfare into direct and indirect components.

\section{Decomposition of Differences in Average Welfare}
\label{sec:oaxaca}

To motivate our analysis, we present a simple running example in the context of a randomized controlled trial, in which a one-shot treatment assigned at time $t = 0$ generates dynamic incentives. Suppose we aim to analyze welfare differences between the treated and control populations, denoted by $1$ and $0$, respectively. In each group, the time-varying state variable is denoted by $S$. The full state vector is $X = (S, K)$ where $K$ is the indicator of the treatment status. We adopt the notation in \cite{chernozhukovmelly}.

Let $V_0^1(s) = V_0(s,1)$ and $V_0^0(s)=V_0(s,0)$ denote the treated and control value functions. Define the treated average welfare as
\begin{align}
\delta_{\langle 1 \mid 1 \rangle} &= \E [ V_0(X) \mid K=1]
\end{align}
and the control average welfare as 
\begin{align}
\delta_{\langle 0 \mid 0 \rangle} &= \E [ V_0(X) \mid K=0].
\end{align}
Both quantities are special cases of the group average welfare defined in Example~\ref{ex:group}. For $k \in \{1, 0\}$, we have
\begin{align}
\delta_{\langle k \mid k \rangle} &= \E [ V_0(X) \mid K=k] = \int_{s} V_0^k(s) \pi^k(s) \, ds,
\end{align}
where $\pi^k(s)$ denotes the probability distribution function of $S$ conditional on $K = k$. The counterfactual welfare metric
\begin{align}
\delta_{\langle 1 \mid 0 \rangle} = \int_{s} V_0^1(s) \pi^0(s) \, ds
\end{align}
does not correspond to the group average welfare of any observable subpopulation. Instead, it is constructed by integrating the treated value function with respect to the stationary distribution of the control population. Provided that $\pi^1(s)$ and $\pi^0(s)$ share the same support, this parameter is well-defined.

The treatment-control welfare difference can be decomposed in the spirit of \citet{kitagawa55}, \citet{oaxaca73}, and \citet{blinder73}:
\begin{align}
\delta_{\langle 1 \mid 1 \rangle} - \delta_{\langle 0 \mid 0 \rangle} 
= \left[ \delta_{\langle 1 \mid 0 \rangle} - \delta_{\langle 0 \mid 0 \rangle} \right] 
+ \left[ \delta_{\langle 1 \mid 1 \rangle} - \delta_{\langle 1 \mid 0 \rangle} \right]. 
\label{eq:oaxaca}
\end{align}
If treatment is randomly assigned, this welfare difference admits a causal interpretation.

The proposed decomposition has an intuitive interpretation when the treatment affects only per-period utilities and does not impact the state transition. We describe such empirical settings in Examples~\ref{ex:1} and~\ref{ex:2} below.  In this case, the stationary distributions $\pi^1(s)$ and $\pi^0(s)$ differ solely due to agents making different optimal choices in the treated and control states, respectively. The first summand,
\begin{align}
\label{eq:counterfact}
\delta_{\langle 1 \mid 0 \rangle} - \delta_{\langle 0 \mid 0 \rangle} &= \int_{\mathcal{S}} \left( V_0^1(s) - V_0^0(s) \right) \pi^0(s) \, ds,
\end{align}
captures the difference in per-period utilities  holding the distribution of states fixed at the control level. This term can be interpreted as the direct or mechanical effect. The second summand,
\begin{align}
\label{eq:counterfact2}
\delta_{\langle 1 \mid 1 \rangle} - \delta_{\langle 1 \mid 0 \rangle} &= \int_{\mathcal{S}} V_0^1(s) \left( \pi^1(s) - \pi^0(s) \right) \, ds,
\end{align}
reflects changes in agents' behavior that alter the distribution of states, holding the utilities fixed. This term can be interpreted as the indirect or behavioral effect.
The sum of the direct and behavioral effects gives the total welfare difference.

\begin{example}[School attendance]
\label{ex:1}
\citet{DHR} estimates a dynamic behavioral structural model in which teachers choose between working and taking leisure.
\citet{DHR} focuses  the student achievements as the primary target.
In this paper, we take a complementary perspective  and focus on teacher welfare.

Let $K = 1$ indicate the treatment status, where treatment constitutes a cash bonus for each additional day of work once the count of days worked exceeds 10 in a given month.  
The state variable $S$ denotes the number of days worked since the beginning of the month. The full state vector is $X = (S, K)$. 
The direct effect, $\delta_{\langle 1 \mid 1 \rangle} - \delta_{\langle 1 \mid 0 \rangle}$, captures the impact of the bonus on welfare through per-period utilities, holding teachers work decisions fixed. 
The indirect effect, $\delta_{\langle 1 \mid 0 \rangle} - \delta_{\langle 0 \mid 0 \rangle}$, arises from the change in teachers work decisions, holding per-period utilities fixed. 
A similar decomposition applies to the model in \citet{ToddWolpin}.
\end{example}

\begin{example}[Breast cancer screening]
\label{ex:2}
A standard approach to evaluating welfare in the context of cancer screening focuses on mortality. Yet, a broader perspective considers the costs of screening—such as time, resource use, and the psychological burden of false positives— which cannot be captured by observable outcomes. 

Let  $K = 1$ indicate assignment to a novel breast cancer screening technology. The state variable $S$ denotes the time elapsed since the most recent screening, and the full state vector is $X = (S, K)$. 
The direct effect, $\delta_{\langle 1 \mid 1 \rangle} - \delta_{\langle 1 \mid 0 \rangle}$, captures the impact of the new technology on welfare through per-period utilities, holding screening behavior fixed. The indirect effect, $\delta_{\langle 1 \mid 0 \rangle} - \delta_{\langle 0 \mid 0 \rangle}$, reflects changes in screening behavior induced by the treatment, holding utilities fixed.
\end{example}

Proposition~\ref{lem:oaxaca} expresses the counterfactual welfare $\delta_{\langle 1 \mid 0 \rangle}$ as a linear functional of the treated value function.

\begin{proposition}[Decomposition of Differences in Average Welfare]
\label{lem:oaxaca}
Suppose both density functions $\pi^0(s)$ and $\pi^1(s)$ have the same support of the state variable $S$. Then the  counterfactual welfare $ \delta_{\langle 1 \mid 0 \rangle}$ is  a special case of equation \eqref{eq:linfunc} with 
\begin{align}
m(Z, V) &= \dfrac{V(S, 1) 1\{ K = 0\}}{\Pr (K=0)} \label{eq:mzvoaxaca}
\end{align}
whose Riesz representation \eqref{eq:rieszweight} holds with 
\begin{align}
\label{eq:compositionweight}
 w_0(X) = w_0(S,K) =\dfrac{ 1\{ K=1\}  } {\Pr (K=0)}  \dfrac{\Pr (K=0 \mid S)}{\Pr (K=1 \mid S)}.
 \end{align}
\end{proposition}

Recent work has extended classical decomposition methods to modern settings. \citet{chernozhukov2021automatic} derive the Riesz representer for the Average Treatment Effect on the Treated (ATET). 
\citet{vafa2024estimatingwagedisparitiesusing} provides an Oaxaca-Blinder decomposition of wage differences. 
Proposition~\ref{lem:oaxaca} departs from these approaches by offering a decomposition of average welfare in dynamic models where welfare is based on latent utilities rather than observed outcomes.

 We include the average counterfactual welfare $\delta_{\langle 1 \mid 0 \rangle}$ as Example \ref{ex:oaxaca} and discuss its estimation in Section \ref{sec:estimation}. Notice that the parameters $ \delta_{\langle 1 \mid 1  \rangle}$ and $ \delta_{\langle 0 \mid 0 \rangle}$ are special cases of Example \ref{ex:group} with $k=1$ and $k=0$, respectively. Therefore,  it is straightforward to extend this example to accommodate direct and indirect effects.
We discuss the estimation of the counterfactual welfare measure and related effects further in Section \ref{sec:estimation}.

\begin{remark}[Overview of Related Literature]
\citep{kitagawa55,oaxaca73,blinder73} pioneered the use of least squares methods for decomposing differences in average outcomes, such as wages. This approach was later extended to the distributional setting in \cite{dfl96,mm05,flf11,chernozhukovmelly}, with \cite{chernozhukovmelly} also providing inference methods. Other related contributions include \citet{OaxacaRansom,KlineOaxaca2011,chernozhukovmelly,KlineOaxaca,GuoBasse2021,vafa2024estimatingwagedisparitiesusing}.
In the canonical wage example, groups $1$ and $0$ correspond to men and women, respectively. The direct effect captures the difference in wage schedules faced by men and women and is often interpreted as a measure of discrimination or preferential treatment. The indirect effect reflects differences in job-related characteristics, such as skills or experience, and is typically interpreted as a composition or selection effect.
\end{remark}

\section{\citet{DHR} revisited}
\label{sec:empirical}

We study how daily financial incentives affect teacher attendance in single-teacher nonformal education centers (NFEs) operated by the NGO Seva Mandir in tribal villages of Udaipur, Rajasthan, India. From 2003 to 2005, \citet{DHR} conducted a randomized trial in which tamper-proof cameras recorded photographs at school opening and closing. A school day was deemed valid if the two images were at least five hours apart and at least eight students were present. At the end of each month, teachers earned a base salary of 500 Rupees (Rs) if they worked fewer than 10 days, plus a 50 Rs bonus for each additional day of work beyond that threshold. The 10-day cutoff thus created a nonlinear dynamic incentive, which we focus on in this application.\footnote{We abstract from the firing threat, which appears negligible: no teacher was fired during the study period, even in cases of near-total absence. According to \citet{DHR}, Seva Mandir adopts a long-term view in assessing teacher performance, which may explain the lack of dismissals.} Our dataset comprises daily attendance records for 57 teachers over an 18-month period (January 1, 2004 to June 30, 2005), along with a test score administered prior to the start of teaching.\footnote{Following \citet{DHR}, the estimation sample includes only weekdays when teachers actively choose between working and taking leisure. Holidays and weekends, though counted toward pay, are excluded from analysis.}

We revisit the dynamic behavioral model of \citet{DHR}, henceforth DHR. Let $t$ denote the day of the month, ranging from $1$ to $T=30$. On each day $t$, a teacher chooses between working ($j_t = 1$) and taking leisure ($j_t = 0$). On the final day of the month, the consumption utility is determined by the monthly paycheck
\begin{align}
\label{eq:total}
\pi(d_T) = 500 + 50 \cdot \max(d_T - 10, 0),
\end{align}
where $500$ is the base salary, $d_T$ is the total number of days worked by day the final day $T$, and $50$ is the bonus.   For all days $t < T$, there is no consumption; utility accumulates only through leisure and is modeled as
\begin{align}
u(x_t, 0) &= \bar{x}_t^{\prime} \mu_0. \label{eq:u0}
\end{align}
Here, $ \bar{x}_t \in \mathrm{R}^{p_X}$ is the state vector including a constant and possibly other observable characteristics, and $\mu_0  \in \mathrm{R}^{p_X}$ is a parameter to be estimated. The state vector is $X_t = (\bar{X}_t, d_t)$.   For example, in Model I of DHR, the leisure utility is assumed to be the same for all teachers, which corresponds to  $u(x_t, 0) = \mu_0$ and $X_t =d_t$.  The per-period utilities are not discounted. DHR includes only a handful of observables into  $\bar{x}_t$ so as to leverage standard maximum likelihood estimators.

We consider a stylized dynamic binary choice model as described in Section~\ref{sec:setup}. Since teachers cannot be fired, the base salary is assumed not to affect their choice between work and leisure.  We decompose the total monthly bonus into daily payments.  Specifically, we assume the utility of working ($j_t = 1$) is given by:
\begin{align}
u(d_t,1) &= 50 \cdot \mu_1 \cdot 1\{ d_t - 10 \geq 0 \}, \label{eq:u1}
\end{align}
where the indicator function, referred to as “In the money” by \citet{DHR}, captures the bonus structure in the stylized model.   The parameter $\mu_1$ converts monetary rewards (in Rupees) into utility units. The stylized model in equations \eqref{eq:u1}--\eqref{eq:u0} preserves the monetary incentives of the exact model \eqref{eq:total}--\eqref{eq:u0}, aside from discounting of the bonus. To abstract from the finite-horizon considerations, we restrict attention to calendar days 15, 16, and 17 of each month. If this model is relevant, 
the methods developed in this paper permit the state vector $\bar{X}_t$ to include high-dimensional covariates.
 

Table~\ref{table:welfare} compares the exact estimates reported by \citet{DHR} (Columns (1)-(2)) to our stylized infinite-horizon replications (Columns (3)-(4)) for selected coefficients. The results are encouraging. First, the estimated bonus coefficient under the stylized model falls within the range of the exact estimates. The standard errors of the replicated coefficients are, on average, 2.5 to 3 times larger, as expected given the smaller sample (only three days per month). Second, the coefficient on teacher test scores remains negative across all specifications, consistent with the original findings. Other coefficients (not reported here) also closely match their exact counterparts. These results suggest that the stylized infinite-horizon framework is appropriate for this dataset.

We further investigate the role of prior work history in explaining teachers work decisions, a possibility raised by \citet[footnote 16]{DHR}.
\cite{DHR} included the first lag of work history in Model VIII, and we  extend this idea by incorporating 89 more lags.  Additionally, we summarize work history using a \textit{work streak} variable, defined as the number of consecutive days a teacher has worked without taking leisure:
\begin{align}
a_{t+1} = (a_t + 1) \cdot 1\{ j_t = 1 \}, \quad a_0 = 0.\label{eq:uat}
\end{align}
We investigate the role of prior work history both in the predictive and structural settings.
  
Table \ref{table:mse} shows the out-of-sample mean squared error (MSEs) for predicting a teacher’s decision to work, using models that sequentially expand the covariate set from Model I to Model IV. Adding the teacher's test score on top of days worked yields minimal improvement. Including month dummies (Model II) results in a modest reduction in MSE. Incorporating the work streak (Model III) leads to a substantial improvement: relative to Model II, the MSE falls by roughly 32\% for Logit, 32\% for Probit, and 35\% for Random Forest. Finally, adding the full 90-day work history dummies (Model IV) further improves prediction, halving the MSE relative to the baseline. These findings show that prior work history has high predictive power even after other observables have been taken into account.  

As a next exercise, we revisit the stylized structural model with the aim of flexibly modeling the utility of leisure. 
As noted by \citet{DHR}, decisions to skip work may be influenced by factors such as social norms, informal requests and commitments, accumulated effort, or fatigue. 
These factors are unlikely to be fully captured by basic observables like test scores. 
A longer spell of prior work history may be more granular and thus may better represent teacher's heterogeneity even if the history itself has no structural or causal interpretation. 
Assuming only a small number of lags suffices to capture the history, we include a 90-day window of lagged attendance indicators. This sparsity assumption calls for the use of Lasso estimators of the value function and the dynamic dual representation, developed in Section \ref{sec:valuefunction}.

\begin{table}[H]
\centering
\caption{ Structural Estimates from  Exact (DHR) and Stylized (CNS) Models }
\begin{tabular}{lcc@{\hskip 24pt}cc}
\toprule
\midrule
 & \multicolumn{2}{c}{DHR} & \multicolumn{2}{c}{CNS} \\
\cmidrule(lr){2-3} \cmidrule(lr){4-5}
     & (1) & (2) & (3) & (4) \\
\midrule
Bonus ($\mu_1$) & 0.049 & 0.016 & 0.039 & 0.039 \\ 
      & (0.001) & (0.001) & (0.003) & (0.003) \\ 
Teacher Test Score  &  & -0.005 &  & -0.011 \\ 
                   &  & (0.002) &  & (0.005) \\ 
\bottomrule
\end{tabular}
\caption*{\textit{Notes:} 
Table reports selected coefficients for structural estimates. Columns (1)-(2) replicates exact, finite-horizon, estimates from Models I and VIII using the full sample of 57 teachers observed over 18 months where payoffs are not discounted.   Columns (3)--(4) report their analogs for the stylized model defined in equations~\eqref{eq:u1}--\eqref{eq:u0} with distinct choices of state components.  The stylized,  infinite-horizon, model is estimated using only the observations from the middle days of each month (15, 16, and 17) and the discount factor is $\beta=0.99$.  See main text for details.}
\label{table:welfare}
\end{table}
\begin{table}[H]
\caption{Out-of-Sample Mean Squared Error for Predicting Decision to Work}
\centering
\begin{tabular}{clrrr}
\toprule
\midrule
Model & Covariates (State Components) & Logit & Probit & RF  \\
 & & (1) & (2) & (3) \\
\midrule
%I   & Days Worked                    & 0.174 & 0.175 & 0.153 \\
I  &Days Worked  $+$ Test Score                     & 0.173 & 0.174 & 0.153 \\
II & $+$ Month Dummies                 & 0.171 & 0.171 & 0.147 \\
III  & $+$ Work Streak             & 0.116 & 0.116 & 0.095 \\
IV   & $+$ Work History Dummies & 0.062 & 0.063 & 0.069 \\
\bottomrule
\end{tabular}
\caption*{\textit{Notes}: 
Table reports  the Mean Squared Error (MSE) for predicting teacher decision to work, using models that progressively add explanatory variables. 
Row I includes only the number of days worked in the current month and the teachers test score, Row II adds month dummies, and so on. 
Models are trained on 2004 data and evaluated on 2005 observations. 
Columns correspond to Logistic Regression (Logit), Probit Regression (Probit), and Random Forest (RF). 
See Table \ref{table:ds} in Appendix for descriptive statistics of the variables.}
\label{table:mse}
\end{table}

Table~\ref{table:welfare2} reports selected coefficients for the structural parameter estimates (Panel A) as well as welfare metrics (Panels B, C and D). 
Columns (1)-(2) correspond to a simple model of \eqref{eq:u0} whose only observable is teacher's test score. 
 Columns (3)-(4) correspond to a more sophisticated model of \eqref{eq:u0} where observables include prior work history. 
 In both cases, the model is estimated using Algorithm \ref{alg:leebound2} described in Appendix G  based on Logit and Random Forest estimators of conditional choice probability. 
The welfare metrics are estimated using the dual estimator described in Section \ref{sec:estimation}. Instead of using the value function, it combines the structural estimates of Panel A with the choice probabilities.

Our findings are as follows. 
First and foremost,  the structural parameters—particularly the bonus coefficient ($\mu_1$) and the effect of teacher test score —are robust to the inclusion of the work history as a state component as well as to the choice of CCP  estimator. 
In contrast, the welfare metrics are more sensitive to model specification.  
In particular, failure to account for the prior work history results in overestimating  teacher welfare by 13-20$\%\%$. 
This overstatement persists across both the full sample and the subgroups defined by test scores values. 
Finally, the average welfare is lower for teachers with test scores at or below 30 than for those with scores above 40, as shown in Panels C and D, 
which is consistent with we include a 90-day window of lagged attendance indicators. 
Similar to DHR's interpretation, more skilled teachers are more committed to work and receive higher utility from teaching.

\begin{table}[h]
\centering
\caption{Welfare Estimates for Structural Model}
\label{table:welfare2}
\begin{tabular}{lcccc}
\toprule
\midrule
Model & \multicolumn{2}{c}{I} & \multicolumn{2}{c}{IV} \\
\cmidrule(lr){2-3} \cmidrule(lr){4-5}
CCP Estimators & Logit & RF & Logit  & RF \\
               & (1) & (2) & (3) & (4) \\
\multicolumn{5}{l}{\textit{Panel A: Structural Estimates}} \\
\midrule
Bonus ($\mu_1$)                                 & 0.039 & 0.039 & 0.049 & 0.046 \\ 
                                    & (0.003) & (0.003) & (0.004) & (0.005) \\ 
Teacher Test Score                                 & -0.011 & -0.011 & -0.012 & -0.013 \\
                                   & (0.005) & (0.005) & (0.009) & (0.010) \\ 
\midrule                                   
\multicolumn{5}{l}{\textit{Panel B: Average Welfare}} \\
\midrule
                                    & 183.232 & 187.446 & 158.784 & 158.444 \\ 
 & (2.33) & (2.401) & (5.126) & (5.222) \\
\midrule
\multicolumn{5}{l}{\textit{Panel C: Average Welfare for Low-Scored Teachers}} \\
\midrule
                                   & 176.183 & 185.578 & 151.138 & 154.877 \\ 
 & (3.749) & (4.082) & (9.383) & (9.664) \\ 
\midrule
\multicolumn{5}{l}{\textit{Panel D: Average Welfare for High-Scored Teachers}} \\
\midrule
                                   & 186.353 & 188.917 & 158.617 & 156.221 \\ 
 & (3.251) & (3.328) & (7.568) & (7.641) \\  
\bottomrule
\end{tabular}
\caption*{\textit{Notes:} 
Table reports  findings for the stylized model  defined in  equations~\eqref{eq:u1}--\eqref{eq:u0}. 
The choice of components  corresponds to Models I and IV in Table~\ref{table:mse}. 
Panel A reports selected coefficients of structural parameter estimated using Algorithm \ref{alg:leebound2} in the Appendix. 
Panels B, C and D report welfare metrics defined in Examples \ref{ex:average} and  \ref{ex:group}.  
Teachers are classified as \textit{Low-Scored} if their test score is less than or equal to 30, and \textit{High-Scored} if their score is greater than or equal to 40. 
See main text for further details.}
\end{table}

\section{Dynamic Dual and Doubly Robust Representations of Welfare Metrics}
\label{sec:dual}

\subsection{Dynamic Dual Representation}

In this Section, we give a dual representation of the parameter of interest.   This representation is important for several purposes.
When $w_0(X)$ depends only on $K$, and so is time-invariant, the dual representation gives a simplified formula for $\delta_0$ that does not require solving any dynamic problem.
Otherwise,  the dual representation leads to a doubly robust moment condition for identification and estimation of the parameter of interest. 
The dual representation\footnote{See equations (16)-(17) in the first version of the paper \cite{CNS}.  }   was derived in the previous version of this paper \citet*{CNS}.

A key part of the dual representation is a function of the state variable that is a backward discounted value of $w_0(X)$, given by
\begin{align}
\label{eq:alpha0}
\alpha_0(X) := \sum_{t \geq 0} \beta^t \E [ w_0(X_{-t}) \mid X],
\end{align}
where $X_{-t}$ is the state variable in period $-t$ in the extended stochastic process ${X_t}$ where $t$ ranges over all the integers. 
Alternatively, $\alpha_0(X)$ is a fixed point of the backward  dynamic operator 
\begin{align}
\label{eq:bdp}
w_0(X_{+})  + \beta \E[ \alpha_0(X) \mid X_{+} ] = \alpha_0(X_{+}).
\end{align}

The following result gives the dynamic dual representation of weighted average welfare. 

\begin{proposition}[Dynamic Dual Representation]
\label{lem:riesz}
Let $V_{\zeta}$ be a net present discounted value of per-period utility $\zeta(x)$ as in \eqref{eq:npv2} with $\zeta(\cdot)$ replacing $\zeta_0(\cdot)$. 
The function $\alpha_0(X)$ in equation \eqref{eq:alpha0} is the unique function such that 
\begin{align}
\label{eq:riesz1}
\E [w_0(X) V_{\zeta}(X)] = \E[ \alpha_0(X) \zeta(X)] \quad
\end{align}
 for any $\zeta(X)$ with finite second moment.
\end{proposition}

An interesting implication of this dual representation is that if $w_0(X)$ is time-invariant, then $\delta_0$ depends only on the per-period expected utility $\zeta_0(X)$. 

\begin{corollary}[Dynamic Dual Representation With Time-Invariant Weight]
\label{cor:riesztimeinvar}
If $m(Z,V)=w_0(X)V(X)$ and $w_0(X)=w_0(K)$ depends only on a time-invariant variable $K$ then $\alpha_0(X)=(1-\beta)^{-1}w_0(K)$ and
\begin{align}
\label{eq:dualtimeinvar}
\delta_0 = \E [w_0(X)V_{\zeta}(X)] = (1-\beta)^{-1}\E [w_0(K)\zeta(X)].
\end{align}
\end{corollary}

In each of our first three examples, the weight was time-invariant so that Corollary \ref{cor:riesztimeinvar} applies and the parameter of interest $\delta_0$ depends only on $\zeta_0(X)$. Here are expressions for $\delta_0$ for Examples \ref{ex:average}-\ref{ex:stock}. 

\begin{example}[continues=ex:average]
The average welfare is given by 
\begin{align}
\label{eq:alpha0average}
\delta_0=\E[V_0(X)]=(1-\beta)^{-1}\E[\zeta_0(X)]
\end{align}
\end{example}

\begin{example}[continues=ex:group]
The group average welfare is given by 
\begin{align}
\label{eq:dualtimeinvar2}
\delta_0 = (1-\beta)^{-1} \E[1(K=k)\zeta_0(X)]/\Pr(K=k).
\end{align}

\end{example}

\begin{example}[continues=ex:stock]
The policy effect is given by 
\begin{align}
\label{eq:stock2}
\delta_0 = (1-\beta)^{-1}\E[w_0(K)\zeta_0(X)], w_0(K) = [\pi^{*} (K) - \pi (K)]/\pi(K).
\end{align}

\end{example}

\begin{remark}[Implications for \cite{Rust} and \cite{HotzMiller}]
Consider a dynamic binary choice model where one of the actions has a terminal property similar to \cite{Rust}. Furthermore, suppose the deterministic utilities take a linear index form
 \begin{align}
 \label{eq:ux1}
 u(x,1)=D_1(x)^{\prime} \theta_{11}, \quad u(x,0) =D_0(x)^{\prime} \theta_{10},
 \end{align}
where $\theta_0= (\theta_{11}, \theta_{10})$. This parameter can be identified via a semiparametric moment condition whose only nuisance parameter is conditional choice probability \cite{HotzMiller,LRSP}.  Stacking this moment condition with \eqref{eq:alpha0average} in Example \ref{ex:average} gives a semiparametric moment condition for $(\theta_0, \delta_0)$ where neither value function nor any other fixed point of Bellman equation needs to be estimated.
\end{remark}

\begin{remark}[Implications for dynamic discrete choice models as in \cite{AMira2002}]
In a broad class of dynamic discrete choice models, including the one in \cite{DHR}, neither action has a terminal choice property. In that case, the structural parameter can be identified by a PMLE moment condition described in \cite{AMira2002} whose debiased analog is proposed in \cite{AdsmEck2022}. When the state space is continuously supported, the nuisance components includes choice-specific value functions that are fixed points of a nonparametric IV problem.  Appendix \ref{sec:dynamicbinary2} describes a related yet different moment condition that we utilize in Section \ref{sec:empirical}.
\end{remark}

When the weight is not time-invariant, $\alpha_0$ will not generally have a closed form or explicit expression because it depends in a complicated way on the dynamic distribution of the state vector $X_t$.
To help understand better the nature of $\alpha_0$ we revisit Example \ref{ex:avder} where the state variable follows an autoregressive process of order 1 with a Gaussian innovation. While this example may not correspond to a state distribution under a dynamic discrete choice model, we include it for pedagogic purposes to help explain the nature of $\alpha_0$.

\begin{proposition}[Dynamic Dual Representation for Average Marginal Effects]
\label{prop:avder}
Consider an AR(1) model with a Gaussian innovation
\begin{align}
\label{eq:arma}
S_{t+1} =  \rho(K) S_t + U_t, \quad U_t \sim \text{IID} N(0, 1),
\end{align} 
where $\rho(K) \in (-1,1) \text{ a.s.}$  is an autoregressive coefficient that may depend on $K$. Then the weighting function in Example \ref{ex:avder} is linear in  $S^2$
\begin{align}
\label{eq:linearweight}
w_0(X) = \gamma_1(K) S^2 + \gamma_0(K)
\end{align}
whose intercept $\gamma_0(K)$ and the slope $\gamma_1(K)$ are functions of the time-invariant type $K$ given in \eqref{eq:linearweightapp}.   The dynamic dual representation is
\begin{align}
\label{eq:linearweightalpha}
\alpha_0(X) =  \gamma_1(K) \dfrac{S^2}{1 - \beta \rho^2(K)}  +  (1-\beta)^{-1}  \gamma_1(K) \dfrac{\beta }{(1 - \beta \rho^2(K))} +   (1-\beta)^{-1} \gamma_0(K).
\end{align}
\end{proposition}

\begin{corollary}
Consider a white noise model with i.i.d states $S_t$, which is a special case of  \eqref{eq:arma} with  $\rho(K)=0$. Then the dynamic dual representation  is time-invariant
\begin{align*}
w_0(X) = w_0(K) = \gamma_0(K) =  - \partial_K \ln f_K(K), \qquad \alpha_0(X) = (1-\beta)^{-1} \gamma_0(K).
\end{align*}

\end{corollary}

\subsection{Doubly Robust Representation}
\label{sec:dr}
In this section, we give an identifying moment condition for the parameter of interest that is doubly robust in the sense that it holds if just one of $V(\cdot)$ or $\alpha(\cdot)$ is the true function. 
This moment condition uses the identifying conditional moment restriction for $V_0$ in equation \eqref{eq:fdp}.
Let $Z$ denote a data observation which includes $(X,X_{+})$, $V$ denote a possible value function, and $\lambda(Z, V):= \beta V(X_{+}) - V(X) + \zeta_0(X)$.
Equation \eqref{eq:fdp} is equivalent to the conditional moment restriction
\begin{align}
\label{eq:npiv}
\E[\lambda(Z, V_0) \mid X]=0. 
\end{align}
This is a nonparametric conditional moment restriction like those of \cite{NeweyPowell} and \cite{AiChen2003} where $X_+$ is an "endogenous" variable, $X$ is an "instrument", and  $\lambda(Z,V)$ is a nonparametric residual as considered in \cite{CNS} and \cite{chen2022wellposedness}. Here we take $\zeta_0(X)$ to be a known function and will consider estimation of $\zeta_0(X)$ in the next Section.

Let $\alpha$ denote a possible function $\alpha_0$. A doubly robust moment function can then be formed as 
 \begin{align}
\label{eq:drmom}
g(Z,V,\alpha,\delta) = m(Z,V)-\delta + \alpha(X)\lambda(Z,V).
\end{align}
Given a function $\xi$ of $X$ define
\begin{align}
\| \xi \| = (\E [\xi(X)^2])^{1/2}.
\end{align}
 \begin{lemma}[Double Robustness of Moment Function  \eqref{eq:drmom}]
\label{lem:orthod}
The moment function satisfies
\begin{align}
\label{eq:mainorthog}
\E [g(Z,V,\alpha, \delta_0)] = \E[(\alpha(X)-\alpha_0(X))(\lambda(Z,V)-\lambda(Z,V_0))].
\end{align}
Also
\begin{align}
\label{eq:doublerobustnessmain}
| \E [g(Z, V,\alpha, \delta_0)] | \leq (1+\beta) \| \alpha - \alpha_0 \| \| V - V_0 \|.
\end{align}
\end{lemma}

Lemma \ref{lem:orthod} establishes double robustness of the moment function $g(Z,V,\alpha,\delta)$ which has zero expectation at $\delta=\delta_0$ if either $V=V_0$ or $\alpha=\alpha_0$ by equation  (\ref{eq:mainorthog}). A doubly robust estimator of the average treatment effect was given in (\cite{Robins}) and (\cite{LRSP}) characterize doubly robust moment functions as being linear in both non-parameric components.

\begin{example}[continues=ex:avder]
The doubly robust representation for the average derivative is 
\begin{align}
\label{eq:alpha0drderivative}
\E [g(Z, V, \alpha,\delta_0)] &= \E [ \partial_K V(X) + \alpha(X)  (\beta V(X_{+}) - V(X) + \zeta_0(X))  - \delta_0 ]
\end{align}
where the true value of $\alpha$ is the backward discounted value \eqref{eq:alpha0} based on $w_0(X) = - \partial_k \ln f(K| S)$. \end{example}

\section{Overview of Estimation and Inference}
\label{sec:estimation}

In this Section, we give estimators of the welfare metrics we introduced in Sections \ref{sec:setup} and \ref{sec:oaxaca}. These estimators will account for the estimation of $\zeta_0(\cdot)$ and of $w_0(\cdot)$ or $m(Z,V)$ by including influence functions for their effect on identifying moments. The inclusion of these influence functions debiases for model selection and/or regularization in the estimation of unknown functions and corrects resulting standard errors for their estimation, as in \cite{LRSP}.

For simplicity of exposition, we focus on panels with $T=2$ time periods where the pairs of consequent states $(X_{i1},X_{i2})_{i=1}^n$ are i.i.d. We use standard cross-fitting for i.i.d data (\cite{schick1986asymptotically})
as common in work on debiased machine learning, \cite{chernozhukov2016double}.
For a weakly dependent time series with $T\geq 3$ periods, cross-fitting along both unit and time dimension is possible by leaving out neighboring folds, as discussed in \cite{CGST}. 
Related work on conditional moment restrictions with weak dependence includes \citet{ChenSieveRiesz,ChenLiao2015,ChenLiaoWang2024}.

\subsection{Average Welfare and Related Averages}

We will first consider a weighted average value function parameter with time-invariant weight that is possibly estimated. This case includes average welfare and related averages in Examples \ref{ex:average}--\ref{ex:stock}. Let $F$ denote an unrestricted distribution for $Z$ and $w(K,F)$ and $\zeta(X,F)$ denote the probability limit (plim) of an estimated weight $\widehat{w}(K)$ and an estimator $\widehat\zeta(X)$ respectively.
Let $\phi_{w}(Z)$ and  $\phi_{\zeta}(Z)$ be the influence functions of $(1-\beta)^{-1}\E[ w(K,F)\zeta_0(X)]$ and $(1-\beta)^{-1}E[w_0(K)\zeta(X,F)]$ respectively.
To nonparametrically debias for the estimation of $w(K)$ and $\zeta(X)$ and so construct a Neyman orthogonal moment function we add $\phi_{w}(Z)$ and $\phi_{\zeta}(Z)$ to the identifying moment function, as in \cite{LRSP}, to obtain
\begin{align}
\label{eq:timeinvarmoment}
\psi(Z, \gamma, \phi, \delta)  &= (1-\beta)^{-1}w(K)\zeta(X) - \delta +  \phi_w(Z) + \phi_\zeta(Z)\\
\gamma &=(w,\zeta), \phi=(\phi_w,\phi_\zeta)
\end{align}
where the true parameter $\delta_0$ solves
\begin{align}
\E [\psi(Z, \gamma_0, \phi_0, \delta_0)] = 0
\end{align}
at the true value $\gamma_0$ of $\gamma$ and $\phi_0$ of $\phi$.

For cross-fitting purposes, we partition the set of data indices ${1, \ldots, n}$ into $L$ disjoint subsets $I_\ell$ of about equal size, $\ell = 1, \ldots, L$.  
Let $\widehat\gamma_\ell=(\widehat w_{\ell}, \widehat \zeta_{\ell})$ and $\widehat\phi_\ell =(\widehat\phi_{w\ell},\widehat\phi_{\zeta\ell})$ be estimators of the weight, per-period utility, and influence functions constructed using all observations not in  $I_\ell$.
Also let $\psi(Z,\widehat\gamma_\ell,\widehat\phi_\ell,\delta)$ be as in equation \eqref{eq:timeinvarmoment} with $\widehat\gamma_\ell$ and $\widehat\phi_\ell$ in place of $\gamma$ and $\phi$.
A cross-fit estimator of $\delta_0$ can be obtained from solving 
$\sum_{\ell = 1}^L \sum_{i \in I_\ell} \psi(Z_i,\widehat\gamma_\ell,\widehat\phi_\ell,\delta)/n =0$ for $\delta$ giving 
\begin{align}
\widehat{\delta} &= \frac{1}{n} \sum_{\ell = 1}^L \sum_{i \in I_\ell}[(1-\beta)^{-1}\widehat{w}_\ell(K_i)\widehat\zeta_\ell(X_i) +  \widehat\phi_{w\ell}(Z_i) + \widehat\phi_{\zeta\ell}(Z_i)],
\label{eq:estimate} \\
\widehat{\Omega}&=\frac{1}{n} \sum_{\ell = 1}^L \sum_{i \in I_\ell} \psi^2(Z_i,\widehat\gamma_\ell,\widehat\phi_\ell,\widehat\delta). \label{eq:sterror}
\end{align}
An example of estimated per-period utility $\zeta$ and its correction term for dynamic binary choice is given in Appendix \ref{sec:dynamicbinary}. The Lemma \ref{lem:alinearity} in Appendix \ref{sec:atheory} provides sufficient conditions for the validity of asymptotic inference. 

\begin{example}[continues=ex:average]
The estimate of the average welfare is
\begin{align*}
\widehat \delta &= \frac{1}{n}\sum_{\ell = 1}^L \sum_{i \in I_\ell} [ (1-\beta)^{-1} \widehat{\zeta}_{\ell}(X_i)+ \widehat \phi_{\zeta\ell} (Z_i)]
\end{align*}

\end{example}

We next continue with the description of group average welfare in Example \ref{ex:group}. 
A key difference from Example \ref{ex:average} is that the time-invariant weighting function $w$ 
depends on the  group probability $\Pr (K=k)$ which needs to be estimated. 

\begin{example}[continues=ex:group]
Let $\phi_{\zeta}$ be the influence function of $\E[ (1-\beta)^{-1} 1\{ K=k\} \zeta(X,F)/\Pr (K=k)]$.  The weighting function $w(K) = 1\{K=k\}/\Pr (K=k)$. The influence function for $\Pr (K=k)$ is 
\begin{align*}
\phi_w(Z) =  -\dfrac{ \delta_0}{\Pr(K=k)}(1\{K=k\} - \Pr (K=k)).
\end{align*}
Thus, the estimator of $\delta_0$ reduces to
\begin{align*}
\widehat \delta &=  \frac{1}{n}\sum_{\ell = 1}^L \sum_{i \in I_\ell} [[ (1-\beta)^{-1}  1\{ K_i = k\}/ \widehat \Pr (K=k) ]\widehat{\zeta}_{\ell}(X_i)+\widehat{\phi}_{\zeta\ell}(Z_i)+\widehat{\phi}_{w\ell}(Z_i)],
\end{align*}
whose standard error in \eqref{eq:sterror} accounts for estimation of $\Pr (K=k)$ by including the term $\phi_w(\cdot)$.
\end{example}

\subsection{General case.} In this Section we descrite the estimator of $\delta_0=\E[m(Z,V_0)]$ when $w_0(\cdot)$ varies with time.  Let $m(Z,V,F)$ denote the plim of the estimated $m(Z,V)$ function, $\phi_m(Z)$ the influence function of $\E[m(Z,V_0,F)]$, and $\phi_{\zeta}(Z)$ the influence function of $\E[\alpha_0(X)\zeta(X,F)]$. The orthogonal moment function is
\begin{align}
\label{eq:timevarmoment}
\psi(Z, \gamma, \phi, \delta)  &= m(Z,V) - \delta + \alpha(X)
(\beta V(X_{+}) - V(X) + \zeta(X))
+ \phi_m(Z) + \phi_\zeta(Z), \\
\gamma &=(m,V,\zeta), \phi=(\alpha,\phi_m,\phi_\zeta)
\end{align}
where $\phi_m$ corrects for the estimation of $m(\cdot)$ and $\phi_{\zeta}$ corrects for the estimation of $\zeta$.  
Algorithm \ref{alg:leebound} below gives the proposed estimator of the parameter of interest.  
Lemma \ref{lem:alinearity2} in Appendix \ref{sec:atheory} establishes the validity of asymptotic inference.

\begin{example}[continues=ex:avder]
The estimate of the average derivative is
\begin{align}
\label{eq:alpha0average3}
\widehat \delta = \frac{1}{n}\sum_{\ell = 1}^L \sum_{i \in I_\ell}  [\partial_k \widehat V_{\ell} (X_i) + \widehat \alpha_{\ell}(X_i)(\beta\widehat V_{\ell} (X_{+i})-\widehat V_{\ell} (X_i)+\widehat \zeta_{\ell}(X_i)) +\widehat{\phi}_{\zeta\ell}(Z_i)]. 
\end{align}
\end{example}
Here there is no correction $\phi_m$ since the functional $m(Z,V)=\partial_K V(X)$ of $V$ does not involve any unknown components.

\begin{example}[Counterfactual Welfare]
\label{ex:oaxaca}
The counterfactual welfare $ \delta_{\langle 1 \mid 0 \rangle}$ in equation \eqref{eq:counterfact} is a special case of \eqref{eq:linfunc} with $m(z,V)$ in \eqref{eq:mzvoaxaca} and 
 $w_0$ in  \eqref{eq:compositionweight}. Let $\phi_{\zeta}(Z)$ be the first step influence function (FSIF) of $\E[ \alpha_0(X) \zeta(X,F) ]$ where $\alpha_0$ is given in \eqref{eq:alpha0} based on the weighting function $w_0$ in \eqref{eq:compositionweight}.  The influence function for $\Pr (K=0)$ is
$$
\phi_w(Z) = - \dfrac{\delta_{\langle 1 \mid 0  \rangle}}{\Pr (K=0)} \left (1\{K=0\} -\Pr (K=0)\right).
$$
Thus the estimator of $\delta_{ 1 \mid 0}$ reduces to 
$$
\widehat \delta_{\langle 1 \mid 0  \rangle}=\frac{1}{n}\sum_{\ell = 1}^L \sum_{i \in I_\ell} \dfrac{1\{ K_i = 0 \}}{\widehat \Pr (K=0)} \widehat V_{\ell}(S_i, 1) + \widehat \alpha_{\ell}(X_i) \lambda(Z_i, \widehat V_{\ell}) +\widehat{\phi}_{\zeta\ell}(Z_i)+\widehat{\phi}_{w\ell}(Z_i),
$$
whose standard error in \eqref{eq:sterror} accounts for estimation of $\Pr (K=0)$ by including the term $\phi_w(\cdot)$. 
\end{example}

Algorithm \ref{alg:leebound} summarizes the estimation steps of welfare metrics. 
 \begin{algorithm}
\small
\begin{algorithmic}[1]
	\STATE  Partition the set of data indices $\{1,2,\dots, n\}$ into  $L$ disjoint subsets of about equal size with where $L$ is an odd number $L \geq 3$.

	\STATE  Estimate the value function  by minimizing sample cross-fit criterion function 
\[
\widehat{V}_{\ell} = \arg\min_{V} L^V_n (V)
\]

\STATE Estimate the dynamic dual representation by minimizing sample cross-fit criterion function 
\[
\widehat{\alpha}_{\ell} = \arg\min_{\alpha}  L^{\alpha}_n (\alpha)
\]

\STATE Estimate the per-period reward $\zeta(\cdot)$ and correction terms $\widehat  \phi_{m\ell}, \widehat \phi_{\zeta\ell} $ using \textit{ all } observations in $I^c_{\ell}$

\STATE Estimate the target welfare metric $\delta_0$

\begin{align*}
\widehat{\delta} &= \frac{1}{n} \sum_{\ell = 1}^L \sum_{i \in I_\ell} [m(Z_i, \widehat V_{\ell} ) + \widehat \alpha_{\ell}(X_i)(\beta\widehat V_{\ell} (X_{+i})-\widehat V_{\ell} (X_i)+\widehat \zeta_{\ell}(X_i))  + \widehat  \phi_{m\ell}(Z_i) + \widehat \phi_{\zeta\ell}(Z_i)],
\end{align*}
and its standard error  $\sqrt{ \widehat{\Omega} /n}$ with $\psi$ given in \eqref{eq:timevarmoment} as
\begin{align*}
\widehat{\Omega}&=\frac{1}{n} \sum_{\ell = 1}^L \sum_{i \in I_\ell} \psi^2(Z_i,\widehat\gamma_\ell,\widehat\phi_\ell,\widehat\delta).
\end{align*}

\end{algorithmic}
\caption{
Step 1 partitions the data into $L$ folds to enable sample splitting. 
Steps 2 and 3 estimate the value function and the dynamic dual representation on each fold by minimizing cross-fit loss functions. 
Step 4 estimates the per-period utility $\zeta(\cdot)$ and the bias-correction terms using observations not in fold $\ell$. 
Step 5 computes the welfare estimate $\widehat{\delta}$ as a bias-corrected plug-in estimator. Step 6 computes the standard error of $\widehat{\delta}$ using the empirical variance of the influence function $\psi$, defined in equation~\eqref{eq:timevarmoment}.}
\label{alg:leebound}
\end{algorithm}

\section{Estimation of Value Function and Dynamic Dual Representation}
\label{sec:valuefunction}

This section introduces novel least squares estimators for both the value function and the dynamic dual representation.
In contrast to the welfare metrics, defined in Section \ref{sec:setup}, these objects do not require strict stationarity to be well-defined. 
Thus, the estimators of the value function and the dynamic dual representation delivered here do not require the time series to be strictly stationary.
Consequently, these estimators apply to any fixed point of a second-kind integral operator.

Sections \ref{ext1} and \ref{sec:lasso} develop a new least squares criterion for the value function that accommodates high-dimensional covariates through penalization, enabling consistent estimation in a high-dimensional state space. Section \ref{ext2} proposes a distinct least squares criterion for the dynamic dual representation, which depends only on the welfare metric of interest. This innovation permits automatic debiasing in the style of \citet{chernozhukov2021automatic} and \citet{chernozhukov2024qm}. Section \ref{sec:discussion} discusses the results.

\subsection{Least Squares Criterion for Value Function}
\label{ext1}

The starting point of our analysis is the expectation operator  $\bbA_0$ defined as
 \begin{align}
\label{eq:operatorlin}
(\bbA_0 \phi) (x) := \beta \E [ \phi (X_{+}) \mid X=x]
\end{align}
Rewriting  \eqref{eq:npv2}  in terms of $\bbA_0$ gives
\begin{align}
\label{eq:vee}
(\bbI - \bbA_0) V_0  = \zeta_0,
\end{align}
or, equivalently, $V_0 = (\bbI - \bbA_0)^{-1} \zeta_0$. The operator $\bbA_0$ is akin to the integral equation operator in \cite{SLinton}. 

The value function can be represented as a minimizer of a criterion function that depends on $\bbA_0$. $V_0$ will minimize the expected squared difference of the left and right-hand sides of equation \eqref{eq:vee}, that is
\begin{align}
V_0&= \arg \min_{V}  \E [(((\bbI - \bbA_0) V)(X)- \zeta_0(X))^2]
\label{eq:vee1} \\
&=\arg \min_{V}  \E [((\bbI - \bbA_0) V)(X)^2- 2((\bbI - \bbA_0) V)(X)\zeta_0(X)]
\label{eq:vee2} \\
&=\arg \min_{V}  \E [((\bbI - \bbA_0) V)(X)^2- 2(V(X)-\beta V(X_+))(X)\zeta_0(X)]
\label{eq:vee3} \\
&=\arg \min_{V}  \E [(V(X)-\beta V(X_+))((\bbI - \bbA_0) V)(X)- 2\zeta_0(X))], \label{eq:vee4}
\end{align}
where the second equality follows by squaring and dropping the term that does not depend on $V$ and the third and fourth equalities by iterated expectations.
The expression minimized following the first equality is the nonparametric two-stage least squares criterion for \cite{NeweyPowell}, \cite{Newey1991}, and \cite{AiChen2003}.
The expression following the third equality is a hybrid that uses iterated expectations to remove the conditional expectation $\bbA_0$ from all but one term. 

\begin{proposition}[Least Squares Criterion for Value Function]
\label{prop:vee}
The  value function $V_0$ in \eqref{eq:npv2} is the unique minimizer of  least squares criterion functions
\begin{align}
V_0&= \arg \min_{V}  \E [((\bbI - \bbA_0) V)(X)^2- 2(V(X)-\beta V(X_+))(X)\zeta_0(X)] \label{eq:ellv2}  \\
&= \arg \min_{V}  \E [(V(X)-\beta V(X_+))((\bbI - \bbA_0) V)(X)- 2\zeta_0(X))]. \label{eq:ellv} 
\end{align}
\end{proposition}

Proposition \ref{prop:vee} gives two least squares criterion functions. We use \eqref{eq:ellv2} and \eqref{eq:ellv} to construct a Lasso and a Neural Network estimator, respectively. We describe the Lasso estimator in Section \ref{sec:lasso} and Neural Network estimator in Appendix D.

To describe the Lasso estimator of the value function let $$b(x) = (b_1(x), \dots, b_p(x)) \in \mathrm{R}^p,$$ be a vector of basis functions. We approximate the value function using a linear form 
$$
V(x) \approx \sum_{j=1}^{p} b_j(x) \rho_{Vj} =  b(x)^{\prime} \rho_V,
$$
where $\rho_V = (\rho_{V1}, \rho_{V2}, \dots, \rho_{Vp}) \in \mathrm{R}^p$ is a  $p$-vector of coefficients. 
The vector is chosen to minimize an approximate least squares criterion
$$
\rho_V =  \arg \min_{\rho \in \mathrm{R}^p} \rho^{\prime} G^V \rho - 2 M^V \rho
$$
where $G^V$ is a symmetric $p \times p$ matrix
\begin{align}
\label{eq:gveemvee}
G^V &= \E [((\bbI - \bbA_0) b)(X) ((\bbI - \bbA_0) b)^{\prime}(X)]
\end{align}
and $M^V$ is a linear term
\begin{align*}
M^V &= \E [(b(X) - \beta b(X_{+})) \zeta_0(X)].
\end{align*}
The FOC reduces to
$$
G^V \rho_V =  M^V.
$$
We choose the criterion \eqref{eq:ellv2} as opposed to \eqref{eq:ellv} so that the sample version of matrix $G^V$ is symmetric and positive-definite.

Given an i.i.d sample  $(X_i, X_{i+})_{i=1}^n$,  we construct a sample estimate of $\rho_V$ in the regime where $\dim (\rho_V) = p_V \gg n$.  For simplicity of exposition, we abstract away from subsequent estimation steps and drop respective cross-fitting indices.  Given a plug-in estimate $\widehat \bbA  b$ of $\bbA_0 b$ and $\widehat \zeta$ and $\zeta_0$ estimated on a hold-out sample, define
\begin{align*}
\widehat G^V &= n^{-1} \sum_{i=1}^n  (b(X_i) - (\widehat \bbA  b) (X_i)) ( b(X_i) - (\widehat \bbA  b) (X_i))^{\prime}, \\
 \widehat M^V  &=  n^{-1} \sum_{i=1}^n  (b(X_i) - \beta b (X_{i+}))  \widehat \zeta (X_i)
\end{align*}
Given a radius $\rho_V$, an $\ell_1$-regularized estimator of the value function takes the form
\begin{align}
\widehat{V}(x) &= b(x)^{\prime} \widehat{\rho}_V, \\
\widehat{\rho}_V &= \arg \min_{\rho \in \mathrm{R}^p}  \rho^{\prime} \widehat G^V   \rho -2 \widehat M^V \rho + r_V \| \rho \|_1.
\end{align}

\newpage

\subsection{Mean Square Convergence for Lasso Estimator}
\label{sec:lasso} 

Assumption \ref{ass:approx} requires that $V_0$ belongs to the mean square closure  $\Gamma$ of linear combinations $b(x)^{\prime} \bar{\rho}$, as well as that the approximating coefficients $\bar{\rho}$ are sufficiently sparse.  

\begin{assumption}
\label{ass:approx}
(1) There exist constants $C>1$ and  $\xi_V>1/2$ and $d_V \in (0, 1/2)$  such that  for each positive integer $s_V\leq Cn^{-2(d_V)/(2\xi_V+1)}$ there is $ \bar{\rho}$ with $s_V$ nonzero elements such that
\begin{equation*}
\mathrm{E}[ (V_0(X)- b^{\prime}(X) \bar{\rho} )^{2}]\leq Cs_{V}^{-2\xi_V}.
\end{equation*}
(2) The matrix $\E [ b(X) b(X)^{\prime}]$ is positive definite with eigenvalues bounded from above by $\bar{\lambda}$ and below by $\underline{\lambda}>0$.
(3) $\sup_j |b_j(X)|$ are bounded a.s.
(4) The radius $r_V$ is chosen such that  $\varepsilon_{n}=o(r_V),$ $r_V=o(n^{c}\varepsilon_{n})$ for all $c>0$, and there exists $C>0$ such that $p_V\leq Cn^{C}.$
(5)  The first-stage estimators  $\widehat \zeta$ of $\zeta_0$ and $\widehat \bbA b$ of $\bbA_0 b$ converge as $\| \widehat \zeta - \zeta_0 \| = o_P (\zeta_n)$ and 
$\| \widehat \bbA - \bbA_0 \| = o_P(a_n)$ where $\zeta_n + a_n = o (n^{-d_{V}})$ for some positive constant $d_{V} \in (0, 1/2)$.  
\end{assumption}

Assumption \ref{ass:approx}(1) requires value function to be approximately sparse in the chosen basis. Assumptions \ref{ass:approx}(2)-(4) are standard regularity conditions. Assumption \ref{ass:approx}(5) reduces to a rate condition on the first-stage estimators. 

\begin{theorem}[Mean Square Rate for Lasso estimator of Value Function]
\label{lem:lassovalue}
If  Assumption \ref{ass:approx} holds,  then, for any $c>0$, 
\begin{equation}
\label{eq:lassorate}
\Vert \widehat V- V_0\Vert =o_{p}(n^{c} n^{-2 d_{V} \xi_V/(2\xi_V+1)}).
\end{equation}
\end{theorem}

Theorem \ref{lem:lassovalue} gives a mean square convergence rate for the value function. The rate is determined by the sparsity parameter $\xi_V$ of the value function and the first-stage rate parameter $d_V \in (0, 1/2)$. 

\begin{remark}[Verification of Rate Condition (5) in Assumption \ref{ass:approx}]
To give an example of operator $\bbA_0$, we consider the dynamic discrete choice problem discussed in Section \ref{sec:setup} where the state transition is deterministic conditional on action. Then the estimator of $\bbA_0$ reduces to an estimator of choice probabilities, and the rate $a_n = O (p_n)$ where $p_n$ is the $\ell_{\infty}$-rate for CCPs.
\end{remark}

\begin{remark}[Lasso estimator of $\bbA_0$]
The following example of $\bbA_0$ is based on a Lasso estimator. For $j=1,2, \dots, p$, define the estimators as
\begin{align}
\widehat G  &= n^{-1} \sum_{i=1}^n b(X_i) b^{\prime}(X_i), \quad \widehat M_j =  n^{-1} \sum_{i=1}^n b(X_i) b_j(X_{+i}). \\
\widehat \rho_j & = \arg \min_{\rho} \rho^{\prime} \widehat G \rho - 2 \widehat M_j  \rho + r_j \| \rho \|_1
\end{align}
where
\begin{align*}
(\widehat \bbA b)_j = b(x)^{\prime} \widehat \rho_j, \quad j=1,2,\dots, p.
\end{align*}
Then, the rate condition (5) in Assumption \ref{ass:approx} on $a_n$ can be verified by the first-case rate of $\sup_{ 1 \leq j \leq p }\| \widehat \rho_j - \rho_j \|$ and can be established using the tools of \cite{CGST}.
\end{remark}

\subsection{Least Squares Criterion for Dynamic Dual Representation}
\label{ext2}

In this Section, we derive dynamic dual criterion function. Define
\begin{align}
\label{eq:operatorlindual}
(\bbA_0^{*} \phi) (x) := \beta \E[ \phi (X_{-}) \mid X=x]
\end{align}
From the dual representation of  \eqref{eq:bdp} we know that $\alpha_0$ satisfies 
\begin{align}
\label{eq:alp}
(\bbI - \bbA_0^{*}) \alpha_0  = w_0
\end{align}
and $\bbI - \bbA^{*}_0$ is invertible.

We show that $\alpha_0$ can be represented as a minimizer of a criterion function that depends on $\bbA^{*}_0$.  $\alpha_0$ will minimize the expected squared difference of the left and right-hand sides of equation \eqref{eq:alp}, that is
\begin{align}
\alpha_0&= \arg \min_{\alpha}  \E [(((\bbI - \bbA^{*}_0) \alpha)(X)- w_0(X))^2]
\label{eq:alpha1} \\
&=\arg \min_{\alpha}  \E [((\bbI - \bbA^{*}_0) \alpha)(X)^2- 2((\bbI - \bbA^{*}_0) \alpha)(X)w_0(X)]
\nonumber \\
&=\arg \min_{\alpha}  \E [((\bbI - \bbA^{*}_0) \alpha)(X)^2- 2m(Z,(\bbI - \bbA^{*}_0)\alpha)]
\label{eq:alpha2}  \\
&=\arg \min_{\alpha}  \E [(\alpha(X)-\beta \alpha(X_-))((\bbI - \bbA^{*}_0) \alpha)(X)- 2m(Z,(\bbI - \bbA^{*}_0)\alpha)],
\end{align}
where the second equality follows by squaring and dropping the term that does not depend on $\alpha$, the third equality follows the Riesz representation in \eqref{eq:rieszweight}, and the third equality by iterated expectations. 

\begin{proposition}[Least Squares Criterion for Dynamic Dual Representation]
\label{prop:alpha}
The dynamic dual representation $\alpha_0$ in \eqref{eq:bdp}  is the unique minimizer of the quadratic criterion function  
\begin{align}
\alpha_0 &=\arg \min_{\alpha}  \E [((\bbI - \bbA^{*}_0) \alpha)(X)^2- 2m(Z,(\bbI - \bbA^{*}_0)\alpha)]  \label{eq:alpha3} \\
&=\arg \min_{\alpha}  \E [((\bbI - \bbA^{*}_0) \alpha)((\bbI - \bbA^{*}_0) \alpha)(X)- 2m(Z,(\bbI - \bbA^{*}_0)\alpha)]. \label{eq:alpha6} \\
&=\arg \min_{\alpha}  \E [(\alpha(X)-\beta \alpha(X_-))((\bbI - \bbA^{*}_0) \alpha)(X)- 2m(Z,(\bbI - \bbA^{*}_0)\alpha)]. \label{eq:alpha4} 
%&=: \arg \min_{\alpha}  \E [\ell_{\alpha}(Z, m, \bbA^{*}_0) ]  \label{eq:ellalpha}
\end{align}
\end{proposition}

Given a vector of basis functions $$b(x) = (b_1(x), b_2(x), \dots, b_j(x), \dots, b_p(x)) \in \mathrm{R}^p,$$ where $p$ can differ from the one Section \ref{ext1}, we approximate the dynamic dual representation via a linear form
 $$
 \alpha(x) \approx \sum_{j=1}^{p} b_j(x) \rho_{\alpha j} = b(x)^{\prime} \rho_{\alpha}
 $$
where $\rho_{\alpha} = (\rho_{\alpha 1}, \rho_{\alpha 1}, \dots, \rho_{\alpha p})  \in \mathrm{R}^p$ is a  $p$-vector. The vector is chosen to minimize an approximate least squares criterion
$$
\rho_{\alpha} =  \arg \min_{\rho \in \mathrm{R}^p} \rho^{\prime} G_{\alpha} \rho - 2 M_{\alpha} \rho
$$
where the $p \times p$ matrix is an outer product of 
\begin{align*}
G_{\alpha} &= \E [((\bbI - \bbA^{*}_0) b)(X) ((\bbI - \bbA^{*}_0) b)^{\prime}(X)]
\end{align*}
and the free term is 
\begin{align*}
M_{\alpha} &= \E [m(Z,(\bbI - \bbA^{*}_0)\alpha) ]
\end{align*}
We choose the criterion \eqref{eq:alpha6} as opposed to \eqref{eq:alpha4} so that the sample version of matrix $G_{\alpha}$ is symmetric and positive-definite. Given a radius $r_{\alpha}$, an $\ell_1$-regularized minimum distance estimator of the dynamic dual representation
\begin{align}
\widehat \alpha(x) &=  b(x)^{\prime} \widehat{\rho}_{\alpha} \\
\widehat{\rho}_{\alpha} &= \arg \min_{\rho \in \mathrm{R}^p} \rho^{\prime} \widehat G^{\alpha}   \rho - 2 \widehat M^{\alpha} \rho + r_{\alpha} \| \rho \|_1
\end{align}
where 
\begin{align}
\label{eq:galpha}
 \widehat G^{\alpha}  &= n^{-1} \sum_{i=1}^n  (\bbI - \widehat{\bbA}^{*} ) b (X_i) (\bbI - \widehat{\bbA}^{*} ) b^{\prime} (X_i), \qquad \widehat M^{\alpha} = n^{-1} \sum_{i=1}^n   m(Z_i, (\bbI - \widehat{\bbA}^{*} ) b ). \nonumber
\end{align}

\begin{theorem}[Mean Square Rate for Lasso estimator of Dynamic Dual Representation]
\label{lem:lassovalue2}
Suppose Assumption \ref{ass:approx} holds with $\alpha_0$ in place of $V_0$ and $\xi_{\alpha}$ in place of $\xi_V$, and  suppose $\xi_{\alpha}>1/2$ and $a^{*}_n = o(n^{-d_{\alpha}})$. Then
\begin{equation}
\label{eq:lassorate2}
\Vert\widehat \alpha- \alpha_0\Vert=o_{p}(n^{c}n^{ (-2d_{\alpha}\xi_{\alpha})/(2\xi_{\alpha}+1)}).
\end{equation}
\end{theorem}

Theorem \ref{lem:lassovalue2} gives a mean square convergence rate for dynamic dual representation. The rate is determined by the sparsity parameter $\xi_{\alpha}$ of the dynamic dual representation and the first-stage rate parameter $d_{\alpha}$. The estimator is automatic in the sense that it only requires knowledge of the linear functional $m(Z, (\bbI - \bbA^{*})b)$.

\subsection{Discussion and Related Results}
\label{sec:discussion}
In this Section, we discuss related results. Remark \ref{rm:cross} discusses cross-fitting. Remark \ref{rm:nn} introduces a neural network estimator of the value function.  Remark \ref{rm:nn2} introduces a neural network estimator of the dynamic dual representation.
Remark \ref{rm:autom} describes the automatic property of the dynamic dual criterion function. Remark \ref{rm:rate} verifies rate conditions for asymptotic theory. 
%Remark \ref{rm:dantzig} sketches a Dantzig estimator of the value function.

\begin{remark}[Cross-fitting]
\label{rm:cross}
To ensure that the nuisance components $\zeta_0, \bbA_0, \bbA^{*}_0$ and their respective criterion functions are estimated on different samples,  the standard cross-fitting procedure is modified as follows. Let $L \geq 3$ be the number of partitions. For each partition  $\ell \in \{1,2,\dots, L\}$, let  $I^c_{\ell} = (Z_i)_{i \notin I_{\ell}}$ denote the set observations \textit{not} in $I_{\ell}$.   Partition  $I^c_{\ell} = I^{c1}_{\ell} \sqcup I^{c2}_{\ell}$ into two halves. For each nuisance parameter $  \gamma \in \{ \zeta, \bbA, \bbA^{*} \}$, let $\widehat \gamma^{1}_{\ell}, \gamma^{2}_{\ell}$ denote the estimator computed on $I^{c1}_{\ell}$ and $ I^{c2}_{\ell}$, respectively. A cross-fit criterion for the value function is 
\begin{align*}
L^V_n (V) = \sum_{i \in I^{c1}_\ell} \ell_V (Z_i, V, \widehat\omega^{2}_{\ell} ) + \sum_{i \in I^{c2}_\ell} \ell_V (Z_i, V, \widehat\omega^{1}_{\ell}), \quad \omega = (\zeta, \bbA).
\end{align*}
In a special case of Lasso estimator, the criterion $L^V_n (V)$ reduces to
\begin{align*}
L^V_n (\rho) &= \rho^{\prime} \widehat G^V \rho - 2 \widehat M^V \rho + r_V \| \rho \|_1
\end{align*}
where
\begin{align*}
\widehat G^V &= \sum_{i \in I^{c1}_\ell} (\bbI - \widehat \bbA^{2}_{\ell}) b (X_i) (\bbI - \widehat \bbA^{2}_{\ell}) b^{\prime} (X_i)   + \sum_{i \in I^{c2}_\ell} (\bbI - \widehat \bbA^{1}_{\ell}) b (X_i) (\bbI - \widehat \bbA^{1}_{\ell}) b^{\prime} (X_i) \\
 \widehat M^V &= \sum_{i \in I^{c1}_\ell} (b (X_i) - \beta b(X_{i+}))  \widehat \zeta^{2}_{\ell} (X_i) + \sum_{i \in I^{c2}_\ell} (b (X_i) - \beta b(X_{i+}))   \widehat \zeta^{1}_{\ell} (X_i).
\end{align*}
\end{remark}

\begin{remark}[Neural network estimator of value function]
\label{rm:nn}
Proposition \ref{prop:vee} facilitates a general plug-in estimator of the value function with an arbitrary function class.   Given a first-stage estimator of the per-period utility $\widehat \zeta$ of $\zeta_0$ and the  expectation operator  $\widehat \bbA$ of $\bbA_0$,   define 
\begin{equation}
\widehat{V}:=\arg\min_{V \in\mathcal{V}_{n}} n^{-1} \sum_{i=1}^n   \ell_V (Z_i, V, \widehat \zeta, \widehat \bbA),  \label{eq:veehat}
\end{equation} 
where $\ell_V(Z, V,  \zeta_0, \bbA_0)$ is taken to as in \eqref{eq:vee4}.  Theorem \ref{thm1} in Appendix \ref{sec:firststage} establishes mean square consistency of $\widehat{V}$ to $V_0$ for an arbitrary function class. Corollary \ref{cor1} establishes mean square consistency of the neural network estimator of $V_0$. 
\end{remark}

\begin{remark}[Neural network estimator of dynamic dual representation]
\label{rm:nn2}
Proposition \ref{prop:alpha} facilitates a general plug-in estimator of dynamic dual representation with an arbitrary function class.  Given a first-stage estimator of the operator $\widehat {\bbA}^{*}$ of $\bbA^{*}_0$,   define 
\begin{equation}
\widehat{\alpha}:=\arg\min_{\alpha \in\mathcal{A}_{n}} n^{-1} \sum_{i=1}^n   \ell_{\alpha} (Z_i, \alpha,  \widehat \bbA^{*}).  \label{eq:alphahat}
\end{equation}
Theorem \ref{thm2} in Appendix \ref{sec:firststage} establishes mean square consistency of $\widehat{\alpha}$ to $\alpha_0$ for an arbitrary function class. It can be used to derive 
mean square consistency of neural network estimator of $\alpha_0$. 
\end{remark}

\begin{remark}[Automatic property of criterion function \eqref{eq:alpha4}]
\label{rm:autom}
The criterion function \eqref{eq:alpha4} have  a convenient property that it only depends on the parameter of interest through $m(Z, (\bbI - \bbA_0) \alpha)$ and does not require an explicit formula for the weighting function $w_0$. A similar property has been establishes for the results in \cite{chernozhukov2021automatic} and \cite{chernozhukov2024qm}. When the model is static, that is $\beta=0$, Theorem \ref{lem:lassovalue2} recovers Theorem 1 of \cite{chernozhukov2021automatic}. Likewise, the criterion function \eqref{eq:alpha4} reduces to
$$
\alpha_0 =\arg \min_{\alpha}  \E [\alpha^2(X)- 2m(Z,\alpha)]
$$
proposed in \cite{chernozhukov2024qm}.
\end{remark}

\begin{remark}[Verification of rate conditions for asymptotic theory.]
\label{rm:rate}
Suppose the conditions of Theorems \ref{lem:lassovalue} and \ref{lem:lassovalue2} hold with $\xi_V, \xi_{\alpha}$ and $d_V, d_{\alpha}$. Furthermore, suppose
$$
2(d_{\alpha}\xi_{\alpha})/(2\xi_{\alpha}+1) + 2(d_{V}\xi_{V})/(2\xi_{V}+1) > 1/2.
$$
For example, if $\min (\xi_{\alpha}, \xi_V) > 1 + 0.01$ and $\min (d_V, d_{\alpha}) \in (3/8, 1/2)$, the product of mean square rates can be upper bounded as
$$
n^{1/2} \| \widehat V_{\ell} - V_0 \|   \| \alpha_{\ell} - \alpha_0 \|  = o_P (n^{1/2} n^{-1/4} \cdot n^{-1/4}) = o_P(1)
$$
which suffices for the product rate condition
$$
n^{1/2} \| \widehat V_{\ell} - V_0 \|   \| \alpha_{\ell} - \alpha_0 \| = o_P(1)
$$
of Assumption \ref{ass:ratealpha} in Appendix \ref{sec:atheory}.
\end{remark}

%\begin{remark}[Dantzig estimator of the value function]
%\label{rm:dantzig}
%Earlier work has focused on an instrumental variable (IV) approach corresponding to a  minimum distance problem
%\begin{align}
%\label{eq:bjuncond}
%&\E [ b(X) (b (X) -\beta  b(X_{+}))^{\prime} ] \rho_V -  \E [ b(X) \zeta_0(X) ] \\
%&= G \rho_V - M = 0. \nonumber
%\end{align}
%Unlike $G_V$  in \eqref{eq:gveemvee},   the matrix $G=\E [ b(X) (b (X)-\beta  b(X_{+}))^{\prime} ]$ is not symmetric and does not correspond to a FOC of a least squares criterion function. Yet, a sparsity restriction of $\rho_V$ can be imposed using a Dantzig-type estimator
%\begin{align}
%&\min_{\rho \in \mathrm{R}^p}  \| \rho \|_1 \text{    subject to   }  \| \widehat G \rho - \widehat M \|_{\infty} \leq \lambda_n, 
%\end{align}
%where $\lambda_n$ is an appropriately chosen sequence.
%\end{remark}

\section{Conclusions}
\label{sec:conclusion}

In this paper we introduce welfare metrics -- including welfare decompositions into direct and indirect effects -- and give a complete set of estimation and inference results for them in the presence of high-dimensional state space.
The results are presented for the dynamic binary choice model of \cite{Rust,HotzMiller,AMira2002} but are applicable to other dynamic models (e.g., dynamic games \cite{AMira2007,BBL}).
For the case of average welfare and related metrics, the proposed estimator is a known function of choice probabilities and the structural parameter. In particular, if the model has ``terminal action'' property, value function or any other 
dynamic object does not have to be estimated at all.  We have applied these methods to estimate the average teachers welfare in an application to teachers absenteeism as in \cite{DHR}.

\newpage

\appendix

\section{Correction Term for Expected Per-Period Utility}
\label{sec:dynamicbinary}

In Section \ref{sec:estimation}, we introduce the estimator of welfare metric parameter $\delta_0$. In this Section, we derive the correction term $\phi_{\zeta}$ for the expected per-period utility. We focus on the  dynamic binary choice model in Section \ref{sec:setup}. The correction term is a sum of two terms 
\begin{align}
\label{eq:phizetaddc}
\phi_{\zeta}(Z) = \phi_{p}(Z)+  \phi_{\theta}(Z)
\end{align}
where $\phi_{p}(Z)$ accounts for estimation of CCPs and $\phi_{\theta}(Z)$ accounts for estimation of $\theta$.  Since $p(x)$ is a conditional mean function, the correction term follows from \cite{Newey1994}
\begin{align}
\label{eq:phicpp}
\phi_{p}(Z)   = \alpha(X) (u(X,1) - u(X,0) +\ln (1-p(X)) - \ln p(X))  (J-p(X)).
\end{align}
To specify the correction term $\phi_{\theta}(Z)$, we impose the following assumption.

 \begin{assumption}[Structural parameter]
 \label{ass:struct}
 We assume that $\theta_0$ is identified via a moment equation 
 $$
\E [ g_{\theta}(Z, \theta_0) ] =0.
$$
Furthermore, there exists an stimator $\widehat \theta$ of $\theta_0$ that is asymptotically linear 
$$
\sqrt{n} (\widehat \theta-\theta_0) = \E_n \psi_{\theta} (Z_i, \theta_0) + o_P(1),
$$
where $\psi_{\theta}$ is an influence function of $\theta_0$.
 \end{assumption}

Assumption \ref{ass:struct} simplifies our exposition. For the case of \cite{Rust} model, this assumption is satisfied for the debiased estimator proposed in e.g. \cite{LRSP} under some conditions. Then,  invoking \cite{Chernozhukov_2015} gives the  correction term for $\theta$ as
$$
\phi_{\theta}(Z) = \Gamma_{\theta}^{\prime} \Omega_{\theta}^{-1} g_{\theta}(Z, \theta)
$$
where the matrices are 
\begin{align*}
\Gamma_{\theta} &=    (\E[ \alpha(X) D_1(X)  p(X) ], \E[  \alpha(X) D_0(X) (1-p(X)) ])^{\prime}, \qquad \Omega_{\theta} = \text{Var} (g_{\theta}(Z, \theta_0)).
\end{align*}

\section{Proofs for Results referenced in Main Text }
\label{sec:proofs}

\begin{lemma}[Equivalence]
\label{lem:equivalence}
The fixed point of \eqref{eq:fdp} coincides with the net present discounted value \eqref{eq:npv2}.

\end{lemma}
\begin{proof}[Proof of Lemma \ref{lem:equivalence}]

Decomposing value function into the current $(t=0)$ and the future $(t \geq 1)$ periods  gives
\begin{align*}
V_0(X) &= \zeta_0(X) + \sum_{t=1}^{\infty} \beta^t \E [\zeta_0(X_t) \mid X]
\end{align*}
Rearranging time indices in the continuation value and concentrating $\beta$ out
$$
V_0(X) =  \zeta_0(X) + \beta \sum_{t=0}^{\infty} \beta^t \E [\zeta_0(X_{t+1}) \mid X].
$$
By Law of Iterated Expectations, the continuation value can be represented as a conditional expectation
$$
\sum_{t=0}^{\infty} \beta^t \E [\zeta_0(X_{t+1}) \mid X] = \E[ \sum_{t=0}^{\infty} \beta^t \E [\zeta_0(X_{t+1}) \mid X_{+}] \mid X] = \E[ V_0(X_{+}) \mid X].
$$
Putting the summands together gives \eqref{eq:fdp}.
\end{proof}

\begin{lemma}[Theorem 4.8, \cite{Kress}]
\label{lem:linearbounded}
Let $\mathcal{L}_2$ be the Hilbert space with the inner product $\ldot  f g = \E [f(X) g(X)]$. Then the following statement holds. 
(1) The operator $F(\zeta): \mathcal{L}_2 \rightarrow \mathrm{R}$  
\begin{align}
\label{eq:fzeta}
F(\zeta) :&=\E [w(X)V_{\zeta}(X)] = \E \bigg[ w(X) \left( \sum_{t \geq 0} \beta^t \E [ \zeta(X_t) \mid X] \right) \bigg]
\end{align}
is a linear operator 
\begin{align}
\label{eq:linearity}
F(\alpha \zeta_1 + \beta \zeta_2) = \alpha F(\zeta_1) + \beta F(\zeta_2).
\end{align}
(2) $F(\zeta)$ is a mean square continuous functional of $\zeta$, that is, 
\begin{align}
\label{eq:bounded}
| F(\zeta)  | &\leq \| w \|  (1-\beta)^{-1}  \| \zeta \|_2 \quad \forall \zeta.
\end{align}
(3) There exists a unique $\alpha_0 \in \mathcal{L}_2$ such that $F(\zeta) = \E [ \alpha_0(X) \zeta(X) ]$ for any $\zeta \in \mathcal{L}_2$ 
\end{lemma}

\begin{proof}[Proof of Lemma \ref{lem:linearbounded}]
(1) Linearity of \eqref{eq:linearity} follows from linearity of expectation. (2): We have that
\begin{align*}
\| V_{\zeta} (X) \| &= \| \sum_{t \geq 0 } \beta^t  \E[ \zeta(X_t)  \mid X] \|. 
\end{align*}
Triangular inequality for the norm, the properties of expectation, and stationarity
\begin{align*}
\| V_{\zeta} (X) \| & \leq \sum_{t \geq 0} \beta^t  \| \E[ \zeta(X_t)  \mid X] \|  \\
&\leq \sum_{t \geq 0} \beta^t  \| \zeta \| \leq (1-\beta)^{-1} \| \zeta \|.
\end{align*}
Cauchy Schwarz inequality implies
\begin{align*}
| F(\zeta)  | &\leq \| w \| \| V_{\zeta} \| \leq \| w \| (1-\beta)^{-1} \| \zeta \|.
\end{align*}
Therefore, $F(\zeta)$ is bounded.
(3) follows from Riesz representation lemma.
\end{proof}

\begin{proof} [Proof of Proposition \ref{lem:oaxaca}]
 The counterfactual welfare can be represented as
\begin{align*}
\delta_{\langle 1 \mid 0 \rangle}   &= \int_{\mathcal{S}} V_0^1(s) \pi^0(s)  ds = \int_{\mathcal{S}} V_0^1(s) \dfrac{\pi^0(s)}{\pi^1(s)} \pi^1(s) ds.
\end{align*}
Rewriting it in an expectation form is 
\begin{align*}
\delta_{\langle 1 \mid 0 \rangle}  &= \E \bigg[ V_0^1(S) \dfrac{\pi^0(S)}{\pi^1(S)} \mid K=1 \bigg] =^{(i)}\E \bigg[ V_0(X) \dfrac{\pi^0(S)}{\pi^1(S)}   \mid K=1 \bigg] \\
&= \dfrac{\E \bigg[ 1\{ K=1 \}  V_0(X) \dfrac{\pi^0(S)}{\pi^1(S)} \bigg]}{\Pr (K=1)}  \\
& = ^{(ii)} \dfrac{\E \bigg[ 1\{ K=1 \}  V_0(X) \dfrac{\Pr (K=0 \mid S) }{\Pr (K=1\mid S)} \bigg] }{\Pr (K=0)}=  \E [ w_0(X) V_0(X) ]
\end{align*}
which coincides with \eqref{eq:linearweight} with $w_0$ in \eqref{eq:compositionweight}. 
\end{proof}

\begin{proof}[Proof of Lemma \ref{lem:riesz}]

Let $\zeta$ be any integrable function and $V_{\zeta}(X)$ be defined in \eqref{eq:npv2}. We note that
\begin{align*}
\delta_0 = \E [w_0(X)V_{\zeta} (X)] &=^{(i)} \sum_{t=0}^{\infty}  \beta^t  \E [w_0(X) \zeta(X_t)]\\
&=^{(ii)} \sum_{t=0}^{\infty} \beta^t   \E[ w_0(X_{-t}) \zeta(X)] \\
&=^{(iii)} \E  [\alpha_0(X) \zeta(X)],
\end{align*}
where (i) follows from the definition of $V(X)$ in  equation \eqref{eq:npv2}, 
(ii) from strict stationarity and $$\E [w_0(X) \zeta (X_t)] = \E [w_0(X_{-t}) \zeta(X)]$$ for any $t$ and 
(iii) from the definition of dynamic dual representation $\alpha_0(X)$ is in \eqref{eq:alpha0}.  
Rewriting \eqref{eq:alpha0} and separating $w_0(X)$ from the rest  of the terms gives
\begin{align*}
\alpha_0(X) &= w_0(X) + \sum_{t \geq 1} \beta^t \E [ w_0(X_{-t}) \mid X] \\
&= w_0(X) + \beta \sum_{t \geq 0} \beta^t \E [ w_0(X_{-(t+1)}) \mid X] \\
&= w_0(X) + \beta \E [ \sum_{t \geq 0} \beta^t \E [ w_0(X_{-(t+1)}) \mid X_{-1}] \mid X]= w_0(X) + \beta \E [ \alpha_0(X_{-1}) \mid X].
\end{align*}
Replacing $X$ by $X_{+}$ gives $\alpha_0(X)$ as a solution to an integral equation  \eqref{eq:bdp}. Uniqueness of $\alpha_0$ follows from Riesz representation lemma whose sufficient conditions are verified in  Lemma \ref{lem:linearbounded}.

\end{proof}

\begin{proof}[Proof of Corollary \ref{cor:riesztimeinvar}]
Plugging  $w_0(X)=w_0(K)$ into equation \eqref{eq:alpha0}  for $\alpha_0(X)$  gives
\begin{align}
\label{eq:alphatimeinvar}
\alpha_0(X) = \sum_{t \geq 0} \beta^t \E [ w_0(X_{-t}) \mid X] = \sum_{t \geq 0} \beta^t \E [ w_0(K) \mid X] =(1-\beta)^{-1}w_0(K).
\end{align}
\end{proof}

 \begin{proof}[Proof of Lemma \ref{lem:orthod}]
 
 Let $V_0(X)$ be the value function defined in \eqref{eq:fdp}, and $V(X)$ be any other integrable function. Define the difference between the two
\begin{align*}
\Delta_V(X):= V(X) - V_0(X).
\end{align*}
Since $\lambda(Z, \cdot)$ is linear in $V$,  
\begin{align}
\label{eq:lambda}
 \lambda(Z,  V) -  \lambda(Z, V_0) &= \beta \Delta_V(X_{+}) - \Delta_V(X).
\end{align}

\textbf{ Step 1 }. We define the error terms. Decomposing 
\begin{align*}
\E [g(Z,  V, \alpha, \delta_0) ] = &\E [g(Z,  V, \alpha, \delta_0) - g(Z,  V_0, \alpha_0, \delta_0)]  \\
&= \E [ m(Z, V) - m (Z, V_0) ]  + \E [ \alpha (X)  \lambda(Z,  V) - \alpha_0(X) \lambda (Z, V_0) ] \\
&= \E [ m(Z, V) - m (Z, V_0) ] \\
&+ \E [ \alpha_0(X) ( \lambda(Z,  V) -  \lambda(Z, V_0)) ] \\
&+ \E [  (\alpha (X) - \alpha_0(X)) \lambda (Z, V) ].
\end{align*}
Invoking Riesz representation \eqref{eq:rieszweight} gives $\E [ m(Z, V) - m (Z, V_0) ] = \E [  w_0(X)  \Delta_V(X) ]=: S_1$.  Plugging \eqref{eq:lambda} into the second term 
$$
\E [ \alpha_0(X) ( \lambda(Z,  V) -  \lambda(Z, V_0)) ] = \E [ \beta \alpha_0(X) \Delta_V(X_{+})  -  \alpha_0(X)  \Delta_V(X)] =: S_2 + S_3.
$$
Then
\begin{align*}
\E [g(Z,  V, \alpha, \delta_0) ]=: S_1 + S_2 + S_3 +\E [  (\alpha (X) - \alpha_0(X)) \lambda (Z, V) ].
\end{align*}

\textbf{ Step 2 }.  We show that $S_1 + S_2 + S_3=0$, which suffices to prove \eqref{eq:mainorthog}.  Invoking stationarity allows to replace $X$ by $X_{+}$ in the first and third summand
\begin{align*}
S_1 = \E  [w(X_{+})  \Delta_V(X_{+})], \qquad S_3 = \E  [\alpha(X_{+})  \Delta_V(X_{+})],
\end{align*}
which allows concentrating out $\Delta_V(X_{+})$, that is
\begin{align*}
S_1 + S_2 + S_3 &=  \E [(w(X_{+}) +  \beta  \alpha_0(X)  -  \alpha_0(X_{+}))  \Delta_V(X_{+})].
\end{align*}
By definition of $\alpha_0(X)$ as a fixed point of \eqref{eq:bdp}, $S_1+S_2 +S_3=0$ which gives \eqref{eq:mainorthog}. Since $\E [ (\alpha (X) - \alpha_0(X)) \lambda (Z, V_0) ] = 0$, the final term can be written as the product
$$ \E [  (\alpha (X) - \alpha_0(X)) \lambda (Z, V) ] =  \E [  (\alpha (X) - \alpha_0(X)) (\lambda (Z, V) - \lambda (Z, V_0)) ],$$ 
which gives \eqref{eq:mainorthog}.

\textbf{ Step 3 }. We prove \eqref{eq:doublerobustnessmain}. Invoking Cauchy Schwartz  gives
\begin{align*}
| \E [  (\alpha (X) - \alpha_0(X))  \Delta_V(X) ] | \leq \| \alpha - \alpha_0 \| \| V - V_0 \|.
\end{align*}
Invoking Cauchy Schwartz and stationarity gives
\begin{align*}
| \E [  (\alpha (X) - \alpha_0(X))  \Delta_V(X_{+}) ] | &\leq (\E [  (\alpha (X) - \alpha_0(X))^2 ])^{1/2}  (\E [  (V (X_{+}) - V_0(X_{+}))^2 ])^{1/2} \\
& \leq    \| \alpha - \alpha_0 \| \| V - V_0 \|.
\end{align*}
Adding the terms together gives an upper bound
\begin{align*}
| \E [g(Z,  V, \alpha, \delta_0) ]  | &\leq \beta |  \E [  (\alpha (X) - \alpha_0(X)) \Delta_V(X_{+}) ] | +  |  \E [  (\alpha (X) - \alpha_0(X)) \Delta_V(X) ] | \\
&\leq (\beta+1) \| \alpha - \alpha_0 \| \| V - V_0 \|,
\end{align*}
which coincides with \eqref{eq:doublerobustnessmain}.

\end{proof}

\begin{proof}[Proof of Proposition \ref{prop:avder}]
The proof has three steps. Step 1 establishes \eqref{eq:linearweight}. Steps 2 and 3 establish  \eqref{eq:linearweightalpha}.

\textbf{Step 1. }  We prove  \eqref{eq:linearweight}. The stationary distribution of $S$ given $K$ is $N(0, \sigma^2(K))$ where $\sigma^2(K) = (1-\rho^2(K))^{-1}$.

 Taking log of the conditional density gives 
\begin{align*}
\ln f_{S \mid K} (S \mid K ) &=   [ (-2 \sigma^2(K))^{-1}  ]  S^2 + [ -1/2 \ln ( 2 \pi \sigma^2(K)) ],
\end{align*}
whose derivative with respect to $K$ is
\begin{align}
\label{eq:partiak}
\partial_K \ln f_{S \mid K} (S \mid K ) =  \partial_K [ (-2 \sigma^2(K))^{-1}  ]  S^2 + \partial_K [-1/2 \ln ( 2 \pi \sigma^2(K)) ].
\end{align}
Since the log joint density is  
$$
\ln f(S,K) =  \ln f_{S \mid K}(S \mid K) + \ln f_K(K)
$$
differentiating both sides with respect to $K$ gives
\begin{align}
\label{eq:partiak2}
\partial_K \ln f(S,K)   =  \partial_K  \ln f_{S \mid K}(S \mid K) + \partial_K  \ln f_K(K).
\end{align}
Plugging \eqref{eq:partiak} into \eqref{eq:partiak2} gives 
\begin{align}
\label{eq:linearweightapp}
w_0(X) = \underbracket{\partial_K [ (-2 \sigma^2(K))^{-1}  ]}_{\gamma_1(K)}  S^2 + \underbracket{\partial_K [-1/2 \ln ( 2 \pi \sigma^2(K)) -  \ln f_K(K) ]}_{\gamma_0(K)}
\end{align}
which is a linear function of $S^2$ with slope $\gamma_1(K)$ and intercept $\gamma_0(K)$.

\textbf{Step 2. } For any step $k \geq 1$ and any time $t \geq 0$, we show that 
\begin{align}
\label{eq:expk}
\E [ S^2_t \mid S_{t+k}=S, K ] = \dfrac{1 -  \rho^{2k} (K) }{1-\rho^2 (K)} + \rho^{2k} (K) S^2.
\end{align}
The proof is established using an inductive argument. In the base case $k=0$, \eqref{eq:expk} holds trivially. To verify the inductive hypothesis for 
$k=l$, note that  
$$
\E [ S_t \mid S_{t+1} =S, K] = \rho(K) S, \quad \text{Var} (S_t \mid S_{t+1}, K) = \sigma^2(K) (1-\rho^2(K)) = 1. 
$$
Squaring the first term and adding variance term gives
\begin{align}
\label{eq:exp1}
\E [ S^2_t \mid S_{t+1}, K ] = 1 + \rho^2 (K) S^2,
\end{align}
which is a special case of \eqref{eq:expk} with $k=1$. Next, 
\begin{align*}
\E [ S^2_t \mid S_{t+l+1}=S ] &=^{i} \E [ \E [ S^2_t \mid S_{t+l}, K ] \mid S_{t+l+1}, K] \\
&=^{ii}  \E [ 1 + \rho^2 (K) S^2_{t+l} \mid S_{t+l+1}=S, K]  \\
&=^{iii} 1 + \rho^2(K) \dfrac{1 -  \rho^{2l} (K) }{1-\rho^2 (K)} + \rho^{2l+2} (K) S^2 \\
&=^{iv} \dfrac{ 1 -  \rho^{2l} (K) + \rho^2(K)  - \rho^{2(l+1)} (K) }{1-\rho^2(K)} + \rho^{2l+2} (K) S^2,
\end{align*}
where (i) follows from Law of Iterated Expectations, (ii) from the inductive hypothesis \eqref{eq:expk} with $k=l$, 
(iii) from the inductive step \eqref{eq:exp1} at $t=(t+l)$. Simplifying the algebra in (iv) gives an expression that is  a special case of \eqref{eq:expk} with $k=(l+1)$.

\textbf{Step 3. } We prove \eqref{eq:linearweightalpha}.  Plugging $S^2$ into \eqref{eq:bdp} gives
\begin{align*}
&\sum_{t \geq 0} \beta^t \E [ S^2_{-t} \mid S_0=S, K ] \\
&= \dfrac{ \sum_{t \geq 0} \beta^t }{1-\rho^2 (K)} + \left(  \dfrac{1 -  \rho^{2t} (K) }{1-\rho^2 (K)} + \rho^{2t} (K) S^2 \right) \\
&= \dfrac{ \sum_{t \geq 0} \beta^t }{1-\rho^2 (K)}  - \dfrac{\sum_{t \geq 0} \beta^t  \rho^{2t} (K) }{1-\rho^2 (K)} + \sum_{t \geq 0} \beta^t \rho^{2t} (K) S^2 \\
&=\dfrac{ \sum_{t \geq 0} \beta^t }{1-\rho^2 (K)} - \dfrac{1}{ (1-\beta \rho^2(K)) (1- \rho^2(K))} + \dfrac{1}{1-\beta \rho^2(K)} S^2 \\
&= \dfrac{1}{1-\beta \rho^2(K)} S^2 + \dfrac{ \beta }{1-\beta \rho^2(K)}.
\end{align*}
Multiplying the discounted sum above by $\gamma_1(K)$ and adding $\gamma_0(K) \sum_{t \geq 0} \beta^t$ gives \eqref{eq:linearweightalpha}.

\end{proof}

\begin{proof}[Proof of Proposition \ref{prop:vee}]
The statement of Proposition follows from \eqref{eq:vee1}--\eqref{eq:vee3}. Since $\beta <1$, the operator $\bbI -\bbA$ is invertible, the minimizer of \eqref{eq:vee3} is unique.

\end{proof}

\begin{proof}[Proof of Proposition \ref{prop:alpha}]
The statement of Proposition follows from \eqref{eq:alpha1}--\eqref{eq:alpha2}. Since $\beta <1$, the operator $\bbI -\bbA^{*}$ is invertible, the minimizer of \eqref{eq:alpha4} is unique.

\end{proof}

\subsection{Proof of Theorem \ref{lem:lassovalue} }
\label{sec:fslasso}

\begin{proof}[Proof of Theorem \ref{lem:lassovalue}]

 The proof builds on an argument similar to  Lemmas A1--A6 and Theorem 1 in  \cite{chernozhukov2021automatic}  (cf \cite{bradic2022}). The proof  of Theorem \ref{lem:lassovalue} below differs in two respects. First, it generalizes the proof of Theorem 1 by considering projections onto \textit{transformed} basis $\psi(x) = (\bbI - \bbA_0) b$.  Second, it drops Assumption 3 of \cite{chernozhukov2021automatic}.
 
 \textbf{ Step 0. (Notation) } Let  $\varepsilon_n$ be an upper bound obeying
\begin{align}
\label{eq:a6}
\| \widehat G - G \|_{\infty} = O_P (\varepsilon_n), \quad \| \widehat M - M \|_{\infty} = O_P (\varepsilon_n), \quad \sqrt{\ln p^2/n} = o(\varepsilon_n).
\end{align}
By Assumption \ref{ass:approx} we can define $J_0$ as a set of indices of a sparse approximation with $|J_0| = s_0$ with $s_0 \geq C  \varepsilon_{n}^{-2/(2\xi+1)}$  and coefficients $\widetilde{\rho}_j$ for $j \in J_0$ such that for $\widetilde{V}_0(x) = \sum_{j \in J_0} \widetilde{\rho}_j b_j(x)$ satisfies 
\begin{align}
\label{eq:bound}
\E ( V_0 (X) -  \widetilde{V}_0(X))^2 \leq C s^{-2\xi}_0
\end{align}
Define $\psi(x) = (\bbI - \bbA_0) b (x)$ as the vector of projected basis functions. Note that the projection coefficient
\begin{align}
\label{eq:projrho}
\rho = \arg \min_{v} \| \zeta_0 - \psi^{\prime} v \|
\end{align}
can be equivalently written as 
\[
G_V \rho - M_V = 0,
\]
with \( G_V = \mathbb{E}[ \psi(X) \psi^{\prime}(X)] \) and $M_V=\mathbb{E} [\psi(X) \zeta_0(X)]$.  Define
\begin{equation}
\rho_{\ast}\in\arg\min_{v}\;(\rho-v)\prime
G_V(\rho-v)+2\varepsilon_{n}\sum_{j\in J_{0}^{c}}|v_{j}|\text{.} \label{pistar}
\end{equation}
Let $J$ to be the vector of indices of nonzero elements of $\rho_{\ast}$ and $\left\vert A\right\vert $ be the number non zero elements of any finite set $A$, and let $J^c$ denote $J^c = \{1, 2, \dots, p \} \setminus J$.

\textbf{ Step 1. } We verify the analogs of Lemmas A1--A2 hold with $G_V= \E [ \psi(X) \psi^{\prime} (X)]$ instead of $\E[ b(X) b^{\prime}(X)]$. The statement of Lemma A1 
\begin{align}
\label{lem:a1}
\| G_V (\rho_{\ast} - \rho) \|_{\infty} \leq \varepsilon_n 
\end{align}
follows from the definition of $\rho_{\ast}$ and $\rho$. To verify the statement of Lemma A2, note that
\begin{align*}
(\rho -\rho_{\ast})^{\prime} G_V (\rho - \rho_{\ast}) &\leq (\rho - \bar{\rho})^{\prime} G_V (\rho - \bar{\rho}) \\
&= \| \psi^{\prime} (\rho - \bar{\rho}) \|^2_2 \\
&\leq 2 \| \psi^{\prime} \rho - \zeta_0 \|^2_2 + 2 \| \psi^{\prime} \bar{\rho} - \zeta_0 \|^2_2 \leq 4 \| \psi^{\prime} \bar{\rho} - \zeta_0 \|^2_2,
\end{align*}
where the last inequality follows from the definition of the projection coefficient $\rho$ in \eqref{eq:projrho}. As shown in the proof of Theorem \ref{thm1}, the norm bound
\begin{align}
\label{eq:normbound}
1- \beta \leq \| \bbI - \bbA_0 \|_2 \leq 1+\beta.
\end{align}
implies
\begin{align*}
 (1+\beta)^2 \lambda_{\max} (\E [ b(X) b(X)^{\prime} ])  &\geq \lambda_{\max} (G_V) \\
 &\geq \lambda_{\min} (G_V) \geq (1-\beta)^2 \lambda_{\min} (\E [ b(X) b(X)^{\prime} ] ) >0
\end{align*}
 Plugging $\zeta_0 = (\bbI - \bbA_0) V_0$ and $\psi = (\bbI - \bbA_0) b$  gives
$$
\| \psi^{\prime} \bar{\rho} - \zeta_0 \| = \| (\bbI - \bbA_0) (V_0 - b^{\prime} \bar{\rho})  \| \leq \| \bbI - \bbA_0 \|_2 \| V_0 - b^{\prime} \bar{\rho}  \|_2 \leq (1+\beta) s_0^{-\xi}.
$$
Combining the bounds and noting that $s_0 \geq C \varepsilon^{-2/(2\xi+1)}_n$ 
\begin{align}
\label{lem:a2}
(\rho -\rho_{\ast})^{\prime} G_V (\rho - \rho_{\ast}) &\leq (\rho - \bar{\rho})^{\prime} G_V (\rho - \bar{\rho}) \leq 4 (1+\beta)^2 C^2 \varepsilon^{4\xi/(2\xi+1)}_n.
\end{align}

\textbf{ Step 2. } We show that $\| V_0 - b^{\prime} \rho_{\ast} \|^2 = O_P (\varepsilon^{4\xi/(2\xi+1)}_n)$. Noting that
\begin{align*}
&\| b^{\prime} (\bar{\rho} - \rho_{\ast}) \|^2 = \|( \bbI - \bbA_0)^{-1}  \psi^{\prime} (\bar{\rho} - \rho_{\ast}) \|^2 \\
& \leq^{i} (1-\beta)^{-2} (\bar{\rho} - \rho_{\ast})^{\prime} G (\bar{\rho} - \rho_{\ast}) \\
&\leq^{ii}  2 (1-\beta)^{-2} (\bar{\rho} - \rho)^{\prime} G (\bar{\rho} - \rho) + 2 (1-\beta)^{-2} (\rho -\rho_{\ast})^{\prime} G (\rho - \rho_{\ast}) \\
&\leq^{iii} 8 (1-\beta)^{-2}  (1+\beta)^2 C^2 \varepsilon^{4\xi/(2\xi+1)}_n.
\end{align*}
where (i) follows from \eqref{eq:normbound} and (ii) from triangle inequality and (iii) from \eqref{lem:a2}. Invoking triangle inequality gives
\begin{align}
\label{lem:a22}
\| V_0 - b^{\prime} \rho_{\ast} \| \leq \| V_0 - b^{\prime} \bar{\rho} \| + \| b^{\prime} (\bar{\rho} - \rho_{\ast}) \| = O (\varepsilon^{2\xi/(2\xi+1)}_n) + O (\varepsilon^{2\xi/(2\xi+1)}_n) = O (\varepsilon^{2\xi/(2\xi+1)}_n),
\end{align}
where $\| V_0 - b^{\prime} \bar{\rho} \|  = O (\varepsilon^{2\xi/(2\xi+1)}_n) $ follows from \eqref{eq:bound}.

\textbf{ Step 3. } We verify Lemmas A3--A4. The statement of Lemma A3 holds with 
\begin{align}
\label{lem:a3}
|J| \leq C \varepsilon^{-2/(2\xi+1)}
\end{align}
which follows from Steps 1--3 and $\lambda_{\max} (G_V) \leq (1+\beta) \bar{\lambda}$. The analogs of Lemmas A4 and A5 also hold.

\textbf{ Step 4. } We show that, for any $\Delta \in \mathrm{R}^p: \| \Delta_{J^c} \|_1 \leq 3 \| \Delta_J \|_1$, we have
\begin{align}
\Delta^{\prime} G \Delta \leq 2 \Delta^{\prime} \widehat{G} \Delta \label{eq:multbound}
\end{align}
For any $\Delta \in \mathrm{R}^p$,  the properties of the norm imply
\begin{align}
|\Delta^{\prime} G \Delta- \Delta^{\prime} \widehat{G} \Delta| &\leq \| \widehat{G} - G \|_{\infty} \| \Delta \|^2_1. \label{eq:multbound1}
\end{align}
Since $\Delta \in \mathrm{R}^p$ belongs to restricted set $\| \Delta_{J^c} \|_1 \leq 3 \| \Delta_J \|_1$, additionally, 
$$
\| \Delta \|^2_1 \leq 16  \| \Delta_J \|^2_1.
$$
Next, the term $ \| \Delta_J \|^2_1$ can be upper bounded as 
\begin{align*}
 \| \Delta_J \|^2_1 &\leq  |J| \| \Delta_J \|^2_2 \leq | J |  \lambda^{-1}_{\min}(G) \Delta^{\prime} G \Delta.
\end{align*}
where (i) follows from Cauchy inequality ad (ii) from the definition of $ \lambda_{\min}(G)$. Invoking Lemma A3 in \cite{chernozhukov2021automatic} gives $| J | \leq \varepsilon_n^{-2/(2\xi+1)} $ which gives 
\begin{align*}
| J |  \lambda^{-1}_{\min}(G) \Delta^{\prime} G \Delta \leq \varepsilon_n^{-2/(2\xi+1)} \lambda^{-1}_{\min}(G) \Delta^{\prime} G \Delta.
 \end{align*}
 Invoking \eqref{eq:multbound1} gives a multiplicative error bound
 \begin{align*}
\Delta^{\prime} G \Delta &\leq (1+32  \varepsilon_n^{(2\xi-1)/(2\xi+1)} ) \Delta^{\prime} \widehat{G} \Delta \leq 2  \Delta^{\prime} \widehat{G} \Delta,
\end{align*}
that matches \eqref{eq:multbound}.

\textbf{ Step 5. }  Let $\Delta=\widehat{\rho}-\rho_{\ast}$ be the estimation error.  The following upper bound applies 
\begin{align*}
\Delta^{\prime} G \Delta &\leq^{i} 2 \Delta^{\prime} \widehat{G} \Delta  \\
&\leq^{ii} 6 r \| \Delta \|_1 \leq^{iii} 24 r \| \Delta_J \|_1 \leq^{iv} 24r |J|^{1/2} \| \Delta_J \|_2 \\
&\leq^{v} \dfrac{24r}{\lambda^{1/2}_{\min}(G)} |J|^{1/2}  (\Delta^{\prime} G \Delta)^{1/2},
\end{align*}
where (i) follows from \eqref{eq:multbound},  the statements (ii)--(iii) follow the statement in Lemma A5, (iv) follows from Cauchy inequality, and (v) noting that
$$
\Delta^{\prime} G \Delta \geq \lambda_{\min}(G) \| \Delta \|^2 \geq \lambda_{\min}(G) \| \Delta_J \|^2.
$$
Dividing by $(\Delta^{\prime} G \Delta)^{1/2}$ gives
\begin{align*}
\Delta^{\prime} G \Delta =^{i} O_P ( |J|^{1/2} r) = O_P (\varepsilon_n^{-1/(2\xi_V+1)} \varepsilon_n n^c ) =^{ii} O_P (\varepsilon_n^{ (2 \xi_V)/(2\xi_V+1)} n^c ),
\end{align*}
where (i) follows from Step 3 and (ii) from the choice of $r_V$ in A5 of Assumption \ref{ass:approx}.

\textbf{ Step 6. } We verify that the bound \eqref{eq:a6} holds with $\varepsilon_n = n^{-d_V}$ where $ \sqrt{\ln p_V^2/n} + a_n +\zeta_n = o ( n^{-d_V})$.  Define
\begin{align*}
\check G_V &= \E_n (b(X_i) -  \bbA_0 b) (X_i)) ( b(X_i) - \bbA_0 b (X_i))^{\prime} \\
\check  M_V &= \E_n (b(X_i) - \beta b (X_{i+}))  \widehat \zeta (X_i).
\end{align*} 
where $\E_n[ \cdot ]$ is the sample average calculated on respective partition. Decomposing error term gives
\begin{align*}
\widehat G_V - \check G_V &= \E_n  (b(X_i) -\bbA_0 b (X_i))  ((\bbA_0 b - \widehat A b) (X_i))^{\prime} \\
&  \E_n  (b(X_i) -\bbA_0 b (X_i)) ((\bbA_0 b - \widehat A b) (X_i))^{\prime} \\
&+  \E_n  ( (\bbA_0 b - \widehat A b) (X_i))  ( (\bbA_0 b - \widehat A b) (X_i))^{\prime}   \\
\widehat M_V - \check M_V &= \E_n (b(X_i) - \beta b (X_{i+})) (\zeta_0 (X_i) - \widehat \zeta (X_i)).
\end{align*}
By Law of Large Numbers, the following concentration bounds hold for sample averages
\begin{align*}
\| \check G_V - G_V \|_{\infty} = o_P (\sqrt{\ln p_V^2/n}), \quad \| \check M_V - M_V \|_{\infty} = o_P (\sqrt{\ln p_V^2/n}).
\end{align*}
as well as for demeaned error terms
$$
\| \widehat G_V - \check G_V \|_{\infty} = o_P ( n^{-d_V}), \quad \| \widehat M_V - \check M_V \|_{\infty} = o_P ( n^{-d_V}).
$$

\textbf{ Step 7. (Conclusion) }  The upper bound can be seen as 
\begin{align*}
\Vert\widehat{V}- V_0\Vert^2 &\leq 2\Vert V_0- b^{\prime} \rho_{\ast}  \Vert^2  + 2 \Vert b^{\prime} (\widehat \rho- \rho_{\ast} ) \Vert^2 \\
&\leq 2C s_0^{-2 \xi} + \Delta^{\prime} (\E [ b(X) b(X)^{\prime} ]) \Delta. 
\end{align*}
The approximation error term is bounded by $s_V^{-2 \xi_V} \leq Cn^{-2 d_V \xi_V/ (2\xi_V+1)}$ as assumed in  Assumption \ref{ass:approx}. The second term is bounded as
\begin{align*}
&\Delta^{\prime} (\E [ b(X) b(X)^{\prime} ]) \Delta \\
&\leq  \bar{\lambda} \Delta^{\prime}  \Delta  \leq ( \bar{\lambda}/\underline{\lambda}) \Delta^{\prime}  G   \Delta  =O_P (\varepsilon_n^{ 2(2 \xi)/(2\xi+1)} n^c ).
\end{align*}

\end{proof}

\begin{proof}[Proof of Theorem \ref{lem:lassovalue2}]
We focus on the case when $m(\cdot, \cdot)$ is known. By Law of Large Numbers, the following concentration bounds hold for sample averages
\begin{align*}
\| \widehat G^{\alpha} - G^{\alpha} \|_{\infty} = o_P (\sqrt{\ln p_{\alpha}^2/n}), \quad \| \widehat M^{\alpha} - M^{\alpha} \|_{\infty} = o_P (n^{-d_{\alpha}}).
\end{align*}
which implies $\varepsilon_n = n^{-d_{\alpha}}$ where $ \sqrt{\ln p^2/n} + a^{*}_n  = o ( n^{-d_{\alpha}})$ suffices for 
\begin{align}
\label{eq:a62}
\| \widehat G^{\alpha} - G^{\alpha} \|_{\infty} = O_P (\varepsilon_n), \quad \| \widehat M^{\alpha} - M^{\alpha} \|_{\infty} = O_P (\varepsilon_n).
\end{align}
The rest of the proof follows similarly to the proof of Theorem \ref{lem:lassovalue}.

\end{proof}

\section{Asymptotic Theory}
\label{sec:atheory}

\subsection{Time-Invariant Case. }

In this section we derive sufficient conditions  for the estimator $\widehat{\delta} $ to be asymptotically linear.
\begin{align}
\label{eq:alinearity}
\sqrt{n} (\widehat{\delta}  -\delta_0) &=  \frac{1}{n} \sum_{\ell = 1}^L \sum_{i \in I_\ell}[(1-\beta)^{-1}\widehat{w}_\ell(K_i)\widehat\zeta_\ell(X_i) +  \widehat\phi_{w\ell}(Z_i) + \widehat\phi_{\zeta\ell}(Z_i) + o_P(1).
\end{align}

Assumption \ref{ass:consistency} is a mild  consistency condition.
 
\begin{assumption}
\label{ass:consistency}
 (1)For any partition index $\ell = 1, 2,\dots, L$ the following convergence conditions hold:
 \begin{enumerate}
 \item[A]  $\int ( \widehat\phi_{\zeta\ell} (z) - \phi_{\zeta} (z))^2 F_0(dz)  = o_P(1)$ 
 \item[B] $\int (\widehat{\phi}_{m \ell} (z) - \phi_{m} (z))^2 F_0(dz)  = o_P(1)$
 \item[C] $\| \widehat \zeta_{\ell} - \zeta_0 \| = o_P(1)$.
 \end{enumerate}
(2) For any partition index $\ell = 1, 2,\dots, L$,  $\| \widehat w_{\ell} - w \| = o_P(1)$.
\end{assumption}

Assumption \ref{ass:corhigher}  is a small bias condition that controls higher order bias of plug-in estimators.

\begin{assumption}
\label{ass:corhigher}
 For any partition index $\ell = 1, 2,\dots, L$ the following higher-order bias terms are negligible
 \begin{align*}
 n^{1/2} \int_{\mathcal{Z}} [ (1-\beta)^{-1} w_0(k) (\widehat \zeta_{\ell}(x)- \zeta_0(x)) + \widehat \phi_{\zeta\ell}(z) - \phi_{\zeta} (z)  ] F_0(dz)= o_P(1) \\
n^{1/2} \int_{\mathcal{Z}} [ (1-\beta)^{-1} (\widehat w_{\ell} (k) - w_0(k)) \zeta_0(x) +  \widehat \phi_{w\ell}(z) - \phi_w (z) ] F_0(dz)= o_P(1) 
 \end{align*}
 \end{assumption}

Assumption \ref{ass:rate} includes a product rate condition as well as the technical conditions on the estimators of $w$ and $\zeta$.

\begin{assumption}
\label{ass:rate}
For any partition index $\ell = 1, 2,\dots, L$  (1) the following products of first-stage errors converge fast enough
\begin{align*}
n^{1/2} \| \widehat w_{\ell} - w_0 \| \| \widehat \zeta_{\ell} - \zeta_0 \| = o_P(1).
\end{align*}
(2) The  estimate $\widehat w_{\ell} (k)$, as well as the weight function $w_0(k)$, and  the estimate $\widehat \zeta_{\ell}(x)$, as well as the function $\zeta_0(x)$, are uniformly bounded over the support of $\mathcal{X}$.
\end{assumption}

\begin{lemma}[Asymptotic Theory for Time-Invariant Case]
\label{lem:alinearity}
If Assumptions \ref{ass:consistency}--\ref{ass:rate} hold then  \eqref{eq:alinearity}  holds, and the asymptotic variance estimate is consistent 
\begin{align}\label{eq:consistencyomega}  \widehat \Omega \rightarrow^p \Omega. \end{align}
\end{lemma}

\begin{proof}[Proof of Lemma \ref{lem:alinearity}]
Throughout this proof, let $C > 0$ denote a generic constant (possibly different each time it appears).

\textbf{ Step 1 }. Let $\ell \in \{1,2,\dots, L\}$ indicate the partition index. Define the following error terms
\begin{align*}
R_{1,\ell}(Z) &= (1-\beta)^{-1} w_0(K) (\widehat \zeta_{\ell}(X) - \zeta_0(X)) + \widehat \phi_{\zeta\ell}(Z) - \phi_{\zeta}(Z) \\
R_{2,\ell}(Z) &= (1-\beta)^{-1} (\widehat w_{\ell} (K) - w_0(K)) \zeta_0(X) + \widehat \phi_{w\ell}(Z) - \phi_w(Z) \\
R_{3,\ell}(Z) &= (1-\beta)^{-1} (\widehat w_{\ell} (K) - w_0(K)) (\widehat \zeta_{\ell}(X) - \zeta_0(X)) 
\end{align*}
and decompose the error term as 
$$
\psi(Z_i, \widehat \gamma_{\ell},  \widehat \phi_{\ell}) - \psi(Z_i, \gamma_0, \phi_0) = R_{1, \ell}(Z_i) + R_{2, \ell}(Z_i) + R_{3, \ell}(Z_i).
$$
 Step 2 shows 
 \begin{align}
\label{eq:r2ell}
 n^{-1/2} [\sum_{i=1}^n R_{2, \ell}(Z_i) + R_{3, \ell}(Z_i)]  = o_P(1).
\end{align}
Step 3 shows 
\begin{align}
\label{eq:r1ell}
n^{-1/2} \sum_{i=1}^n R_{1, \ell}(Z_i)  = o_P(1).
\end{align}
Step 4 shows \eqref{eq:consistencyomega}.

\textbf{ Step 2 (Proof of  \eqref{eq:r2ell})}.  By Assumption \ref{ass:consistency}, we have $n^{1/2}\E [ R_{2,\ell}(Z_i) + R_{1, \ell} (Z_i) ] = o_P(1)$. Furthermore, we have 
$$
\E [R^2_{2, \ell} (Z) ] \leq 2  \E (\widehat \phi_{w\ell}(Z) - \phi_w(Z))^2  + 2 (1-\beta)^{-2} C \E (\widehat w_{\ell} (K) - w_0(K))^2 = o_P(1)
$$
and
$$
\E [R^2_{1, \ell} (Z) ] \leq 2  \E (\widehat \phi_{\zeta\ell}(Z) - \phi_{\zeta}(Z))^2  + 2 (1-\beta)^{-2} C \E (\widehat \zeta_{\ell} (X) - \zeta_0 (X))^2 = o_P(1)
$$
where $C$ is a uniform bound in Assumption \ref{ass:rate}.  Collecting the terms gives \eqref{eq:alinearity}.

 \textbf{ Step 3 (Proof of  \eqref{eq:r1ell}) }.  We show \eqref{eq:r1ell}.   By Assumption \ref{ass:rate} (1), 
\begin{align*}
n^{1/2} \E [R_{1,\ell}(Z)] &= n^{1/2} \int_{\mathcal{Z}} (\widehat w_{\ell} (k) - w_0(k)) (\widehat \zeta_{\ell}(x) - \zeta_0(x)) F_0(dx) \\
&\leq n^{1/2} \| \widehat w_{\ell} - w_0 \| \cdot \| \widehat \zeta_{\ell} - \zeta_0 \| = o_P(1).
\end{align*}
By Assumptions \ref{ass:consistency} (3)-(4) and \ref{ass:rate} (2),  the term $\E [R^2_{1,\ell}(Z)]$ is bounded as 
$$
\E [R^2_{1,\ell}(Z)] \leq  2C (1-\beta)^{-1} \min (\| \widehat w_{\ell} - w_0 \|, \| \widehat \zeta_{\ell} - \zeta_0 \|) = o_P(1)
$$
where $C$ is a uniform bound in Assumption \ref{ass:rate}. Steps 1--3 imply that \eqref{eq:alinearity} holds, and, therefore, $\widehat \delta \rightarrow^p \delta_0$.

\textbf{ Step 4 (Proof of \eqref{eq:consistencyomega})}.  Assumption \ref{ass:consistency} gives
\begin{align*}
&\int_{\mathcal{Z}} (\widehat w_{\ell} (k) \widehat \zeta_{\ell} (x) - w_0 (k) \zeta_0(x))^2 F_0(dz) \\
& \leq 2\int_{\mathcal{Z}} (\widehat w_{\ell} (k) \widehat \zeta_{\ell} (x) - w_0(k) \widehat \zeta_{\ell}(x))^2 F_0(dz) + \int_{\mathcal{Z}} (w_0(k) (\widehat \zeta_{\ell}(x)-\zeta_0(x) ))^2 F_0(dz)  \\
& \leq C \| \widehat w_{\ell}  - w_0 \|^2 + C \| \widehat {\zeta}_{\ell}  - \zeta_0 \|^2.
\end{align*}
Decomposing the error of the original moment gives 
\begin{align*}
&\int_{\mathcal{Z}}( (1-\beta)^{-1} \widehat w_{\ell} (k) \widehat \zeta_{\ell} (x) - (1-\beta)^{-1} w_0 (k) \zeta_0(x) - \widehat \delta + \delta_0)^2 F_0 (dz) \\
&\leq  C \| \widehat w_{\ell}  - w_0 \|^2 + C \| \widehat {\zeta}_{\ell}  - \zeta_0 \|^2 + C \int_{\mathcal{Z}} (\widehat \delta -\delta_0)^2 F_0 (dz) = o_P(1) + o_P(1) = o_P(1).
\end{align*}
Invoking Lemma 16 in CEINR gives $\widehat{\Omega} \rightarrow^p \Omega$.
\end{proof}

In this Section we derive sufficient conditions  for the estimator $\widehat{\delta} $ of $\delta_0$ to be asymptotically linear, that is 
\begin{align}
\label{eq:alinearity2}
\sqrt{n} (\widehat{\delta}  -\delta_0) &= \frac{1}{\sqrt{n}} \sum_{i=1}^n [m_0(Z_i, V_0) -\delta_0 + \alpha_0(X_i) \lambda(Z_i, V_0) + \phi_{\zeta}(Z_i) + \phi_w(Z_i)] + o_P(1).
\end{align}

\begin{assumption}
\label{ass:corhigher2}
 For any partition index $\ell = 1, 2,\dots, L$ the following higher-order bias terms are negligible
 \begin{align*}
 n^{1/2} \int_{\mathcal{Z}} [ \alpha_0(x) (\widehat \zeta_{\ell}(x)- \zeta_0(x)) + \widehat \phi_{\zeta\ell}(z) - \phi_{\zeta} (z)  ] F_0(dz)= o_P(1) \\
n^{1/2} \int_{\mathcal{Z}} [ \widehat m_{\ell} (z, V_0) - m_0(z, V_0) +  \widehat \phi_{m\ell}(z) - \phi_m (z) ] F_0(dz)= o_P(1) 
 \end{align*}
 \end{assumption}

\begin{assumption}
\label{ass:ratealpha}
 For any partition index $\ell = 1, 2,\dots, L$  (1) the following products of first-stage errors converge fast enough
\begin{align*}
n^{1/2} (\| \widehat V_{\ell} - V_0 \| + \| \widehat \zeta_{\ell} - \zeta_0 \|) \| \widehat \alpha_{\ell} - \alpha_0 \| = o_P(1).
\end{align*}
(2)  the function $\alpha_0(x)$ and its estimate $\widehat \alpha_{\ell}(x)$ are uniformly bounded over the support of $\mathcal{X}$ by some constant $C_{\alpha}$. \newline
(3) the function $\E [m_0^2 (Z, V)] \leq C_m \| V \|^2$ for some finite $C_m$ and for any integrable function $V$.
\end{assumption}

\begin{assumption}
\label{ass:crossm}
 For any partition index $\ell = 1, 2,\dots, L$  (1) the following condition holds
\begin{align}
\label{eq:doubly}
 n^{1/2} \int_{\mathcal{Z}} ( \widehat m_{\ell} (z, \widehat V_{\ell})  - m_0 (z, \widehat V_{\ell}) - \widehat m_{\ell} (z, V_0) + m_0(z, V_0)) F_0 (dz) = o_P(1).
\end{align}
 \end{assumption}
 
\begin{example}[continues=ex:avder]
Since $m(z,V) = \partial_K V(X)$ is a known functional of $V$,  Assumption \ref{ass:crossm} is automatically satisfied. 
\end{example}

\begin{lemma}[Asymptotic Theory for Time-Variant Case]
\label{lem:alinearity2}
If Assumptions \ref{ass:consistency}(1), \ref{ass:rate}, \ref{ass:corhigher2}, \ref{ass:ratealpha}, \ref{ass:crossm} hold,  then  \eqref{eq:alinearity2}  holds, and the asymptotic variance estimate is consistent $\widehat \Omega \rightarrow^p \Omega$.
\end{lemma}

\begin{proof}[Proof of Lemma \ref{lem:alinearity2}]
We verify Assumptions 1–3 of (\cite{LRSP}, CEINR) where notation is redefined as follows. When writing the residual $\lambda(Z, V)$, we make  the dependence on $\zeta$ in $\lambda(Z, V)$ explicit, that is,
$$
\lambda(Z, V, \zeta) = \beta V(X_{+}) - V(X) + \zeta(X).
$$
The following notation mapping from CEINR to this paper is used
\begin{align*}
\theta_{CNR} &:= \delta, \\
\gamma_{CNR} &:= (m(\cdot, \cdot), V(\cdot), \zeta(\cdot)), \\
w_{CNR} &=z \\
g_{CNR}(w_{CNR}, \gamma_{CNR}, \theta_{CNR})&= m(z, V) - \delta  \\
\alpha_{CNR} &= (\alpha,  \phi_{\zeta}, \phi_m) 
\end{align*}
Note that the correction term does not depend on the target parameter, that is, 
\begin{align*}
\phi_{CNR} (w_{CNR},  \gamma_{CNR}, \alpha_{CNR}, \theta_{CNR}) &= \phi_{CNR} (w_{CNR},  \gamma_{CNR}, \alpha_{CNR}) \\
&=\alpha(x) \lambda(z, V, \zeta) + \phi_{\zeta} (z) + \phi_m (z)
\end{align*}
Steps 1, 2, 3 verify Assumptions 1,2,3, respectively. Step 4 establishes \eqref{eq:consistencyomega}.

\textbf{ Step 1 }.  The term in Assumption 1(ii) of CEINR is bounded as 
\begin{align*}
&\int_{\mathcal{W}_{CNR}} (\phi_{CNR} (w_{CNR}, \alpha_{0,CNR}, \widehat \gamma_{\ell,CNR}) -\phi_{CNR}(w_{CNR},  \alpha_{0,CNR}, \gamma_{0,CNR}))^2 F_0(d w_{CNR}) \\
&\leq \int_{\mathcal{Z}}  \alpha^2_0(x) \lambda^2 (z, \widehat V_{\ell}-V_0, \widehat \zeta_{\ell} - \zeta_0)  F_0(dz) \\
&\leq \beta^2 \int_{\mathcal{Z}} \alpha^2_0(x) (\widehat V_{\ell} (x_{+}) - V_0(x_{+}))^2 d x_{+} +  \int_{\mathcal{Z}} \alpha^2_0(x) (\widehat V_{\ell} (x) - V_0(x))^2 d x \\
&+  \int_{\mathcal{Z}} \alpha^2_0(x) ( \widehat \zeta_{\ell} (x) - \zeta_0(x))^2 dx \\
&\leq [\sup_{x \in \mathcal{X}} \alpha^2_0(x)]  \left(  (\beta^2+1) \int_{\mathcal{Z}} (\widehat V_{\ell} (x) - V_0(x))^2 d z   +  \int_{\mathcal{Z}}  ( \widehat \zeta_{\ell} (x) - \zeta_0(x))^2 dx \right).
\end{align*}
The term in Assumption 1(iii) of CEINR is bounded as 
\begin{align*}
&\int_{\mathcal{W}_{CNR}} (\phi_{CNR} (w_{CNR},  \widehat \alpha_{\ell,CNR},\gamma_{0,CNR}) -\phi_{CNR} (w_{CNR},  \alpha_{0,CNR}, \gamma_{0,CNR}))^2 F_0(d w_{CNR}) \\
&\leq 3 \underbracket{\int_{\mathcal{Z}}  (\widehat \phi_{m\ell}(z) - \phi_{m} (z))^2 F_0(dz)}_{T_{1\ell}}  +3 \underbracket{ \int_{\mathcal{Z}}  (\widehat \phi_{\zeta\ell} (z) - \phi_{\zeta} (z))^2 F_0(dz)}_{T_{2\ell}} \\ 
&+ 3  \underbracket{\int_{\mathcal{Z}}  (\widehat \alpha_{\ell} (x) - \alpha_0(x))^2 \lambda^2 (z, V_0, \zeta_0)F_0(dz) }_{T_{3\ell}} = 3 \sum_{j=1}^3 T_{j\ell}. 
\end{align*}
Invoking Assumption \ref{ass:consistency} ensures that $T_{1\ell} + T_{2\ell}=o_P(1)$. To bound $T_{3\ell}$, we have 
$$
T_{3\ell} \leq   [\sup_{x \in \mathcal{X}} \E [ \lambda^2 (Z, V_0, \zeta_0) \mid X=x]  ] \int_{\mathcal{Z}}  (\widehat \alpha_{\ell} (x) - \alpha_0(x))^2 F_0(dx)
$$
which implies  $T_{3\ell}= o_P(1)$.

Finally, we verify Assumption 1(i) of CEINR.  By Assumption \ref{ass:ratealpha}(3),
\begin{align*}
 \int_{\mathcal{Z}} (m_0(z, \widehat{V}_{\ell}) -m_0(z, V_0))^2 F_0(dz) \leq C_m \int_{\mathcal{Z}} \| \widehat{V}_{\ell} - V_0 \|^2 F_0(dz)
\end{align*}
Likewise, by Assumption \ref{ass:crossm} with $\Delta_m(\cdot)$ defined therein 
\begin{align*}
&\int_{\mathcal{Z}}  (\widehat m_{\ell}(z, \widehat{V}_{\ell}) -m_0(z, \widehat{V}_{\ell}))^2 F_0(dz) \\
&\leq 2 \int_{\mathcal{Z}}  (\widehat m_{\ell}(z, V_0) -m_0(z, V_0))^2 F_0(dz) + 2\int_{\mathcal{Z}} \Delta^2_{m\ell} (z) F_0(dz)  \\
&\leq 2 C_m \int_{\mathcal{Z}} \| \widehat{V}_{\ell} - V_0 \|^2 F_0(dz) + 2 \int_{\mathcal{Z}} \Delta^2_{m\ell} (z) F_0(dz) = o_P(1) + o_P(1).
\end{align*}
Combining two bounds gives an upper bound 
\begin{align*}
&\int_{\mathcal{W}_{CNR}}  (g_{CNR}(w_{CNR}, \widehat \gamma_{\ell,CNR}, \theta_{0,CNR}) - g_{CNR}(w_{CNR},  \gamma_{0,CNR}, \theta_{0,CNR}))^2 F_0(dw_{CNR}) \\
&= \int_{\mathcal{Z}} (\widehat m_{\ell}(z, \widehat{V}_{\ell}) - m_0(z, V_0))^2 F_0(dz) \\
&\leq 2\int_{\mathcal{Z}}  (\widehat m_{\ell}(z, \widehat{V}_{\ell}) -m_0(z, \widehat{V}_{\ell}))^2 F_0(dz)  + 2\int_{\mathcal{Z}} (m_0(z, \widehat{V}_{\ell}) -m_0(z, V_0))^2 F_0(dz) = o_P(1).
\end{align*}

\textbf{ Step 2 }. We verify Assumption 2(i) of CEINR. Let $\ell \in \{1,2,\dots, L\}$ indicate the partition index and $\Delta_V(X)=V(X) - V_0(X)$. Define the following error terms
\begin{align*}
L_{1,\ell} (Z) &=  \beta (\widehat \alpha_{\ell} (X) - \alpha_0(X)) \Delta_{\widehat V_{\ell}} (X_{+})  \\
L_{2,\ell} (Z) &= - (\widehat \alpha_{\ell} (X) - \alpha_0(X))  \Delta_{\widehat V_{\ell}} (X)  \\
L_{3, \ell} (Z) &= (\widehat \alpha_{\ell} (X) - \alpha_0(X)) (\widehat \zeta_{\ell}(X) - \zeta_0(X)).
\end{align*}
Decomposing the error term in Assumption 2(i) gives
\begin{align*}
\widehat \Delta_{CNR, \ell} (W_{CNR}) = \sum_{j=1}^3 L_{j, \ell}(Z).
\end{align*}
Note that
\begin{align*}
n^{1/2} | \E [L_{1,\ell} (Z)  ] | + n^{1/2} \E [ L_{2,\ell} (Z) ] | &\leq  2 n^{1/2} \| \widehat \alpha_{\ell} - \alpha_0 \| \| \widehat V_{\ell} - V_0 \| = o_P(1)\\
n^{1/2} | \E [ L_{3,\ell}(Z) ] | &\leq n^{1/2} \| \widehat \alpha_{\ell} - \alpha_0  \| \| \widehat \zeta_{\ell} - \zeta_0 \|.
\end{align*}
Finally, since $\sup_{x \in \mathcal{X}} \widehat \alpha_{\ell} (x)$ is bounded a.s., we have $$\sup_{j \in \{1,2 \}} \E [ L^2_{j, \ell} (Z) ] \leq 2 C_{\alpha}^2 \| \widehat V_{\ell}  - V_0 \|^2 = o_P(1).$$ For the third term, 
$$
\E [ L^2_{3, \ell} (Z) ] \leq 2 C^2 \| \widehat \zeta_{\ell} - \zeta_0 \|^2 = o_P(1).
$$

\textbf{ Step 3 }. We verify Assumption 3(iv) of CEINR. Let $\ell \in \{1,2,\dots, L\}$ indicate the partition index. Define the following error terms
\begin{align*}
B_{1, \ell}  &=n^{1/2} \E [ \widehat m_{\ell} (Z,V_0) - m (Z, V_0) +  \widehat \phi_{m\ell}(Z) - \phi_m (Z) ]  \\
B_{2, \ell}  :&= n^{1/2} \E [ \alpha_0(X) (\widehat \zeta_{\ell}(X)- \zeta_0(X)) + \widehat \phi_{\zeta\ell}(Z) - \phi_{\zeta} (Z)  ]  \\
B_{3, \ell} :&=n^{1/2} \E [ \widehat m_{\ell} (Z, \widehat V_{\ell}) - m_0(Z, \widehat V_{\ell}) - \widehat m_{\ell} (Z, V_0) + m_0(Z, V_0) ]\\
B_{4, \ell} :&= \E [w_0(X) \Delta_{\widehat V_{\ell}} (X) +  \alpha_0(X) \lambda(Z, \widehat V_{\ell} - V_0) ].
\end{align*}
Notice that
\begin{align*}
&n^{1/2} \E [ \widehat m(Z, \widehat V_{\ell}) - m (Z, V_0) + \alpha_0(X) \lambda(Z, \widehat V_{\ell} - V_0, \widehat \zeta_{\ell} - \zeta_0)  \\
&+ \widehat \phi_{\zeta} (Z)  - \phi_{\zeta} (Z)+ \widehat \phi_{m\ell}(Z) - \phi_m (Z) ] = \sum_{j=1}^4 B_{j, \ell}.
\end{align*}
By Assumption \ref{ass:corhigher2}, $B_{1, \ell}+B_{2, \ell}= o_P(1)$; By Assumption \ref{ass:crossm}, $B_{3, \ell} = o_P(1)$. Lemma \ref{lem:orthod} implies $B_{4, \ell} =0$, which verifies Assumption 3(iv).  Assumption 3 (i) is automatically satisfied since
\begin{align*}
&\int_{\mathcal{Z}} \phi_{CNR} (w_{CNR}, \widehat \alpha_{\ell,CNR}, \gamma_{0,CNR}) F_0(d w_{CNR}) \\
&= \int_{\mathcal{Z}}  \widehat \alpha_{\ell}(x)  \lambda (z, V_0) F_0 (dz)  = 0.
\end{align*}
Likewise, by definition of $\phi_{\zeta}$ and $\phi_m$ as correction terms, $\int_{\mathcal{Z}}  \widehat{\phi}_{\ell m} (z) F_0 (dz) 
+ \int_{\mathcal{Z}}  \widehat{\phi}_{\ell  \zeta} (z) F_0 (dz) =0$.

\textbf{ Step 4 }. Finally, by the first conclusion, the estimator is asymptotically linear, and therefore consistent $\widehat \delta \rightarrow^p \delta_0$. We verifying the remaining conditions of Lemma 16 in CEINR. Decomposing the error of non-orthogonal moment  gives
\begin{align*}
&\int_{\mathcal{Z}}( \widehat m_{\ell} (z, \widehat V_{\ell}) -m_0(z, V_0) - \widehat \delta + \delta_0)^2 F_0 (dz) \\
&\leq 2  \int_{\mathcal{Z}}(\widehat m_{\ell} (z, \widehat V_{\ell})   -m_0(z, V_0))^2 F_0 (dz) + 2\int_{\mathcal{Z}} (\widehat \delta -\delta_0)^2 F_0 (dz) = o_P(1) + o_P(1) = o_P(1).
\end{align*}
The final condition has been verified in Step 2, which implies $\widehat{\Omega} \rightarrow^p \Omega$.

\end{proof}

\section{First-Stage Rates for Extremum Estimators}
\label{sec:fs}

\subsection{General Case Extremum Estimators}

In this Section, we  give a general statement for plug-in extremum estimators with a nuisance component, nesting Theorems \ref{thm1} and \ref{thm2} as special cases.  Abusing notation, let  $\ell (z,V,\omega): \mathcal{Z} \times \mathcal{V} \times \Omega \rightarrow \mathrm{R}$ denote an $M$-estimator loss function whose arguments are the data vector $z$, the scalar function $V$, and the vector-valued nuisance parameter $\omega: \mathcal{Z} \rightarrow \mathrm{R}^{d_{\omega}}$ whose true value $\omega_0$ is identified.  Define the true value $V_0$ as
\begin{align}
V_0 &= \arg \min L (V, \omega_0) \label{eq:v00}.
\end{align}
and
 \[
V^{\ast}= \arg\min_{V\in\mathcal{V}_n} L (V, \omega_0). 
\]
and a plug-in estimate 
\begin{align}
\label{eq:veesample}
\widehat V =  \arg \min_{\mathcal{V}_n} L^V_n (V)
\end{align}
where the  parameter space $\mathcal{V}$ can be approximated by a growing sieve space $\{ \mathcal{V}_n \}_{n \geq 1}$ and $\widehat \omega$ is estimated on auxiliary sample and $L^V_n(V)= n^{-1} \sum_{i=1}^n \ell (Z_i, V, \widehat \omega)$. The results in this section build on earlier work by \cite{NSS}, \cite{foster2019orthogonal}, and \cite{chernozhukov2024qm}.

\begin{assumption}
\label{ass:cons2}
For some sequence $\omega_n = o(1)$, the rate bound holds $\| \widehat {\omega} - \omega_0 \|_1 = O_P (\omega_n)$ where  $\| \omega \|_{1} = \sum_{j=1}^{d_{\omega}} \|  \omega_j  (Z) \|_{Z \sim P_Z}$.
\end{assumption}

Define
\begin{align*}
\starhull(\mathcal{V}_n-V_{0})=~  &  \{x\rightarrow
t\,(V(x)-V_{0}(x)):V%
\in\mathcal{V}_n,\text{ } t\in\lbrack0,1]\}.
\end{align*}
Let $\delta_n$ be  an upper bound on the critical radius of $\starhull(\mathcal{V}_n-V_{0})$ and all other mean square continuous transformations of $V$.

\begin{assumption}
\label{ass5}
(1)  (Identification) For some positive $ \bar{\lambda} \geq \underline{\lambda}>0$, the following lower bound holds
\begin{align}
\label{eq:identification}
 L (V, \omega_0) - L (V_0, \omega_0) \geq \underline{\lambda} \| V - V_0 \|^2, \quad \text{ for any } V \in \mathcal{V},
\end{align}
and the upper bound holds
\begin{align}
\bar{\lambda} \| V - V_0 \|^2 \geq  L (V, \omega_0) - L (V_0, \omega_0) \label{eq:identification2}.
\end{align}
(2) (Lipshitz in $(V,\omega)$) For some positive finite constant $C_L$, 
\begin{align}
\label{eq:lipshitz}
 |L (V, \omega) - L (V_0, \omega) - L (V, \omega_0) + L (V_0, \omega_0) | \leq C_L  \| V - V_0 \|   \| \omega - \omega_{0} \|_1 \text{ for any } V, \omega.
\end{align}
(3) For each $\omega$ in the realization set $\Omega$, the function $\ell(z, V, \omega)$ is $L$-Lipshitz in some mean square continuous (possibly, vector-valued) transformation of $V$ 
whose critical radius is upper bounded by $\delta_{n}$. Specifically, the following condition holds: with probability $1-\iota$, 
       \begin{align}
L (V, \omega) - L (V^{\ast}, \omega)  &\leq L_n (V, \omega) - L_n (V^{\ast}, \omega) +  \delta \| V-V^{\ast} \| + \delta^2, \label{eq:ineq}
 \end{align}
 where $\delta = \delta_n+ c_0 \sqrt{\ln (c_1/\iota)/n }$ for some $c_0, c_1>0$.

\end{assumption}

\begin{theorem}
\label{lem:mainrate}
Suppose Assumptions \ref{ass:cons2} and \ref{ass5} hold.  Then, with probability $1-\iota$, for some absolute constant $C$
\begin{equation}
\label{eq:mainlossrate}
\left\Vert \widehat{V}-V_{0}\right\Vert ^{2} \leq C \left(\delta_{n}^{2} + \omega^2_n + \| V^{\ast} - V_0 \|^2 + \frac{\ln(1/\iota)}{n}\right).
\end{equation}
\end{theorem}

 \begin{proof}[Proof of Theorem \ref{lem:mainrate}]
 
 Step 1 shows 
 \begin{align}
 \underline{\lambda}  \|  \widehat V - V^{\ast}  \|^2    - C_L \| \widehat \omega - \omega_0 \|_1 \| \widehat  V - V_0 \| \leq  \delta \|  \widehat V - V^{\ast}   \| + \delta^2 + 2\bar{\lambda} \| V^{\ast} - V_0 \|^2 . \label{eq:veehatv0}
\end{align}
Step 2 shows \eqref{eq:mainlossrate}.

\textbf{ Step 1. } Invoking triangular inequality gives
 \begin{align}
  \|  V - V^{\ast}  \|^2  - \|  V_0 - V^{\ast}  \|^2  \leq   \|  V - V_0 \|^2 \label{eq:vvv0} 
   \end{align}
 For any $V \in \mathcal{V}$ and the true value $V_0$,    
    \begin{align*}
& \underline{\lambda} \|  V - V_0 \|^2  - C_L \| \omega - \omega_0 \|_1 \| V - V_0 \|  \\
 &\leq^{(i)} L(V, \omega_0) - L(V_0, \omega_0) - C_L \| \omega - \omega_0 \|_1 \| V - V_0 \| \\
  &\leq^{(ii)}  L(V, \omega) - L(V_0, \omega)
  \end{align*}
  where (i) from Assumption \ref{ass5}(1) and (ii) from Assumption \ref{ass5}(2). Adding and subtracting approximation term $L (V^{\ast}, \omega) $ gives 
      \begin{align*}
  L(V, \omega) - L(V_0, \omega) &= L (V, \omega) - L (V^{\ast}, \omega) + L (V^{\ast}, \omega) - L(V_0, \omega) \\
 &\leq^{(iii)}  L (V, \omega) - L (V^{\ast}, \omega) +  \bar{\lambda} \| V^{\ast} - V_0 \|^2 
 \end{align*}
 where (iii) follows from  \eqref{eq:identification2}.  Plugging  $V := \widehat V$ and $\omega:= \widehat \omega$ into \eqref{eq:ineq} and invoking Assumption \ref{ass5}(3) 
 and   $$ L_n (\widehat  V,  \widehat \omega) - L_n (V^{\ast}, \widehat \omega) \leq 0,$$
 gives \eqref{eq:veehatv0}.

\textbf{ Step 2. } To show \eqref{eq:mainlossrate}, we replace product terms $  \delta \|  \widehat V - V^{\ast}   \|$ and $C_L \| \omega - \omega_0 \|_1 \| V - V_0 \| $ by  upper bounds.  Invoking AM-GM inequality for  $   \delta \|  \widehat V - V^{\ast}   \|$  gives 
\begin{align*}
 \delta \|  \widehat V - V^{\ast}   \| &\leq  \underline{\lambda}/4 \| \widehat V - V^{\ast}   \|^2 + \delta^2/  \underline{\lambda}.
 \end{align*}
 Invoking AM-GM inequality 
 \begin{align*}
  C_L \| \widehat \omega - \omega_0 \|_1 \| \widehat V - V_0 \| \leq \underline{\lambda}/4 \| \widehat V - V_0 \|^2 +   C^2_L/ \underline{\lambda} \| \widehat \omega - \omega_0 \|^2_1.
  \end{align*}
  Combining \eqref{eq:veehatv0} with AM-GM upper bounds  gives
\begin{align*}
3/4 \underline{\lambda}  \|  \widehat V - V^{\ast}  \|^2  \leq 2\bar{\lambda} \| V^{\ast} - V_0 \|^2  + \underline{\lambda}/4 \| \widehat V - V_0 \|^2  + C^2_L/ \underline{\lambda} \| \widehat \omega - \omega_0 \|^2_1 + \delta^2 (1 + 1/ \underline{\lambda}).
\end{align*}
Adding   $ 3/4 \underline{\lambda}  \| V^{\ast} - V_0 \|^2$ to each side and invoking 
$$
3/8 \|  \widehat V - V_0  \|^2  \leq 3/4 \|  \widehat V - V^{\ast}  \|^2 + 3/4  \| V^{\ast} - V_0 \|^2
$$
gives
\begin{align*}
3/8 \|  \widehat V - V_0  \|^2  \leq 3 \bar{\lambda} \| V^{\ast} - V_0 \|^2  + \underline{\lambda}/4 \| \widehat V - V_0 \|^2  + C^2_L/ \underline{\lambda} \| \widehat \omega - \omega_0 \|^2_1 + \delta^2 (1 + 1/ \underline{\lambda}).
\end{align*}
which implies 
\begin{align*}
 \underline{\lambda}/8\|  \widehat V - V_0  \|^2   \leq 3\bar{\lambda} \| V^{\ast} - V_0 \|^2 +     C^2_L/ \underline{\lambda} \| \widehat \omega - \omega_0 \|^2_1 + \delta^2 (1 + 1/ \underline{\lambda}).
 \end{align*}
Dividing by $ \underline{\lambda}/8$ gives
\begin{align*}
\|  \widehat V - V_0  \|^2   \leq 24\bar{\lambda}/ \underline{\lambda} \| V^{\ast} - V_0 \|^2 +     8C^2_L/ \underline{\lambda}^2 \| \widehat \omega - \omega_0 \|^2_1 +8 \delta^2 (1/\underline{\lambda} + 1/ \underline{\lambda}^2).
 \end{align*}
which matches \eqref{eq:mainlossrate} with $C \geq  \max (24\bar{\lambda}/ \underline{\lambda},  8C^2_L/ \underline{\lambda}^2, 8/\underline{\lambda},  8/ \underline{\lambda}^2)$ on the event $\| \widehat \omega - \omega_0 \|^2_1  \leq \omega^2_n$.

  \end{proof}

\section{First-Stage Rates for Neural Network Estimators of Value Function and Dynamic Dual Representation}

\label{sec:firststage}

In this Section, we derive mean square convergence rates for estimators of the value function and dynamic dual representation  based on Theorem \ref{lem:mainrate}. Let $\ell_V (z, V, \omega)$ denote an $M$-estimator loss function as in \eqref{eq:ellv}.

\begin{assumption}
\label{ass:cons}
(A) There exists a vanishing numeric sequence $\zeta_n = o(1)$ for which  $\| \widehat \zeta - \zeta_0 \| = O_P (\zeta_n)$. \newline
(B)  There exists a vanishing numeric sequence  $a_n = o(1)$ for which $\| \widehat A - A_0 \| = O_P (a_n)$. \newline
\end{assumption}

Assumption \ref{ass:cons} is a basic consistency condition for the plug-in estimators of per-period utility $\zeta$. For the dynamic binary choice  in Section \ref{sec:dynamicbinary}, the per-period utility $\zeta(x)$ is a known smooth transformation  of the conditional choice probability and a structural parameter given in \eqref{eq:zetax}--\eqref{eq:ux1}.

\begin{assumption}
\label{assmsq}
 The estimated operator $\widehat A$ belongs to a realization set $\mathcal{\bbA}_n$  with probability $1-o(1)$. For some constant $C_A$, for any $\bbA$  in $\mathcal{\bbA}_n$, 
\begin{align}
\sup_{\bbA \in \mathcal{\bbA}_n} \| \bbA \phi \|^2 \leq C_A \| \phi \|^2. \label{eq:meansq}
\end{align}
\end{assumption}

\begin{assumption}[Regularity conditions for the search set $\mathcal{V}_n$]
\label{ass2}
I) $\left\Vert f\right\Vert _{\infty}\leq 1$ for all $f\in{\mathrm{star}}(\mathcal{V}_n-V_{0})$ and  II) There exists a sequence $\delta_n= o(1)$ that is an upper bound on the critical radius of $\mathrm{star}(\mathcal{V}_n-V_{0})$.
\end{assumption}

Assumption \ref{ass2} is a standard regularity condition that depends on the class of estimators of the value function (see, e.g., Assumption 2 in \cite{chernozhukov2024qm}).

\begin{theorem}
\label{thm1} 
If Assumptions \ref{ass:cons} and  \ref{ass2} and \ref{assmsq} are satisfied, then, with probability $1-\iota$, for some constant $C_V$  large enough 
\[
\left\Vert \widehat{V}-V_{0}\right\Vert ^{2} \leq C_V \left( \delta^2_n + a^2_n + \zeta^2_n +\left\Vert V^{\ast}-V_{0}\right\Vert ^{2}+\frac{\ln(1/\iota)}{n}\right),
\]
where $V^{\ast}$ is the best approximation of $V_0$ by an element of $\mathcal{V}_n$, that is, 
\[
V^{\ast}=\arg\min_{V\in\mathcal{V}_n}  \E [ \ell_V (Z, V, \omega_0) ].
\]

\end{theorem}

Theorem \ref{thm1} gives a convergence rate in terms of the critical radius and the first-stage rate of plug-in estimators. To make use of this Theorem, one needs to know the size of the critical radius $\delta_n$ and the rate of approximation error $\left\Vert V^{\ast}-V_{0}\right\Vert ^{2}$. For the case of MLP neural network, those have been verified in \cite{chernozhukov2024qm}.

\begin{corollary}\label{cor1}
If (i) the support of $X$ is contained in a Cartesian product of compact intervals and $V_{0}(X)$ can be extended to a function that is continuously differentiable with $\beta_V$ continuous derivatives; (ii) $\mathcal{V}_n$ is an MLP network with $d_V$ inputs, width $K_V$, and depth $m_V$ with $K_V\to\infty$ and $m_V\to\infty$; (iii) for any estimate $\bbA \in \mathcal{A}_n$,  $(\bbA V)_{V \in \mathcal{V}_n}$ is representable as such a network;  then there is $C>0$ such that, for any $\varepsilon >0$, 
\[
\left\Vert \widehat{V}-V_{0}\right\Vert^{2}=O_{p}(K_V^{2}m_V^{2}\ln
(K_V^{2}m_V)\ln(n)/n+[K_Vm_V\sqrt{\ln(K^{2}_V m_V)}]^{-2(\beta_V/d_V)+\varepsilon_V} + \zeta^2_n + a^2_n).
\]
\end{corollary}

%\begin{lemma}[Verification of Assumption \ref{ass:cons} for density-based estimators of $\bbA$ and $\bbA^{*}$]
%\label{ex:density}
%Given estimated conditional densities $\widehat  f (x_{+} \mid x)$ and $\widehat f (x \mid  x_{+})$, define
%\begin{align*}
%(\widehat {\bbA} \phi) (x):&= \beta \int_{\mathcal{X}} \phi (x_{+}) \widehat  f (x_{+} \mid x) dx_{+} \\
%(\widehat {\bbA}^{*} \phi) (x_{+}) :&= \beta \int_{\mathcal{X}} \phi (x) \widehat f (x \mid  x_{+}) d x.
%\end{align*}
%Then, the plug-in estimators $\widehat {\bbA} $ and $\widehat {\bbA}^{*}$ obey Assumption \ref{ass:cons}(B) with $a_n \leq f_n$ and  Assumption \ref{assmsq} with $C_A \leq \beta C$, where $C$ is a uniform bound on the ratio $f(x_{+} \mid x)/f_0(x_{+} \mid x)$.
%\end{lemma}

Define
\begin{align*}
\starhull(\mathcal{A}_n-\alpha_{0})=~  &  \{x\rightarrow
t\,(\alpha(x)-\alpha_{0}(x)): \alpha \in \mathcal{A}_n,\text{ } t\in\lbrack0,1]\}.
\end{align*}
and
\begin{align*}
\starhull(m \circ \mathcal{A}_n- m \circ \alpha_{0})=~  &  \{x\rightarrow
t\,( m(z, \alpha) -m(z, \alpha_0)): \alpha \in \mathcal{A}_n,\text{ } t\in\lbrack0,1]\}.
\end{align*}

\begin{assumption}
\label{ass3}
(A) For some sequence $a^{*}_n = o(1)$ we have $\| \widehat A^{*} - A^{*}_0 \| = O_P (a^{*}_n)$. \newline
(B)   There exists a sequence $\delta_n= o(1)$ that is an upper bound on the critical radius of $\mathrm{star}(\mathcal{A}_n-\alpha_{0})$ and $\starhull(m \circ \mathcal{A}_n- m \circ \alpha_{0})$.
(C) The estimated operator $\widehat A^{*}$ belongs to a realization set  with probability $1-o(1)$. Each element of this realization set is mean square continuous, that is, \eqref{eq:meansq} holds for $\bbA^{*}$ for some absolute constant $C^{*}_A$.
\end{assumption}

\begin{theorem}
\label{thm2}
 If Assumptions \ref{ass:ratealpha}(C)  with $\bbA^{*}$ in place of $\bbA$ and \ref{ass3} are satisfied, then, with probability $1-\iota$, for some $C_{\alpha}$ large enough
\[
\left\Vert \widehat{\alpha}-\alpha_{0}\right\Vert ^{2} \leq C_{\alpha} \left( \delta^2_n + (a^{*})^2_n +\left\Vert \alpha^{\ast}-\alpha_{0}\right\Vert ^{2}+\frac{\ln(1/\iota)}{n}\right).
\]
where $\alpha^{\ast}$ is the  best approximation of $\alpha_0$ by an element of $\mathcal{A}_n$
\[
\alpha^{\ast}=\arg\min_{\alpha \in\mathcal{A}_n}  \E [ \ell_{\alpha} (Z, \alpha, \bbA^{*}_0) ].
\]

\end{theorem}

\subsection{Proofs of Results of Section \ref{sec:firststage}}
 \begin{proof}[Proof of Theorem \ref{thm1}]
The statement   Theorem \ref{thm1} is a special case of Theorem \ref{lem:mainrate} with $$\omega = (\zeta, \bbA), \quad \ell (z, V,\omega) =  \ell_V (Z, V, \omega). $$ Steps 1, 2,3 verify   Assumption \ref{ass5} (1)--(3), respectively.

\textbf{Step 1. } Consider  $\bar{\lambda} = (1+\beta)^2 \geq \underline{\lambda} = (1- \beta)^2>0$. Proposition \ref{prop:vee} gives
$$
L (V, \omega_0)  = \| (\bbI - \bbA_0) V - \zeta_0 \|^2.
$$
Taking $\Delta_V (X)  = V (X)- V_0 (X)$ gives 
\begin{align}
L (V, \omega_0) - L(V_0, \omega_0) &= \| (\bbI - \bbA_0) \Delta_V \|^2  \label{eq:milimet}. 
\end{align}
Stationarity implies $ \| \bbA_0 \Delta_V \|  \leq \beta \|  \Delta_V \|$. Thus, the lower bound \eqref{eq:identification} follows from 
\begin{align*}
\| \Delta_V \| (1 - \beta) = \| \Delta_V \| - \beta \| \Delta_V \| &\leq \| \Delta_V  \| - \| \bbA_0 \Delta_V \| \leq \| (\bbI - \bbA_0) \Delta_V \|,
\end{align*}
and the upper bound follows from 
$$
\| (\bbI - \bbA_0) \Delta_V \| \leq \| \bbI - \bbA_0 \|  \| \Delta_V \| \leq ( \| \bbI \| + \| \bbA_0 \|) \| \Delta_V \| \leq (1+\beta) \| \Delta_V\|.
$$

\textbf{Step 2. }  We verify Assumption \ref{ass5}(2) with $C_L = 3 (1+\beta) C + 2 (1+\beta)$ where here $C$ is an upper bound on $\| \Delta_V \| \leq \| V_0 \| \leq C$. Let $\Delta_{\zeta} = \zeta - \zeta_0$ and $\Delta_{\bbA} = \bbA - \bbA_0$.  For any $V \in \mathcal{V}_n$, note that
\begin{align*}
L (V, \omega) - L(V, \omega_0) &=  \E [ ((\bbI - \bbA_0) V)  ( (\Delta_{\bbA}) V - 2 (\Delta_{\zeta}) ) ]  \\
L (V_0, \omega) - L(V_0, \omega_0) &=  \E [ ((\bbI - \bbA_0) V_0)  ( (\Delta_{\bbA}) V_0 - 2 (\Delta_{\zeta}) ) ].
\end{align*}
Define the error terms
\begin{align*}
K_1 &= \E [ ((\bbI - \bbA_0) \Delta_V) \cdot  ( (\Delta_{\bbA}) \Delta_V  ] \\
K_2 &=  \E [ ( (\bbI - \bbA_0)  \Delta_V) \cdot   ( (\Delta_{\bbA}) V_0  ] \\
K_3 &= \E [ ( (\bbI - \bbA_0)  V_0) \cdot   ( (\Delta_{\bbA}) \Delta_V  ]   \\
K_4 &=-2 \E [ (\bbI - \bbA_0) \Delta_V \cdot  (\Delta_{\zeta}) )].
\end{align*}
and note that 
\begin{align*}
L (V, \omega) - L(V, \omega_0) - (L (V_0, \omega) - L(V_0, \omega_0)) &= \sum_{j=1}^4 K_j
\end{align*}
Cauchy inequality implies an upper bound 
\begin{align*}
| K_4 | \leq 2 \| (\bbI - \bbA_0) \Delta_V \| \| \Delta_{\zeta} \| \leq 2 (1+\beta)  \| \Delta_{\zeta} \|  \| \Delta_V \|,
\end{align*}
and 
$$
\max_{j \in \{1,2,3\}} | K_j |  \leq (1+\beta) C  \| \Delta_{\bbA} \| \| \Delta_V \|.
$$

\textbf{Step 3. }  We verify Assumption \ref{ass5}(3) with $\ell_V (z, V, \omega)=f_V (V (X), V(X_{+}), \bbA V)$ where
 $$f_V (t_1, t_2, t_3) = (t_1 - \beta t_2) ( t_1 t_3 - \beta t_2 t_3) - 2 (t_1 - \beta t_2) \zeta $$
that is Lipshitz with respect to the vector $(V (X), V(X_{+}), \bbA V (X))$. For any $\bbA \in \mathcal{A}_n$ and $\zeta$,  Lemma 11 from \cite{foster2019orthogonal} gives
 \begin{align*}
&|(L_n (V, \omega) - L_n (V^{\ast}, \omega)) - (L (V, \omega) - L(V^{\ast}, \omega)) | \\
& \leq^{i} O(\delta (\sqrt{ \| V (X)- V_0 (X) \|^2}  + \sqrt{\beta^2 \| V (X_{+}) - V_0 (X_{+}) \|^2} +\sqrt{ C \| \bbA V - \bbA V_0 \|^2 }) + \delta^2) \\
&\leq^{ii} O(\delta  \| V - V_0 \|  (1+\beta) +\sqrt{ C \| \bbA V - \bbA V_0 \|^2 })  + \delta^2) \\
&\leq^{iii} O(\delta  \| V - V_0 \|  (1+\beta + \sqrt{C C_A})  + \delta^2)
\end{align*}
where (i) follows from Lemma 11,  (ii) follows from stationarity and $\| \Delta_V (X_{+}) \| = \| \Delta_V (X) \|$ and (iii) from MSE continuity of $\bbA$ stated  in Assumption \ref{assmsq}.

\end{proof}

 \begin{proof}[Proof of Theorem \ref{thm2}]
The statement   Theorem \ref{thm1} is a special case of Theorem \ref{lem:mainrate} with $$\omega = \bbA^{*}, \quad \ell (z, V, \bbA^{*}) =  \ell_{\alpha} (Z, V, \bbA^{*}). $$ Thus, it suffices to verify Assumption \ref{ass5} (1)--(3).

\textbf{Step 1. } Stationarity implies $ \| \bbA_0 \Delta_{\alpha} \|  \leq \beta \|  \Delta_{\alpha} \|$. An argument similar to Step 1 in the proof of Theorem \ref{thm1} verifies Assumption \ref{ass5}(1) with  $\bar{\lambda} = (1+\beta)^2 \geq \underline{\lambda} = (1- \beta)^2>0$.

\textbf{Step 2. }  We verify Assumption \ref{ass5}(2). For any $\alpha \in \mathcal{A}_n$, note that
 \begin{align*}
L (\alpha, \bbA^{*} ) - L(\alpha, \bbA^{*}_0)  &=  \E [ ((\bbI - \bbA^{*}_0) \alpha)  (\bbA^{*} - \bbA^{*}_0) \alpha  - 2m (Z,  (\bbA^{*} - \bbA^{*}_0)\alpha ) ]  \\
L (\alpha_0, \bbA^{*} ) - L(\alpha_0, \bbA^{*}_0) &=  \E [ ((\bbI - \bbA^{*}_0) \alpha_0)  (\bbA^{*} - \bbA^{*}_0) \alpha_0  - 2m (Z,  (\bbA^{*} - \bbA^{*}_0)\alpha_0 ) ] 
\end{align*}
Subtracting the second line from the first one gives 
\begin{align*}
L (\alpha, \bbA^{*} ) - L(\alpha, \bbA^{*}_0) - (L (\alpha_0, \bbA^{*} ) - L(\alpha_0, \bbA^{*}_0)) &= \sum_{j=1}^4 K^{*}_j
\end{align*}
where
\begin{align*}
K^{*}_1 &= \E [ ((\bbI - \bbA^{*}_0) (\alpha-\alpha_0)) \cdot  (\bbA^{*} - \bbA^{*}_0) (\alpha-\alpha_0)  ] \\
K^{*}_2 &=  \E [ ( (\bbI - \bbA^{*}_0)  (\alpha-\alpha_0)) \cdot    (\bbA^{*} - \bbA^{*}_0) \alpha_0  ] \\
K^{*}_3 &= \E [ ( (\bbI - \bbA^{*}_0)  \alpha_0) \cdot    (\bbA^{*} - \bbA^{*}_0) (\alpha - \alpha_0)  ]   \\
K^{*}_4 &=-2 \E [ m (Z,  (\bbA^{*} - \bbA^{*}_0) (\alpha - \alpha_0 ))].
\end{align*}
Invoking Riesz reprepresentation \eqref{eq:rieszweight} gives
$$
K^{*}_4 = -2 \E [ w_0(X)  (\bbA^{*} - \bbA^{*}_0) (\alpha - \alpha_0)   (X) ]. 
$$
Invoking Cauchy Schwartz gives an upper bound 
\begin{align*}
| K^{*}_4 | \leq 2 \| w_0 \| \| \bbA^{*} - \bbA^{*}_0 \| \| \alpha - \alpha_0 \|.
\end{align*}
For the remaining terms, assuming $\| \alpha - \alpha_0 \| \leq \| \alpha_0 \| \leq C$ gives
\begin{align*}
\max_{j \in \{1,2,3\}} | K^{*}_j |  &\leq \|  (\bbI - \bbA^{*}_0) \| \| \alpha_0 \|  \| \bbA^{*} - \bbA^{*}_0 \| \| \alpha - \alpha_0 \| \\
&\leq (1+ \beta) \| \alpha_0 \|  \| \bbA^{*} - \bbA^{*}_0 \| \| \alpha - \alpha_0 \|.
\end{align*}

\textbf{Step 3. }  We verify Assumption \ref{ass5}(3) with $$f_{\alpha} (t_1, t_2, t_3, t_4) = (t_1 - \beta t_2) ( t_1 t_3 - \beta t_2 t_3) - 2  t_4 $$
that is Lipshitz with respect to the vector  $(\alpha (X), \alpha(X_{-}), \bbA^{*} \alpha, m (Z, (\bbI -  \bbA^{*}) \alpha )  )$.  For any $\bbA^{*} \in \mathcal{A}_n$,  Lemma 11 from \cite{foster2019orthogonal} gives
\begin{align*}
&|(L_n (\alpha, \bbA^{*} ) - L_n (\alpha^{\ast}, \bbA^{*})) - (L (V, \bbA) - L(V^{\ast}, \bbA^{*})) | \\
& \leq^{i} O( (\delta ( \| \alpha - \alpha_0 \| (1+\beta) + \sqrt{  \|  m (Z, (\bbI -  \bbA^{*}) \Delta_\alpha ) \|^2 })  + \delta^2)\\
&\leq^{ii} O( (\delta ( \| \alpha - \alpha_0 \| (1+\beta)  )  + \delta^2)\
\end{align*}
where (i) follows from invoking Lemma 11 in \cite{foster2019orthogonal} and  stationarity and (ii) from Assumption \ref{ass:ratealpha}(C) as well as from MSE continuity of $\bbA^{*}$ assumed in the analog of  Assumption \ref{assmsq}.

  \end{proof} 
 
%\begin{proof}[Proof of Lemma \ref{ex:density}]
%If there exists $f_n$ such that 
%\begin{align*}
%\| \widehat  {f} (x_{+} \mid s,k) - f (x_{+} \mid s, k) \|_{\infty} = O_P (f_n),
%\end{align*}  then Assumption \ref{ass:cons}(B) is satisfied with $a_n = f_n$.  Furthermore, taking $\mathcal{A}_N = \{  \bbA \phi = \beta \int_{\mathcal{X}} \phi (x_{+}) f (x_{+} \mid x) d x,  \quad  f(x_{+} \mid x) \leq C_f \}$ gives a realization set for the operator $\bbA$ obeying Assumption \ref{ass:cons}(C) with $C_A = \beta^2 C_f$
%\begin{align*}
%\sup_{ \bbA \in \mathcal{A}_n} \|  \bbA \phi \|^2 \leq \beta^2 \int_{\mathcal{X}} \phi^2 (x_{+}) \dfrac{ f (x_{+} \mid x)}{f_0( x_{+} \mid x)} f_0( x_{+} \mid x) dx_{+}   f_X(x) dx \leq \beta^2 C_f \| \phi \|^2.
%\end{align*}
%
%
%
%
%\end{proof}

\section{Extension to Nonlinear Models}
\label{sec:nonlinear}

In this Section, we generalize the results of Sections \ref{sec:setup}--\ref{sec:estimation} to allow the value function $V$ to appear nonlinearly in the model. Consider a semiparametric moment problem
\begin{align}
\label{eq:maincmrmoment}
\E [ m (Z, \delta_0, V_0)  ] =0,
\end{align}
where  $Z \in \mathcal{Z} \subseteq \mathrm{R}^Z$ is the data vector, $\delta$ is a  finite-dimensional target  parameter whose true value is $\delta_0$,  and $V_0$ is value function defined in \eqref{eq:npv2}.  A special case of this problem is considered in \eqref{eq:linfunc} where $m(Z,V)$ is assumed linear in $V$. In this Section, we allow both $\delta_0$ and $V_0$ to appear nonlinearly in the model.

The problem above can be translated into the theoretical  framework of Section \ref{sec:setup} via linearization of the moment function with respect to $V$, following \cite{Newey1994}. Let $D(Z, V)$  denote a function of $Z$ and a linear function $V(\cdot)$ (i.e. $D(Z,V)$ is a linear functional of $V$.) We assume that for all $V$ such that $\| V - V_0 \|$ small enough, for some positive finite constant $\bar B< \infty$, 
 \begin{align}
\| m(Z, \delta_0, V) - m(Z, \delta_0, V_0) - D (Z, V - V_0)  \| \leq \bar{B} \|  V - V_0 \|^2. \label{eq:linearization}
\end{align} 
We will impose throughout that the expectation $\E[D(Z, V)]$ is mean square continuous as a function of $V$, meaning that there is a constant $C$ such that for all $V(X)$ with $\E[V(X)^2]<\infty$,
\begin{align}
|\E[D(Z, V)]|\leq C (\E[V(X)^2])^{1/2},
\end{align}
which is equivalent to existence of a function $w_{D0}(X)$ with $\E[w_{D0}(X)^2]<\infty$ such that 
\begin{align}
\label{eq:riesz2}
\E[D(Z,V)]=\E[w_{D0}(X)V(X)], 
\end{align}
for all $V(\cdot)$ with $\E[V(X)^2]<\infty$ (see also e.g., \cite{Newey1994}). In contrast to \eqref{eq:rieszweight}, the LHS of \eqref{eq:riesz2} involves the derivative functional rather than the original functional itself.

Let $\alpha_0$ be the dynamic dual representation corresponding to $w_{D0}$ in \eqref{eq:riesz2}. The following extremum characterization applies for the $M$-estimator loss function
\begin{align}
\label{eq:propalphad}
\ell^D(Z, \alpha, \omega) = (\alpha(X)-\beta \alpha(X_-))((\bbI - \bbA^{*}) \alpha)(X)- 2D(Z,(\bbI - \bbA^{*})\alpha), \quad \omega = (\bbA^{*}, D).
\end{align}
where true value of $\omega$ is $\omega_0 = (\bbA^{*}_0, D_0)$. When $\beta=0$, the Proposition \ref{prop:alpha2} recovers the extremum representation of \cite{chernozhukov2024qm}.

\begin{proposition}[Extremum Representation of $\alpha_0$ in Nonlinear Case]
\label{prop:alpha2}
The dynamic  Riesz representer $\alpha_0$ in \eqref{eq:bdp}  is the unique minimizer of the quadratic objective function  
\begin{align}
\alpha_0&=\arg \min_{\alpha}  \E [(\alpha(X)-\beta \alpha(X_-))((\bbI - \bbA^{*}) \alpha)(X)- 2D(Z,(\bbI - \bbA^{*})\alpha)]. \label{eq:alpha5} \\
&= \arg \min_{\alpha}  \E [((\bbI - \bbA^{*}) \alpha)(X)^2- 2((\bbI - \bbA^{*}) \alpha)(X)w_{D0}(X)]
\nonumber \\
&= \arg \min_{\alpha}  \E [(((\bbI - \bbA^{*}) \alpha)(X)- w_{D0}(X))^2] \nonumber.
\end{align}
\end{proposition}

Finally, let us describe the construction of the orthogonal moment for $\delta_0$.  Let $D(Z,V,F)$ denote the plim of the estimated $D(Z,V)$ function, $\phi_m(Z)$ the influence function of $\E[D(Z,V_0,F)]$, and $\phi_{\zeta}(Z)$ the influence function of $\E[\alpha_0(X)\zeta(X,F)]$. The orthogonal moment is
\begin{align}
\label{eq:timevarmoment2}
\psi(Z, \gamma, \phi, \delta)  &= m(Z, \delta, V) + \alpha(X)(\beta V(X_{+}) - V(X) + \zeta(X))
+ \phi_m(Z) + \phi_\zeta(Z), \\
\gamma &=(m,V,\zeta), \phi=(\alpha,\phi_m,\phi_\zeta). \nonumber
\end{align}

\begin{remark}[Riesz representation of  Conditional Expectations of $V(X)$]
\label{lem:condriesz}
For any mean square integrable function $\rho(z) \in \mathcal{L}_2$, the linear functional
\begin{align}
\label{eq:mainlinearfunctional}
D_{\rho}(Z, V):= \E[ \rho(Z) V(X_{+}) \mid X]
\end{align}
is also mean square integrable. Its Riesz representation takes the form
\begin{align}
\label{eq:riesz3}
\E[D_{\rho} (Z,V)]=\E[w_{D0}(X_{+})V(X_{+})] =\E[w_{D0}(X)V(X)], \quad w_{D0} (X_{+}):= \E[ \rho(Z) \mid X_{+}].
\end{align}
\end{remark}

\section{Dynamic binary choice revisited}
\label{sec:dynamicbinary2}

In this Section, we demonstrate the application of Appendix \ref{sec:nonlinear} for a single-agent dynamic binary choice  model obeying Assumptions 1 and 2 in  \cite{AMira2002}, described herein.
At each period $t \in \{0,1,\dots, \infty\}$, the agent chooses an action $j$ in a binary set $\mathcal{A} = \{1, 2\}$
after observing the state realization $x$ and the vector of private shocks $\epsilon = (\epsilon (j))_{j \in \mathcal{A}}$ independent of states. 
The shocks $\epsilon (j)$ are independent across $\mathcal{A}$ and type 1 extreme value distributed.  
For each action $j$, the state process $(X_t \mid X_{t-1}, A_t=j)_{t \geq 1}$ is a  first-order time-homogeneous Markov chain.
 The action $j$ results in $u(x,j) + \epsilon(j)$  immediate (i.e., per-period) utility.  
 The agent chooses actions according to the optimal Markov policy
\begin{align}
\label{eq:agentchoice}
\arg \max_{j \in \mathcal{A}}  \{ v(x,j) + \epsilon(j)\},
\end{align}
where $v(x,j)$ is the choice-specific value function defined as
\begin{align}
    \label{eq:choicespec}
    v(x,1) &=  u(x,1) +\beta \E [ V_0(X_{+}) \mid X=x,J=1] \\
     v(x,0) &=  u(x,0) +\beta \E [ V_0(X_{+}) \mid X=x,J=0], 
\end{align}
and $V_0(x)$ is the agent's value function, determined as a fixed point of the Bellman equation
\begin{align}
    \label{eq:bellman}
    V_0(x)     &=\E_{\epsilon} \big[\max_{j \in \mathcal{A}} \big[   v(x,j)  + \epsilon(j) \mid X=x\big] \big].
    \end{align}
As demonstrated in    \cite{HotzMiller} and \cite{AMira2002}, $V_0(x)$  can be represented as a  net present value \eqref{eq:npv2}  based on the per-period utility  $\zeta_0(x)$ that we introduced in \eqref{eq:zetax}, Section \ref{sec:setup}.  Furthermore, the log-odds ratio
\begin{align}
\label{eq:hotzmiller}
\ln p_0(x) - \ln (1-p_0(x))  &= v(x,1) - v(x,0) 
 \end{align}
simplifies to
 \begin{align}
\label{eq:hotzmiller2}
v(x,1) - v(x,0)  = D(x)' \theta_0 + \beta ( \E [ V_0(X_{+}) \mid X=x,J=1] -  \E [ V_0(X_{+}) \mid X=x,J=0])
 \end{align} 
 for linear index utilities as in \eqref{eq:ux1}. Introducing expectation functions for the choices $J=1$ and $J=0$
\begin{align}
\label{eq:gamma2}
V \rightarrow  \gamma^{2}_0(x, V)  &=\E[ V(X_{+}) \mid X=x, J=1], \\
V \rightarrow \gamma^{1}_0(x, V)  &=\E[ V(X_{+}) \mid X=x, J=0]. \label{eq:gamma1}
\end{align}
 gives a succinct form of log-odds ratio equality
  \begin{align}
 v(x,1) - v(x,0)  = D(x)' \theta_0 + \beta ( \gamma^2_0(x,V_0) -  \gamma^2_1(x,V_0) ).
 \end{align} 
 Applying  the logistic transformation to \eqref{eq:hotzmiller2} gives
\begin{align}
\label{eq:cmr}
\E [ J - \Lambda (D(x)' \theta_0 + \beta ( \gamma^2_0(x,V_0) -  \gamma^2_1(x,V_0)   ) \mid X=x] =0,
\end{align}
where $\Lambda(t)$ and $\Lambda'(t)$ denote  the logistic CDF and the logistic PDF, specifically
\begin{align}
\label{eq:logit}
\Lambda(t) = \dfrac{\exp t}{\exp t +1}, \qquad \Lambda'(t) = \Lambda(t) \cdot (1-\Lambda(t)).
\end{align}
For this approach to yield  constructive identification,  a separate identification argument for value function $V_0(x)$ must be provided. For instance, if the choice $J=0$ has a renewal choice property such a replacement action in bus replacement problem \cite{Rust} or sterilization choice \cite{HotzMiller},   $V_0(x)= C - \ln (1-p_0(x))$  for some $C$ is identified (Example 3, \cite{LRSP}).

\subsection{Identification of $\theta_0$. }

In what follows, let us provide a missing argument for identification of structural parameter does not require $J=0$ to be replacement, a question studied in \cite{AdsmEck2022}.   The first part of the argument exploits  linearity of per-period utility $u(x,1)$ and $u(x,0)$ in $\theta$ and  is standard in the literature.
It  shows that value function $V_0(x)$ can be expressed as a linear function in $\theta$.  Define the ``slope'' function as $$\zeta^S_0 (x) :=(- (1-p_0(x)), D_1(x)p_0 (x))^{\prime}$$  and the ``intercept'' function as
$$
\zeta^I_0(x) := H(p_0(x)), \quad H(t) = \gamma_e - t\ln t -(1-t) \ln (1-t).
$$
Next, consider  a linear form  of $\theta$
\begin{align}
\label{ex:exante}
\zeta(x,\theta) = \zeta^S_0 (x) '\theta +\zeta^I_0(x),
\end{align}
which coincides with the per-period utility $\zeta_0(x)$ in \eqref{eq:zetax} at the true value $\theta=\theta_0$. Likewise,  the linear form 
\begin{align}
\label{eq:valuelinear}
V(x, \theta) =: V^S_0(x)^{\prime} \theta + V^I_0(x)
\end{align}
constructed using the slope and intercept functions
\begin{align}
\label{eq:valuelinear2}
V^S_0 (x)= \sum_{t \geq 0} \beta^t \E [ \zeta^S(X_t) \mid X=x], \quad V^I_0 (x)= \sum_{t \geq 0} \beta^t \E [ \zeta^I(X_t) \mid X=x]
\end{align}
matches the value function  $V_0(x)$ in \eqref{eq:npv2} at at the true value $\theta=\theta_0$. Our proposal is to replace value function $V$ in  \eqref{eq:cmr} by a linear form $V(x, \theta)$,  treating $V^S, V^I$ as nuisance functions. Then, \eqref{eq:cmr} is a semiparametric conditional  moment restriction whose parametric component is $\theta$ and \textit{identified} functional component is $h_0(x) = (V^S_0(x), V^I_0(x),\gamma^2_0(x), \gamma^1_0(x))$ where the true values of expectation functions $\gamma^1, \gamma^2$ are given in \eqref{eq:gamma2}--\eqref{eq:gamma1} and true values of $V^S, V^I$ are given in \eqref{eq:valuelinear2}.  Taking 
\begin{align*}
 \upsilon_1 (X,h) &= D(x) +  \beta (\gamma^2(x, V^S) - \gamma^1(x,V^S)) \\
 \upsilon_0 (X,h) &= \beta (\gamma^2_0(x, V^I) - \gamma^1_0(x,V^I)) \\
 \upsilon(X,h) &= ( \upsilon^{\prime}_1 (X,h),  \upsilon^{\prime}_0 (X,h))^{\prime}
\end{align*}
gives
$$
\upsilon (x,h_0)^{\prime} \theta_0 = v(x,1) - v(x,0)  = \ln p_0(x) - \ln (1-p_0(x)).
$$
 Invoking logistic transform to both sides gives
$$
p_0(x) = \Lambda (\upsilon (x,h_0)^{\prime} \theta_0)
$$
which gives rise to an unconditional moment restriction
\begin{align}
\label{eq:ucmr}
g_{\theta} (Z, \theta, h) = \upsilon(X, h ) (J - \Lambda ( \upsilon (X,h)^{\prime} \theta )).
\end{align}

\subsection{Overview of estimation and inferential strategy. }

Our estimation strategy for the structural parameter $\theta_0$ is to combine the extremum representation of $V^S_0, V^I_0$  with the orthogonal moment $\psi$ which we derive below. We  follow the outline of Section \ref{sec:estimation} and begin with the derivation of   debiased moment function. The moment function  \eqref{eq:timevarmoment2} reduces to
\begin{align}
\psi(Z, h, \alpha, \theta)  &= \upsilon(X,h) (J - \Lambda ( \upsilon (X,h)^{\prime} \theta_0  )) + \sum_{j=1}^2 \phi_{\gamma^j} (Z,h) \label{eq:line1} \\
&+ \alpha(X)(\beta V(X_{+})  - V(X) + \zeta(X)) +   \phi_{\zeta} (Z,h) \label{eq:line3} 
\end{align}
where  the first two correction terms $\phi_{\gamma^1}$ and  $\phi_{\gamma^2} $  in \eqref{eq:line1} account for the estimation of $ \gamma^1$ and $\gamma^2$, respectively and the third one in \eqref{eq:line3} accounts for the  nuisance components of linear form  $V(x, \theta)$.  A standard argument for expectation functions  (\cite{Newey1994}, Proposition 4, p. 1361) gives
\begin{align}
\label{eq:phigamma2}
\phi_{\gamma^2}(Z, h) &=  \alpha_2(X, J) ( V(X_{+}) -\gamma_2( X,V) ) \\
\phi_{\gamma^1}(Z, h) &=  \alpha_1(X, J) ( V(X_{+}) -\gamma_1( X,V) ), \label{eq:phigamma1} 
\end{align}
where the true values of bias correction functions $\alpha_2(X,J)$  and $ \alpha_1(X, J)$ are 
\begin{align}
\label{eq:alphaj}
\alpha_{20}(X, J) &=  - \beta \upsilon(X) J (1-p_0(X)) \\
\alpha_{10}(X, J)&=   \beta \upsilon(X)  (1-J) p_0(X). \nonumber
\end{align}
The third correction term in line \eqref{eq:line3}  is specified up to $\alpha$, which can be derived as described in Section \ref{sec:nonlinear}, and the correction term $\phi_{\zeta}$.  Note that the linearization functional  \eqref{eq:linearization}
\begin{align}
\label{eq:frechet}
D(Z, V) :&=   -\beta \cdot \upsilon(X) \cdot p_0(X) \cdot  (1-p_0(X)) \cdot  \left( \gamma^2_0(X, V)  - \gamma^1_0(X, V) \right) \\
&= - \beta \E[ (\alpha_2(X,J) - \alpha_1(X,J)) V(X_{+}) \mid X] \nonumber
\end{align}
is a special case of  a linear functional  \eqref{eq:mainlinearfunctional} in Remark \ref{lem:condriesz} with $\rho(Z) = \alpha_2(X,J) - \alpha_1(X,J)$. Thus, the true value of $\alpha$ is the dynamic dual representation based on weighting function 
\begin{align}
\label{eq:rieszrepresenter}
w_{D0}(x) &=- \beta \E[ \upsilon(X) (J - p(X)) \mid X_{+}=x]
\end{align}
whose automatic characterization is given in Proposition \ref{prop:alpha2} with $D(Z,V)$ in \eqref{eq:frechet}.  Finally, the fourth correction term accounts for estimation of choice probability $p(x)$ \textit{only} \begin{footnote}{This approach is different from Section \ref{sec:dynamicbinary} where both $p_0(x)$  and $\theta$ are treated as nuisance components.} \end{footnote}
that enters linearly in $\zeta^S (x)$ and nonlinearly in $\zeta^I(x) = H(p(x))$. The correction term is
\begin{align}
\label{eq:phiv}
\phi_{\zeta}(Z,h) &= - \beta \alpha(X) (\gamma^2 (X, V) - \gamma^1(X,V)) (J-p(X)).
\end{align}
 
\begin{remark}[Static models with $\beta=0$]
Notice  all correction terms $\phi_{\gamma^2}(Z, h), \phi_{\gamma^1}(Z, h), \phi_{\zeta}(Z,h)$ and $\alpha(x)$ in \eqref{eq:alphaj}--\eqref{eq:phiv} depend multiplicatively on $\beta$ and, therefore, reduce to zero if $\beta=0$. In this case, the moment equation \eqref{eq:ucmr}  reduces to a FOC for the logistic regression
$$
\E [ D(X) (J - \Lambda (D(X)^{\prime} \theta_0)) ] =0,
$$
which has no nuisance parameters.  The same property hold for the debiased moment condition in Example 3, \cite{LRSP}.  
\end{remark}

Recent work by \cite{AdsmEck2022} considered semiparametric  estimation and inference method for a structural parameter in dynamic discrete choice models.  The approach proposed in Appendix \ref{sec:dynamicbinary2} complements their results in several respects.  First, the authors consider data arriving in state-action pairs $((X,J), (X_{+}, J_{+}))$
while this paper  considers standard triplets $(X,J, X_{+})$, following \cite{AMira2002} and \cite{LRSP}.  Consequently, the nuisance parameters of the moment condition proposed therein are based on \textit{choice-specific} value functions $(v(x,j))_{j \in \mathcal{A}}$, while the nuisance functions of \eqref{eq:valuelinear2}--\eqref{eq:ucmr} are  slope and intercept of \textit{value function} $V_0(x)$.

Algorithm \ref{alg:leebound2} sketches the proposed estimator of $\theta$.

 \begin{algorithm}
\small
\begin{algorithmic}[1]
	\STATE  Partition the set of data indices $\{1,2,\dots, n\}$ into  $L$ disjoint subsets of about equal size with where $L$ is an odd number $L \geq 3$.
	\STATE   Let $I^c_{\ell} = (Z_i)_{i \notin I_{\ell}}$ denote the set observations \textit{not} in $I_{\ell}$.   Partition  $I^c_{\ell} = I^{1}_{\ell} \sqcup I^{2}_{\ell}$ into two halves. For each nuisance parameter $  \omega$, let $\widehat \omega^{1}_{\ell}, \omega^{2}_{\ell}$ denote the estimator computed on $I^{1}_{\ell}$ and $ I^{2}_{\ell}$, respectively.  Estimate
	\begin{enumerate}
	\item[a] The choice probability function $ p_0(x)= \Pr (J=1 \mid X=x)$
	\item[b] The  expectation functions $V \rightarrow \gamma^1(x,V)$ and $V \rightarrow \gamma^2(x,V)$ as in \eqref{eq:gamma1} and \eqref{eq:gamma2}.
	\item[c] The forward operator $\bbA$ in \eqref{eq:operatorlin} and backward operator $\bbA^{*}$ in \eqref{eq:operatorlindual}

	\end{enumerate}
	
	\STATE  Estimate ``slope'' and ``intercept'' value functions  by minimizing sample cross-fit objective function 
\[
\widehat{V}^S_{\ell} = \arg\min_{V \in \mathcal{V}_n} 
\bigg[
\sum_{i \in I^{c1}_\ell} \ell^S (Z_i, \widehat \omega^{2}_{\ell} ) + \sum_{i \in I^{c2}_\ell} \ell^S (Z_i, \widehat \omega^{1}_{\ell})
\bigg], \qquad \widehat \omega = (\widehat {\bbA}, \widehat p)
\]
\[
\widehat{V}^I_{\ell} = \arg\min_{V \in \mathcal{V}_n} 
\bigg[
\sum_{i \in I^{c1}_\ell} \ell^I (Z_i, \widehat \omega^{2}_{\ell} ) + \sum_{i \in I^{c2}_\ell} \ell^I (Z_i, \widehat \omega^{1}_{\ell})
\bigg], \qquad \widehat \omega = (\widehat {\bbA}, \widehat p)
\]
where $ \mathcal{V}_n$ is some set of functions and  $ \ell^S$ and $\ell^I$ are  special cases of $M$-estimator losses in \eqref{eq:ellv} based on $\zeta^S(x)$ and $\zeta^I(x)$, respectively.

\STATE Estimate dynamic dual representation by minimizing sample cross-fit objective function 
\[
\widehat{\alpha}_{\ell} = \arg\min_{\alpha \in \mathcal{A}_n} 
\bigg[
\sum_{i \in I^{c1}_\ell} \ell^D (Z_i,  \widehat{p}_{\ell}, \widehat{\gamma}^2_{\ell},  \widehat{\gamma}^1_{\ell},  \widehat{\bbA}^{2*}_{\ell}) + \sum_{i \in I^{c2}_\ell} \ell^D(Z_i,  \widehat{p}_{\ell}, \widehat{\gamma}^2_{\ell},  \widehat{\gamma}^1_{\ell},   \widehat{\bbA}^{1*}_{\ell})
\bigg],
\]
where $ \mathcal{A}_n$ is some set of functions and  $ \ell^D$ is a special case of $M$-estimator losses in \eqref{eq:propalphad} based on $D(Z,V)$.

\STATE Estimate the nuisance parameters $(\widehat p_{\ell}, \widehat  \gamma^2_{\ell},  \widehat  \gamma^1_{\ell})$ using \textit{ all } observations in $I^c_{\ell}$. Define a collection of estimated functions as  $\widehat h_{\ell} = (\widehat p_{\ell}, \widehat  \gamma^2_{\ell},  \widehat  \gamma^1_{\ell}, \widehat (V^S_{\ell})^\prime \widehat{\theta}_{\ell} + \widehat V^I_{\ell}, \widehat (\zeta^S_{\ell})^\prime \widehat{\theta}_{\ell} + \widehat \zeta^I_{\ell})$ where $\widehat{\theta}_{\ell} $ is a preliminary estimator of $\theta$ based on a non-orthogonal moment condition.

\STATE Estimate the structural parameter using debiased GMM estimator

\begin{align*}
\frac{1}{n} \sum_{\ell = 1}^L \sum_{i \in I_\ell}  \psi (Z_i, \widehat \theta, \widehat h_{\ell}) =0  
\end{align*}

\end{algorithmic}
\caption{Debiased Estimator of Structural Parameter}
\label{alg:leebound2}
\end{algorithm}

\newpage
\subsection{Auxiliary Empirical Details}
     
     \begin{table}[h]
\caption{Descriptive Statistics of Key Variables}
\centering
\begin{tabular}{lcccc}
  \toprule
  & Full & Low  & Medium & High  \\
  & Sample & Score $\in [3,30]$ & Score $\in [30,40]$ & Score $\in [40,54]$ \\
  \midrule
  School Open & 0.78 & 0.73 & 0.80 & 0.80 \\
  & (0.41) & (0.45) & (0.40) & (0.40) \\
  \\
Teacher Test Scores & 35.08 & 20.22 & 36.61 & 43.92 \\ 
 & (10.68) & (6.89) & (2.33) & (3.67) \\ 
Days Worked & 17.15 & 16.07 & 17.64 & 17.50 \\ 
 & (5.55) & (6.29) & (5.18) & (5.18) \\ 
Work Streak & 3.61 & 3.10 & 3.73 & 3.86 \\ 
 & (5.55) & (4.94) & (5.68) & (5.82) \\ 
    \midrule
  Number of Observations &  19016 &  5245 &  5984 &  7787\\
  Number of Teachers & 57 & 16 & 17 & 24 \\
  \bottomrule
\end{tabular}
\label{table:ds}
\caption*{Notes: This table reports sample means and standard deviations of School Open, Days Worked,  Work Streak and  Teacher Test Score where  the latter variable is used to categorize teachers into 
Low-, Medium-, and High-score groups. The data sample comes from \cite{DHR} and spans teacher attendance data between January 1, 2004, and June 30, 2005, excluding holidays and Sundays when all schools were closed. See text for details.}
\end{table}

\bibliographystyle{chicagoa}

\bibliography{/Users/virasemenova/Desktop/my_new_bibtex}

\end{document}